\documentclass[lettersize,journal]{IEEEtran}
\usepackage{amsmath,amsfonts}
\usepackage{algorithmic}
\usepackage{algorithm}
\usepackage{array}
\usepackage{textcomp}
\usepackage{stfloats}
\usepackage{url}
\usepackage{verbatim}
\usepackage{cite}
\usepackage{threeparttable}
\usepackage{diagbox}
\usepackage{subfigure}
\usepackage{graphicx}  
\usepackage{amssymb}
\usepackage{booktabs}  
\usepackage{multirow}  

\begin{document}
\title{Boosting Adversarial Transferability for Hyperspectral Image Classification Using 3D Structure-invariant Transformation and Weighted Intermediate Feature Divergence}

\author{Chun Liu, Bingqian Zhu, Tao Xu, Zheng Zheng, Zheng Li, Wei Yang, Zhigang Han, Jiayao Wang

\thanks{(Corresponding author: Tao Xu.) }

\thanks{Chun Liu and Tao Xu are with the State Key Laboratory of Spatial Datum, College of Remote Sensing and Geoinformatics Engineering, Faculty of Geographical Science and Engineering; the School of Computer and Information Engineering; and the Henan Industrial Technology Academy of Spatio-Temporal Big Data, Henan University, Zhengzhou, Henan 450046, China (e-mail: liuchun@henu.edu.cn; txu@henu.edu.cn).

Bingqian Zhu, Zheng Li, and Wei Yang are with the School of Computer and Information Engineering, Henan University, Zhengzhou, Henan 450046, China (e-mail: 104754231411@henu.edu.cn; lizheng@henu.edu.cn; yangwei@henu.edu.cn).

Zheng Zheng is with the National Key Laboratory of Integrated Aircraft Control Technology, School of Automation Science and Electrical Engineering, Beihang university, Beijing 100091, China (e-mail: zhengz@buaa.edu.cn).

Zhigang Han and Jiayao Wang are with the State Key Laboratory of Spatial Datum, College of Remote Sensing and Geoinformatics Engineering, Faculty of Geographical Science and Engineering; and the Henan Industrial Technology Academy of Spatio-Temporal Big Data, Henan University, Zhengzhou, Henan 450046, China (e-mail: zghan@henu.edu.cn; wjy@henu.edu.cn).
}

\thanks{}}

\markboth{ }
{Shell \MakeLowercase{\textit{et al.}}: A Sample Article Using IEEEtran.cls for IEEE Journals}

\maketitle

\begin{abstract}
Deep Neural Networks (DNNs) are vulnerable to adversarial attacks, which pose security challenges to hyperspectral image (HSI) classification based on DNNs. Studying adversarial attacks on DNN models for HSI classification helps reveal model vulnerabilities but also provides critical directions for improving their resilience. Numerous adversarial attack methods have been designed in the domain of natural images. However, different from natural images, HSIs contains high-dimensional rich spectral information, which presents new challenges for generating adversarial examples. Based on the specific characteristics of HSIs, this paper proposes a novel method to enhance the transferability of the adversarial examples for HSI classification using 3D structure-invariant transformation and weighted intermediate feature divergence. While keeping the HSIs structure invariant, the proposed method divides the image into blocks in both spatial and spectral dimensions. Then, various transformations are applied on each block to increase input diversity and mitigate the overfitting to substitute models. Moreover, a weighted intermediate feature divergence loss is also designed by leveraging the differences between the intermediate features of original and adversarial examples, which enhances the transferability of generated adversarial examples from the perspective of optimizing the perturbation generation process. It constrains the perturbation direction by enlarging the feature maps of the original examples, and assigns different weights to different feature channels to destroy the features that have a greater impact on HSI classification. Extensive experiments demonstrate that the adversarial examples generated by the proposed method achieve more effective adversarial transferability on three public HSI datasets. Furthermore, the method maintains robust attack performance even under defense strategies.
\end{abstract}


\section{Introduction}
\IEEEPARstart{W}{hile} obtaining geometric features of object surfaces through two-dimensional spatial imaging technology, hyperspectral images (HSIs) employ spectral demodulation techniques to interpret the total radiant energy of each pixel as a continuous spectrum across different bands \cite{1}. They build a three-dimensional data cube that integrates the spatial-spectral information of ground targets, enabling the capture of more subtle differences between them. Due to this advantage, HSIs have important application value in meteorology \cite{2}, environmental protection \cite{3}, agriculture \cite{4}, and military fields \cite{5}.

In recent years, DNNs have become the dominant approach for HSI classification due to their powerful feature extraction capabilities \cite{57,11,9,59}. Researchers have developed various methods based on convolutional neural networks (CNNs), recurrent neural networks (RNNs), and other DNN models to extract more discriminative spatial and spectral features from HSIs \cite{wang,Mahmoodi,Cai}. For example, Wang et al. \cite{wang} introduced a dynamic super-pixel normalization (DSN)-based DNN for HSI classification; Mahmoodi et al. \cite{Mahmoodi} developed an effective method capable of focusing on both spectral and spatial features. When considering special circumstances such as insufficient training samples, more methods have been presented based on the techniques like few-shot learning for HSI classification \cite{66,67,68}.

Despite demonstrating significant advantages in HSI classification tasks, DNNs are highly vulnerable to adversarial attacks \cite{12,13}, which cause machine learning models to produce erroneous predictions via imperceptible subtle perturbations added to input example. These crafted inputs are referred to as adversarial examples. Therefore, it is of great significance to study adversarial attacks on the DNN models for HSI classification, which helps reveal the potential vulnerabilities of these models but also provides critical directions for improving their resilience.

Previous research on adversarial attacks mainly focuses on white-box settings where adversaries possess the target model's internal information such as architecture, parameters, and training data. Many classic methods have been proposed along this route, including FGSM \cite{24} and subsequent iterative methods \cite{13,26,27}. In contrast, black-box attacks do not require such access and generate adversarial examples using known substitute model, which are thus more realistic. Among the black-box attacks \cite{14,15}, the attacks based on adversarial transferability \cite{16,17} are more flexible, where transferability refers to the fact that the adversarial examples obtained on the substitute model still have an attacking effect on the unknown target model. Nevertheless, adversarial examples generated using substitute model often overfit to the architecture and feature representations of that model. To address this, various input transformation methods (e.g.,\cite{28,30,34}) have been developed to improve adversarial transferability. For example, the diverse input method (DI) \cite{30} convert the input with a probability $p$, achieved by randomly resizing the image and padding zeros at its edges. At the same time, some methods enhance transferability by modifying feature representations or altering the model's attention \cite{17,21,41,42}. For example, Xu et al. \cite{62} proposed a new black-box adversarial attack method called Mixup-Attack, which finds common vulnerabilities between different networks by attacking the shallow-layer feature of a given substitute model.

However, since the HSIs contains not only a large amount of spatial information but also rich spectral information, this presents new challenges for generating adversarial examples. In previous methods, Shi et al. \cite{19} focused on the robustness of generated adversarial examples against common image processing methods. They generated superpixel segmentation templates and spectral clustering templates in spatial and spectral dimensions, respectively, and constrained the perturbation values within each small block to be the same, improving the global smoothness of the generated perturbations. Although this approach takes into account the spectral information of HSIs, it has not explored the structure of HSIs which contains critical semantics information, and the issue of adversarial examples overfitting to the substitute model has not been alleviated. In addition, existing methods for generating adversarial examples for HSI classification models usually rely solely on classification layer information to guide perturbation optimization. Nevertheless, studies have found that adversarial perturbations generated from intermediate feature maps often have the highest transferability \cite{23}. Although methods such as \cite{62,63} employ shallow-layer feature distance or feature approximation as loss functions, they failed to constrain the perturbation direction to weaken the features related to the true class. And in the feature space, they relied only on overall gradient information, without considering the different contributions of different channels in HSIs.  

To address these issues, this study proposes a method to boost transferability of adversarial examples for HSI classification by using 3D structure-invariant transformation and weighted intermediate feature divergence. In the line of the transformation based black-box attacks, it exploits the large number of spectral bands of HSIs and partition HSIs into 3D blocks across both spatial and spectral dimensions. Random transformation strategies are applied to each block to obtain more diverse inputs, so as to alleviate the problem of overfitting to the substitute models. Meanwhile, inspired by Zhu et al. \cite{23}, a novel weighted intermediate feature divergence loss is also designed from the un-targeted attack perspective. To guide the perturbations to suppress features associated with the true class, it first enlarges the feature values of the original example by a fixed factor, ensuring that the adversarial example's feature values are lower. Next, it computes the variance of each channel within the feature map to assign different weights to different channels, thereby destroying the features that have a greater impact on HSI classification.

Our contributions are summarized as follows:
\begin{enumerate}

\item{By making full use of the large number of spectral bands in HSIs, we propose a 3D structure-invariant transformation method to divide HSIs into blocks in both spatial and spectral directions and apply random transformation strategies to each block. While keeping the global structure unchanged and maintain global semantic information, it enhances the diversity of HSIs and alleviates the problem of generated adversarial examples overfitting to the substitute model.}

\item {We design a weighted intermediate feature divergence loss that enhances the transferability of generated adversarial examples by increasing the differences between the intermediate features of the original and adversarial examples. It constrains the perturbation direction by enlarging the feature maps of the original example, and assigns different weights to different feature channels to destroy the features that have a greater impact on HSI classification.}

\item {We propose a novel attack framework to boost the transferability of adversarial examples for HSI classification by utilizing 3D structure-invariant transformations together with the weighted feature divergence loss. Our framework is combined with two gradient-based attack methods. Extensive experiments on three real HSIs datasets show that the proposed method outperforms the baseline approaches with significant advantages, demonstrating its effectiveness.}
\end{enumerate}

The rest of this paper is organized as follows: Section 2 reviews related works relevant to this study. Section 3 provides a detailed explanation of the proposed adversarial attack methodology. Section 4 introduces the datasets information, experimental settings, and summarizes the experimental results. Section 5 concludes the study.

\section{Related Work}
\subsection{Traditional White-box Attack Methods}
Gradient-based methods are a common and effective category of adversarial attack techniques. The core idea is to maximize the model's loss using its gradient information to generate adversarial examples. Goodfellow et al. \cite{24} proposed the Fast Gradient Sign Method (FGSM), which generates powerful adversarial examples based on the linear properties of CNNs. Wang et al. \cite{13} and Madry et al. \cite{26} further extended FGSM by decomposing the one-step perturbation generation into an iterative process, and proposed I-FGSM and PGD attacks. MI-FGSM \cite{27} introduced a momentum term in iterative attacks, which accumulated previous gradient vectors to guide the calculation of the next gradient. Subsequent methods such as NI-FGSM \cite{28} and VMI-FGSM \cite{29} were also developed. Although these methods achieve high attack success rates on white-box models, their transferability to black-box models is often limited due to the significant differences in gradient information between models.

\subsection{Input Transformation-based Attack Methods}
The approach based on input transformation enhances the diversity of input data by applying specific data transformation operations to the inputs. The DI \cite{30} method is one representative approach. TI \cite{32}, SIM \cite{28}, and Admix attack \cite{34} are classical input transformation techniques that enhance diversity. Dong et al. \cite{35} proposed a superpixel-guided attention attack method, which constrains the perturbation values within each superpixel to be the same, thereby improving the robustness of adversarial examples against defenses based on image processing. Long et al. \cite{36} diversified inputs by performing transformations in the frequency domain. Unlike previous methodologies, Guo et al. proposed MixCam \cite{38}, which enhances adversarial transferability by shifting the regions of model attention. Additionally, Wang et al. introduced BSR \cite{39}, improving the consistency of adversarial example attention heatmaps across different models. Liu et al. \cite{40} proposed a local mixup method to improve the existing mixup method. The transformed examples obtained by these methods can be extended the model's attention areas, mitigate the overfitting to the substitute model during the generation of adversarial examples, and thereby improve the success rate of the attack.

\subsection{Advanced Loss Functions-based Attack Methods}
Yosinski et al. \cite{21} were the first to propose that the features extracted by the network have transferability across models with different architectures. Later, Zhou et al. proposed Transferable Adversarial Perturbations (TAP) \cite{17}, which improve transferability by increasing the feature distance between the original and adversarial images in intermediate layers. Inka et al. \cite{42} modified the features of the original image to align its intermediate-layer representation becomes closer to that of another class. On the other hand, Wu et al. \cite{43} approached the problem by extracting model attention, prioritizing the disruption of key features that may be adopted by multiple model architectures. Li et al. \cite{44} introduced a method that mixes features of clean and adversarial examples, attenuating intermediate-level perturbations from clean features. Additionally, Wang et al. \cite{45} also proposed to increases feature diversity by diversifying the high-level features. Since the mapping distance of intermediate feature maps correlates with the distance of the final output of neural networks \cite{17}, these methods increase the likelihood of erroneous predictions and enhance the success rate of transferability attacks.

\subsection{Adversarial Attack methods in Remote Sensing}
In recent years, research in the area of adversarial examples within the remote sensing field has been steadily increasing. Chen et al. \cite{61} analyzed the vulnerability of CNN-based remote sensing image (RSI) recognition models to adversarial examples and found that models with different architectures exhibit varying degrees of vulnerability when trained on the same RSI dataset. Later, Xu et al. \cite{46} and Chen et al. \cite{47} investigated the performance of classic attack methods in remote sensing scene classification tasks. Burnel et al. \cite{50} used a Wasserstein Generative Adversarial Network-based approach to generate more natural adversarial examples for remote sensing. Methods based on implicit neural representations and feature approximation (i.e., approximating the features of the original image to a virtual image not belonging to any class) provided new insights for this research \cite{rs1,63}. The latest DMFAA method \cite{rs3} is based on training a substitute model using a distillation model, integrating features from other models. Shi et al. \cite{19} obtained smoother examples in the spatial and spectral dimensions through constraints based on superpixels and clustering, exploring their adversarial transferability and robustness. Previous studies that typically added perturbations globally, Peng et al. \cite{rs4} proposed a semi-black-box adversarial attack method based on one-pixel attacks. 

To mitigate the threat of adversarial attacks, various defenses have been proposed. Shi et al.\cite{51} introduced an attention-based adversarial defense model that extracts and separates attack-invariant and attack-variant features from adversarial examples, using an attention-guided reconstruction loss to address the attention shift caused by perturbations. Xu et al. \cite{52} leverage the global context information of HSIs to significantly improve model robustness against adversarial examples. They also proposed the Spatial-Spectral Self-Attention Learning Network \cite{53}, which integrates spatial and spectral information through a multi-scale spatial attention module and global spectral transformer. Similarly, \cite{rs5,rs6} integrated spatial and spectral information for defense, indicating that holistic consideration of global information is an effective defense strategy. Furthermore, Zhao et al. \cite{rs7} proposed a novel anti-perturbation network that enhances the robustness of HSI classification under adversarial attacks by integrating pixel-level features and superpixel priors. To the best of our knowledge, previous studies have paid little attention to improving the transferability of adversarial attacks in HSI classification, which we believe is also of significant research importance.

\section{Proposed Method}
\subsection{Preliminary}
In this study, we refer to the known model used to generate adversarial examples as the substitute model, denoted as $f$. The input HSIs are denoted as $x\in X\subset{R^{C\times H \times W}}$, where $H$ and $W$ represent the height and width, respectively, and $C$ is the number of spectral bands. The ground-truth label of $x$ is $y$, and the adversarial example is $ x_{adv} \in X \subset R^{C\times H \times W}$. The goal of this study is to find a small perturbation $\delta \subset R^{C\times H \times W}$ that satisfies the constraint condition and enables the adversarial example $x_{adv}=x+\delta$ to maximally mislead the model's predictions. The loss function is denoted as $L\left(\cdot\right)$. So the process of generating adversarial examples is naturally achieved by maximizing the loss function, which can be expressed as:
\begin{equation}
{x_{adv}} = \arg \max L\left(f\left(x_{adv}\right),y\right), s.t.{\left\| \delta \right\|_p} \le \varepsilon 
\label{equ1}
\end{equation}

Typically, the $l_p$-norm is used to constrain the perturbation $\delta$ (in this paper, the $l_1$-norm is utilized), where $\varepsilon$ is the perturbation budget.

\begin{figure*}[t]
    \centering
    \includegraphics[width=1\textwidth]{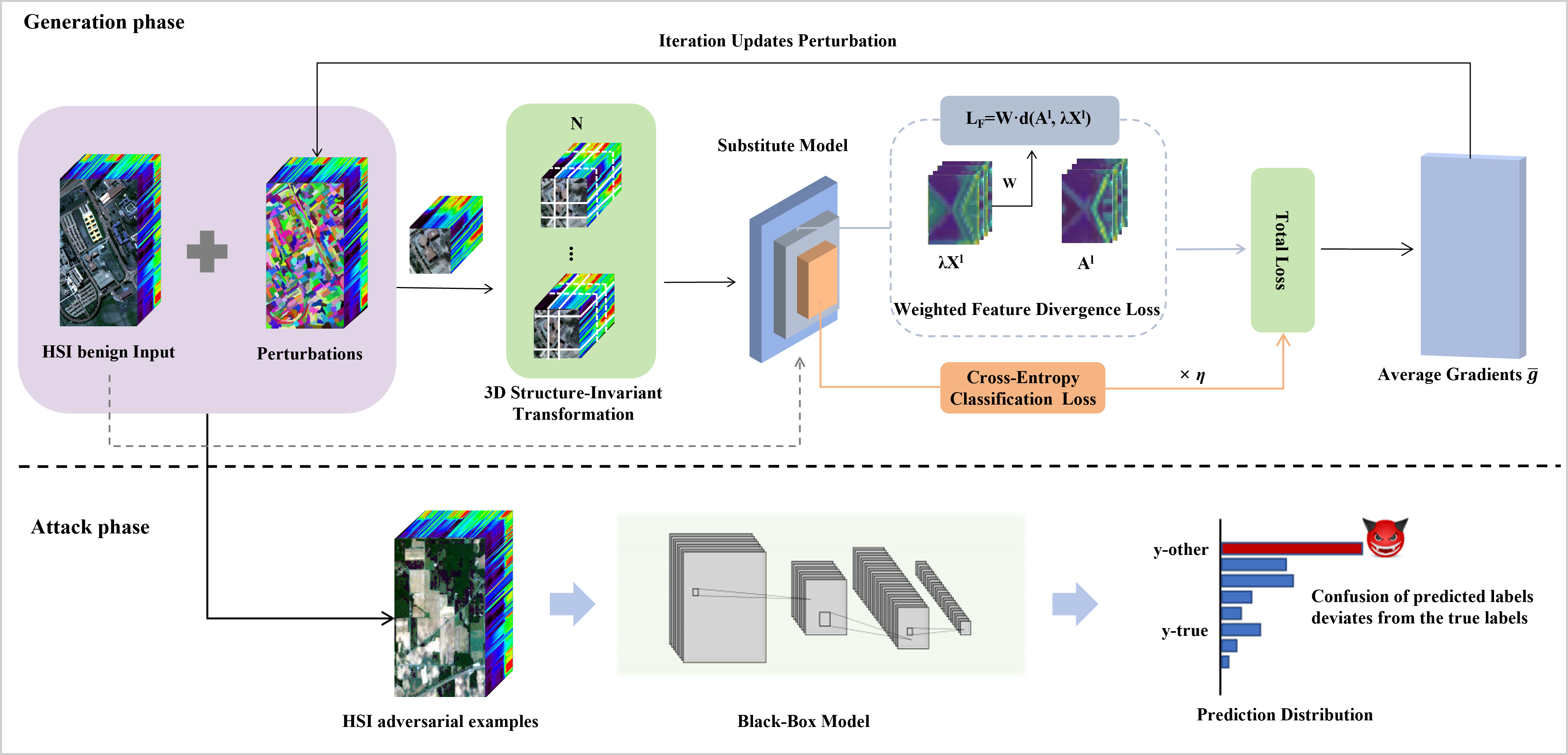}  
    \caption{The overall workflow of the proposed method. Before feeding the image into the substitute model, the 3D structure-invariant transformation method is applied to adversarial examples, which divides the examples into blocks in the spatial and spectral dimensions and then uses random transformation operation to each block. While preserving the image structure, it not only helps generate more diverse images but also maintain the global semantic information. Subsequently, the weighted feature divergence loss and cross-entropy classification loss are computed to obtain the total loss, so as to obtain the average gradient, which is iteratively updated to refine the perturbation. Here, $X^l$ and $A^l$ denote the feature maps of the original and the adversarial examples at layer $l$, respectively.} 
    \label{framework image}  
\end{figure*}

\subsection{Overall Process}
The pipeline of our method is illustrated in Fig. \ref{framework image}. The transferability of adversarial examples is enhanced through two key steps: 3D structure-invariant transformation and weighted feature divergence loss. Before inputting the adversarial examples into the substitute model, $n$ copies will be generated. These $n$ copies are randomly divided along both the spatial and spectral dimensions. Subsequently, each block in the copy will be randomly applied with a data transformation operation to generate a set of enhanced transformed images, which will serve as input for the substitute model. Then, we calculate the weighted feature divergence loss $L_F$ between the original example and the adversarial example, which is combined with the cross-entropy classification loss to form the total loss function. Because the transformed images may differ significantly from the original ones, the calculation of gradients could introduce instability. To solve this problem, we calculate the average gradient according to the average loss of multiple transformed images. Algorithm \ref{adversarial_algorithm} outlines the perturbation generation process integrated with the baseline method MI-FGSM \cite{27}, where $Clip_{x,\varepsilon}$ means to crop the adversarial examples in [$x - \varepsilon, x + \varepsilon$]. The details of the two key steps are introduced below.

\subsection{3D Structure-Invariant Transformation}
Data transformation methods can mitigate the issue of adversarial examples overfitting to the substitute model. However, previous methods usually do not take the image structure into account when transforming the images. This may result in the enhanced image losing important semantic information. Therefore, this paper proposes the structure-invariant method to transform the HSI images when generating adversarial examples. Specifically, given an image $x$, we randomly divide it into blocks and then apply different transformation strategies to each block. In this process, the relative structural relationship between the image blocks can be completely preserved. Since transforming each block does not impair the task such as target recognition \cite{20}, this structure-invariant transformation strategy not only helps generate more diverse images but also maintain the global semantic information which are important for downstream tasks.

\begin{figure}[htbp]
\label{Transform image}  
\centering
\subfigure[]{
    \begin{minipage}[b]{.4\linewidth}
        \centering
        \includegraphics[height=3.2cm,width=3.2cm]{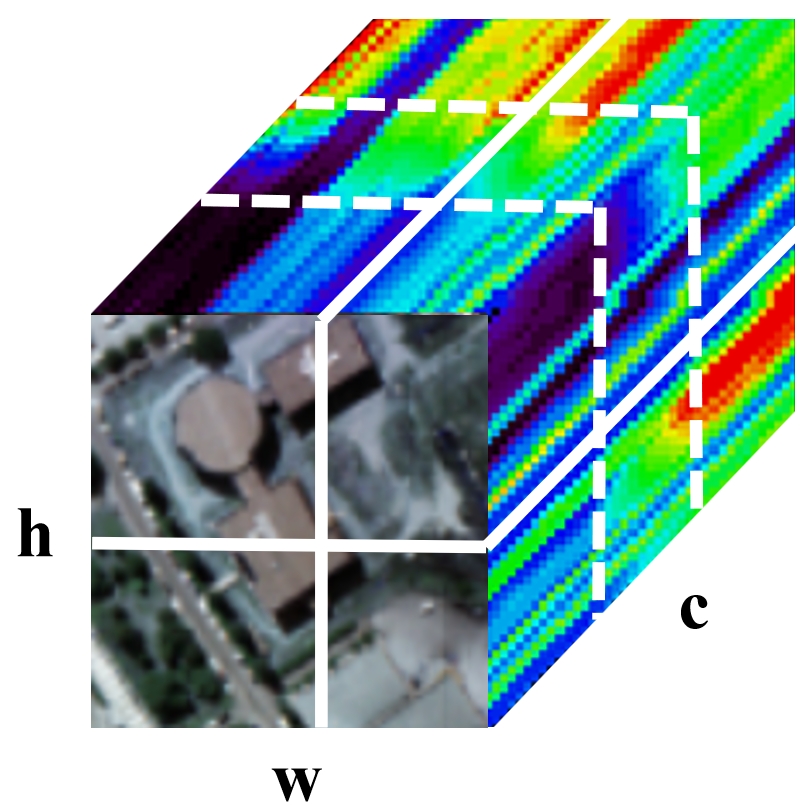}
        \label{subfig:divide}
    \end{minipage}}
\subfigure[]{
 	\begin{minipage}[b]{.4\linewidth}
        \centering
        \includegraphics[height=3.1cm,width=3.1cm]{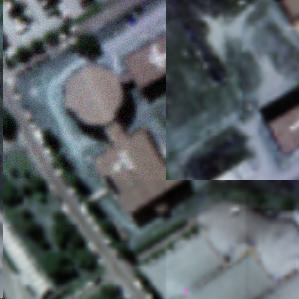}
        \label{subfig:trans example}
    \end{minipage}}
\caption{(a) Schematic of image division. If the spatial dimension division parameter is set to 2, and the spectral dimension is set to 3, then a 2×2×3 block structure will be obtained; (b) Schematic of image transformation. Taking spatial transformation as an example, the image is divided into 2×2 blocks, with different transformations applied to each block as shown.} 
\end{figure}

Since HSIs have rich spatial and spectral information, dividing the HSI can be carried out in both spatial and spectral dimensions. When the structure-invariant transformation is performed only in the spatial dimension, i.e., 2D structure-invariant transformation, the transformation effect will be applied to all spectral channels of each block. Nevertheless, different DNN models usually focus on distinct discriminative spectral bands \cite{19}. Therefore, in order to further improve the diversity of HSIs, we adopt the 3D structure-invariant transformation to HSIs. That is, we randomly divide each image into $a \times a$ blocks in the spatial dimension and $b$ blocks in the spectral dimension, and finally get the three-dimensional block structure of $a\times a\times b$, as shown in Fig. \ref{subfig:divide}. Then, we randomly select different transformation operations from the candidates and apply them to different blocks, thereby generating a set of diverse image variants. The transformation diagram is shown in Fig. \ref{subfig:trans example}. The candidate transformation operations used to blocks are summarized in Table \ref{Table:transformations}. In order to avoid information loss, we add constraints on some operations, such as limiting the scope of scale transformations, preventing over-dark images.

\begin{table*}[b]
\caption{The various transformation methods adopted in our approach and the resulting transformed images}
\centering
\scalebox{0.9}{
    \begin{tabular}{ccccccccccc}
    \toprule
    Raw & VShift & VFlip & Scale & Rotate & Resize & HShift & HFlip & Dropout & DCT & AddNoise \\
    \midrule
    \includegraphics[width=0.08\textwidth]{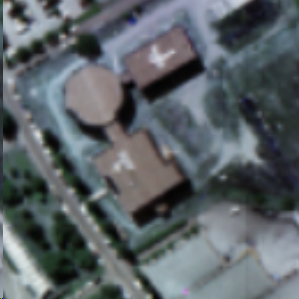} &
    \includegraphics[width=0.08\textwidth]{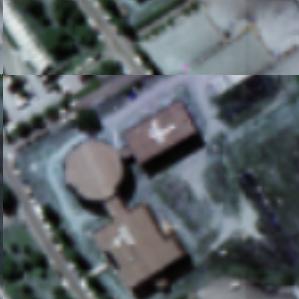} &
    \includegraphics[width=0.08\textwidth]{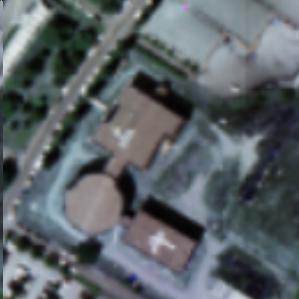} &
    \includegraphics[width=0.08\textwidth]{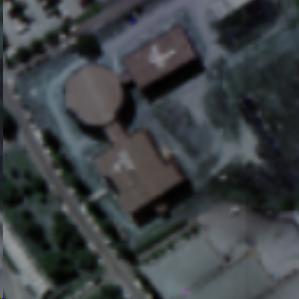} &
    \includegraphics[width=0.08\textwidth]{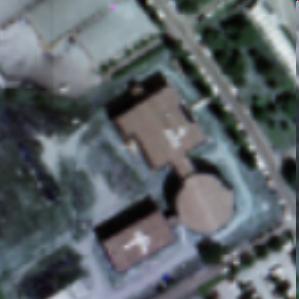} &
    \includegraphics[width=0.08\textwidth]{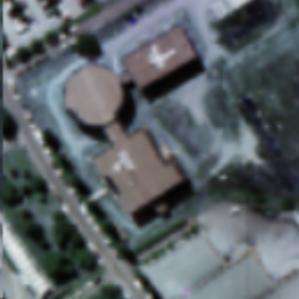} &
    \includegraphics[width=0.08\textwidth]{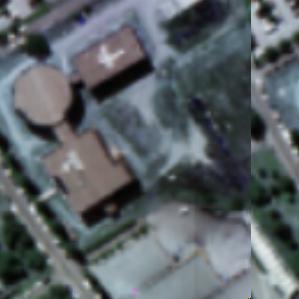} &
    \includegraphics[width=0.08\textwidth]{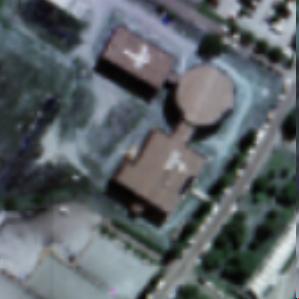} &
    \includegraphics[width=0.08\textwidth]{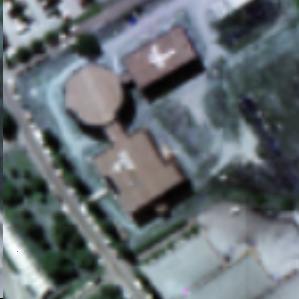} &
    \includegraphics[width=0.08\textwidth]{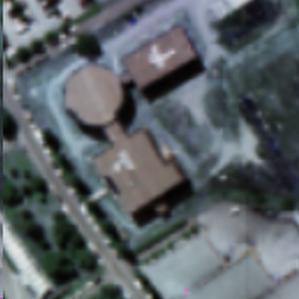} &
    \includegraphics[width=0.08\textwidth]{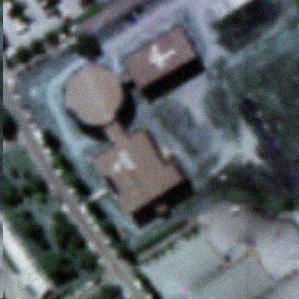} \\
    \bottomrule
    \end{tabular}
}
\label{Table:transformations}
\end{table*}

\begin{figure*}[htbp]
\centering
\subfigure
{
    \begin{minipage}[b]{.12\linewidth}
        \centering
        \includegraphics[height=2.1cm,width=2.1cm]{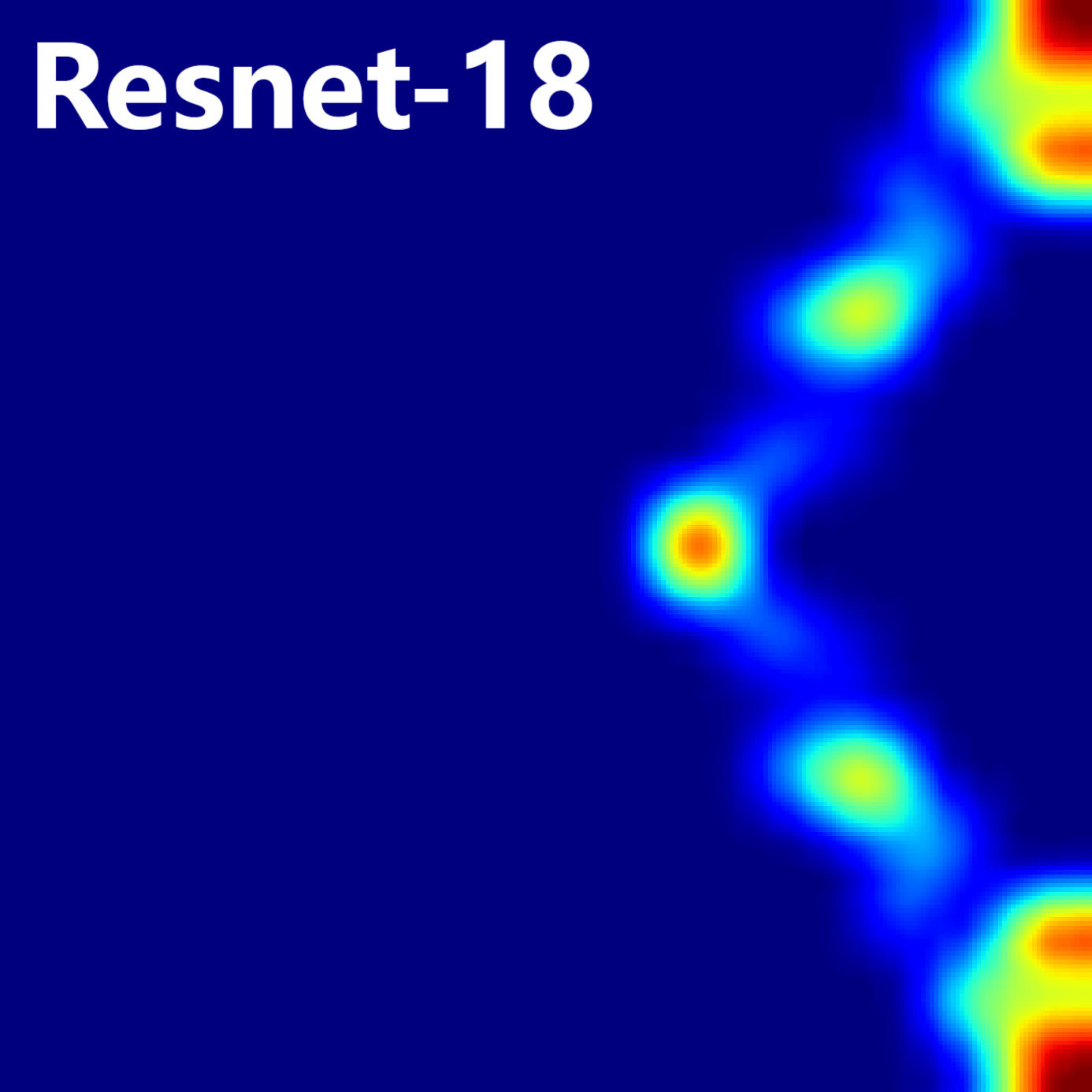}
    \end{minipage}
    
    \begin{minipage}[b]{.12\linewidth}
        \centering
        \includegraphics[height=2.1cm,width=2.1cm]{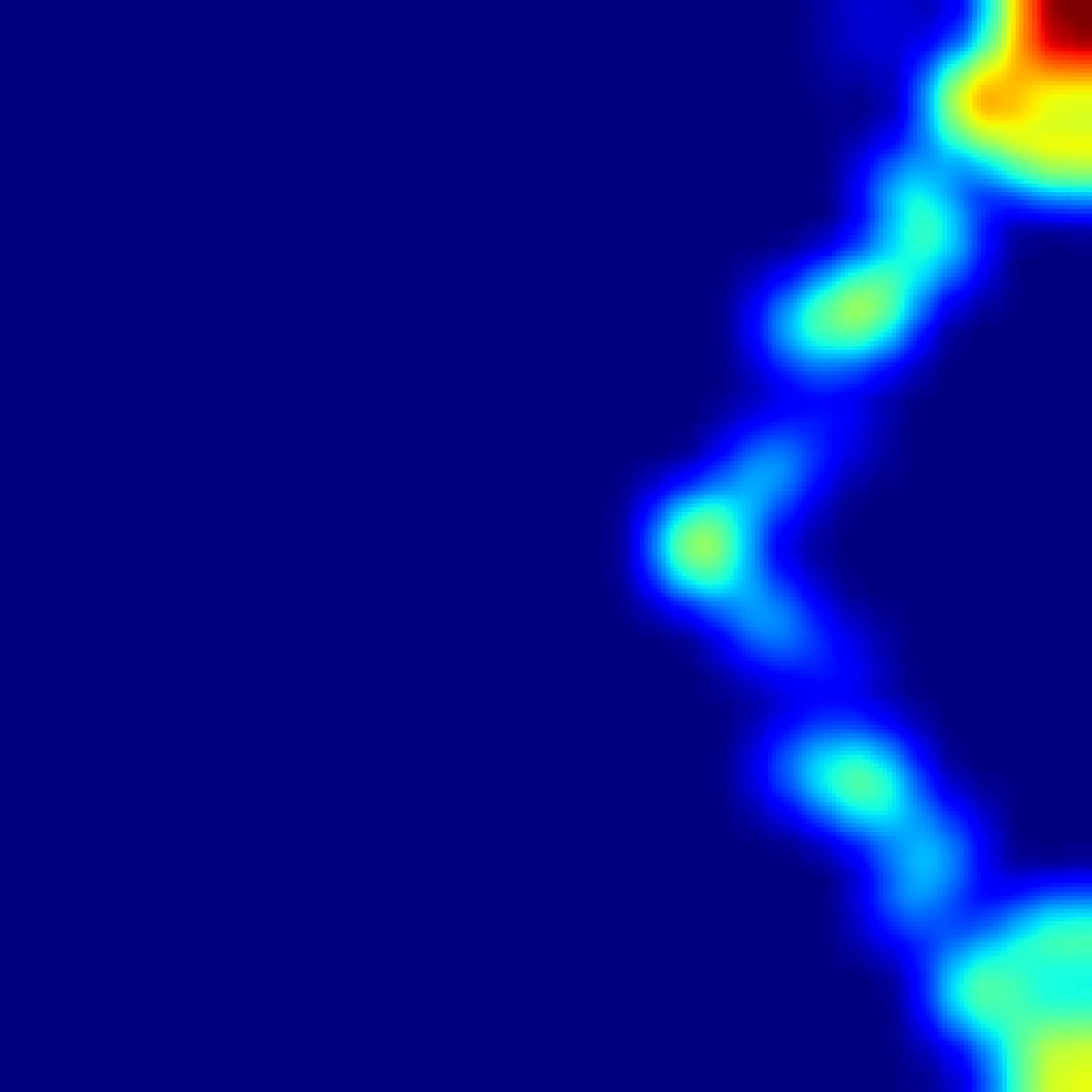}
    \end{minipage}

     \begin{minipage}[b]{.12\linewidth}
        \centering
        \includegraphics[height=2.1cm,width=2.1cm]{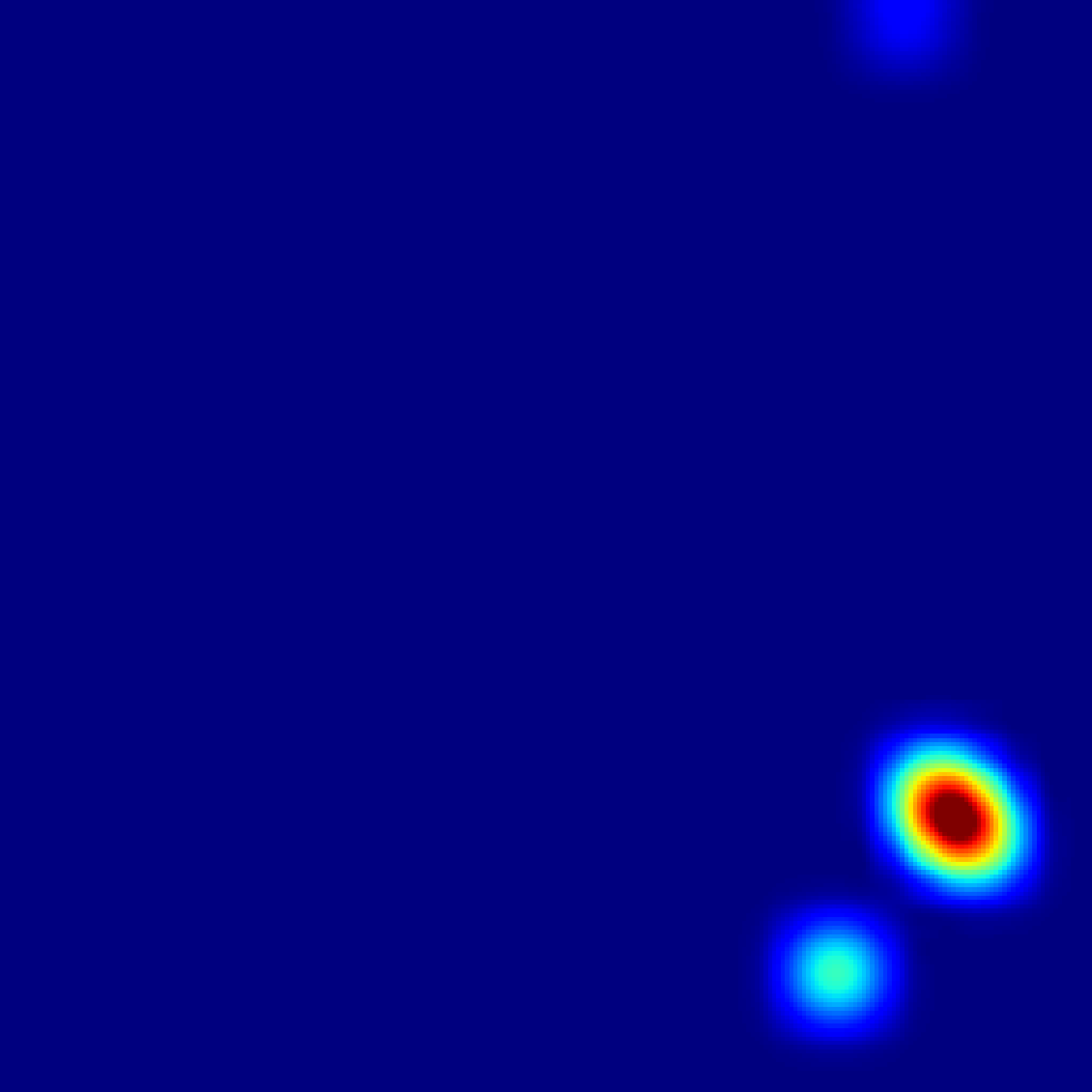}
    \end{minipage}

    \begin{minipage}[b]{.12\linewidth}
        \centering
        \includegraphics[height=2.1cm,width=2.1cm]{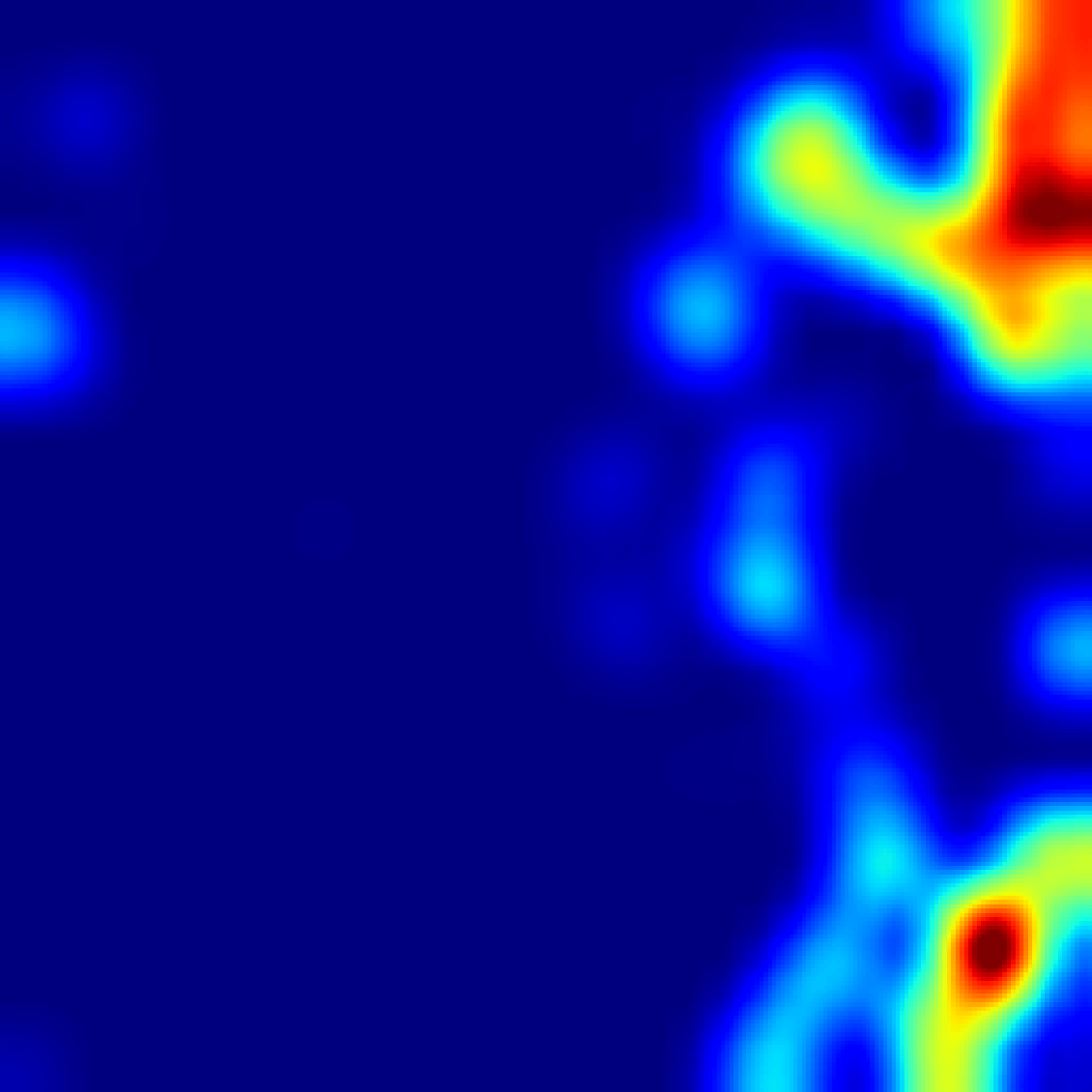}
    \end{minipage}

    \begin{minipage}[b]{.12\linewidth}
        \centering
        \includegraphics[height=2.1cm,width=2.1cm]{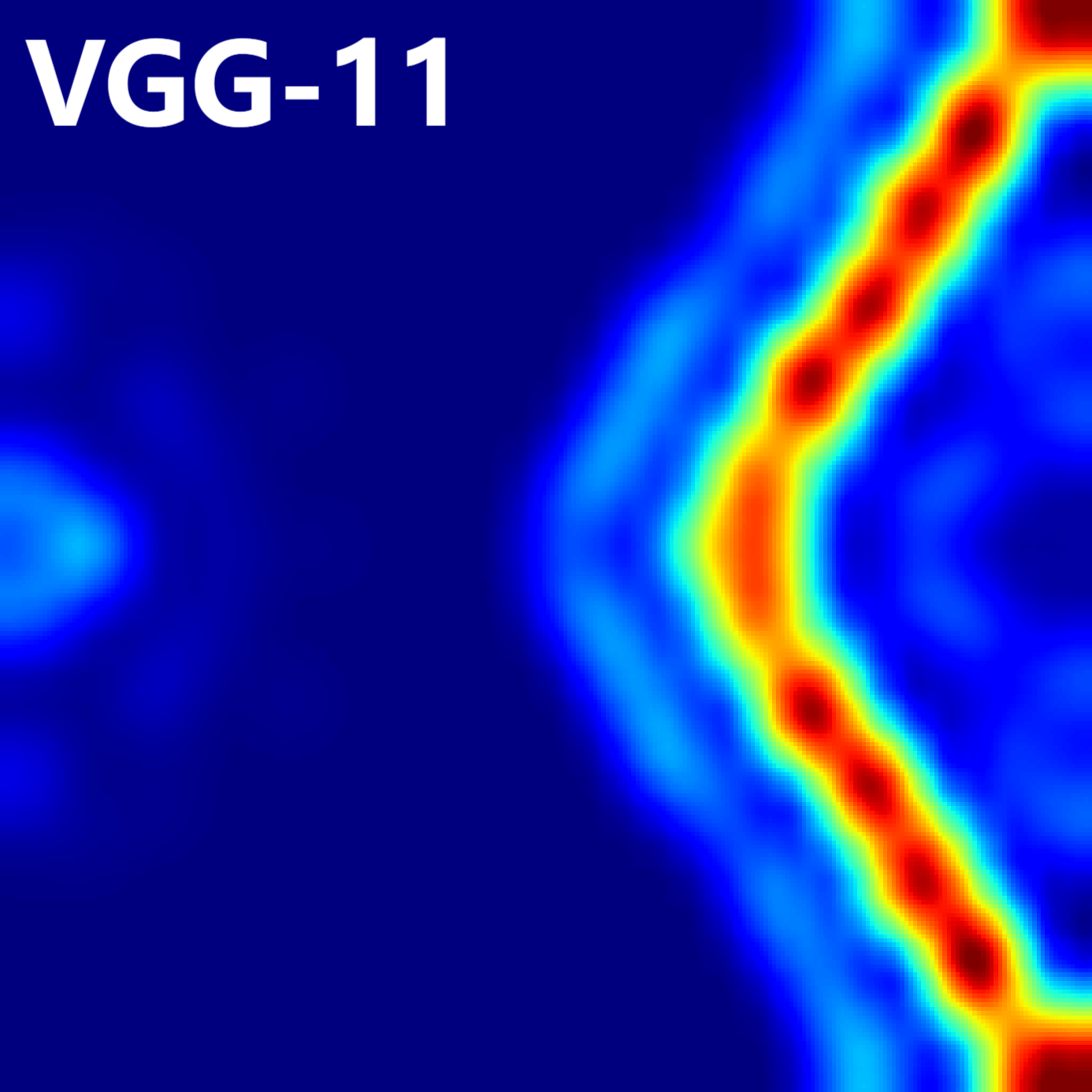}
    \end{minipage}
    
    \begin{minipage}[b]{.12\linewidth}
        \centering
        \includegraphics[height=2.1cm,width=2.1cm]{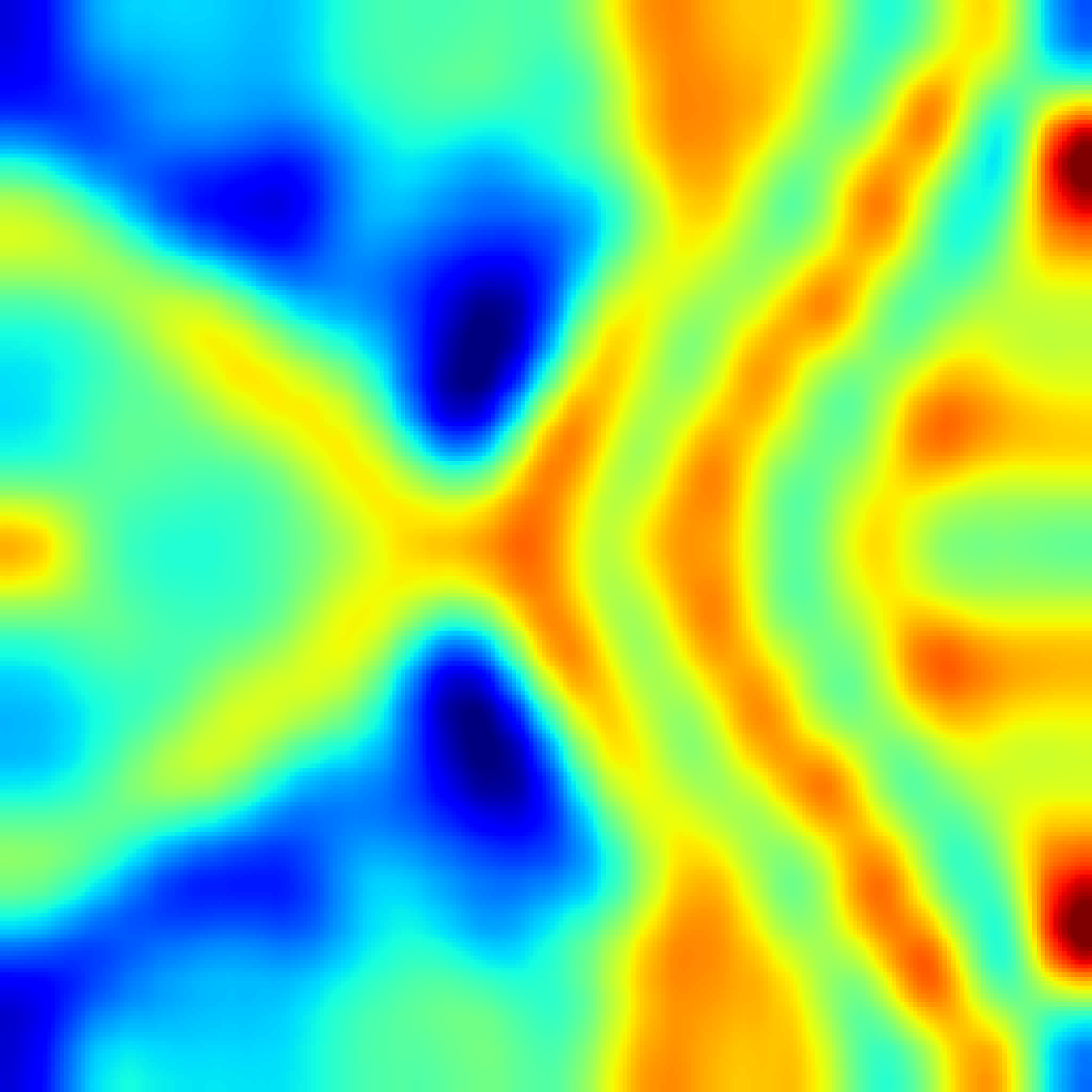}
    \end{minipage}

     \begin{minipage}[b]{.12\linewidth}
        \centering
        \includegraphics[height=2.1cm,width=2.1cm]{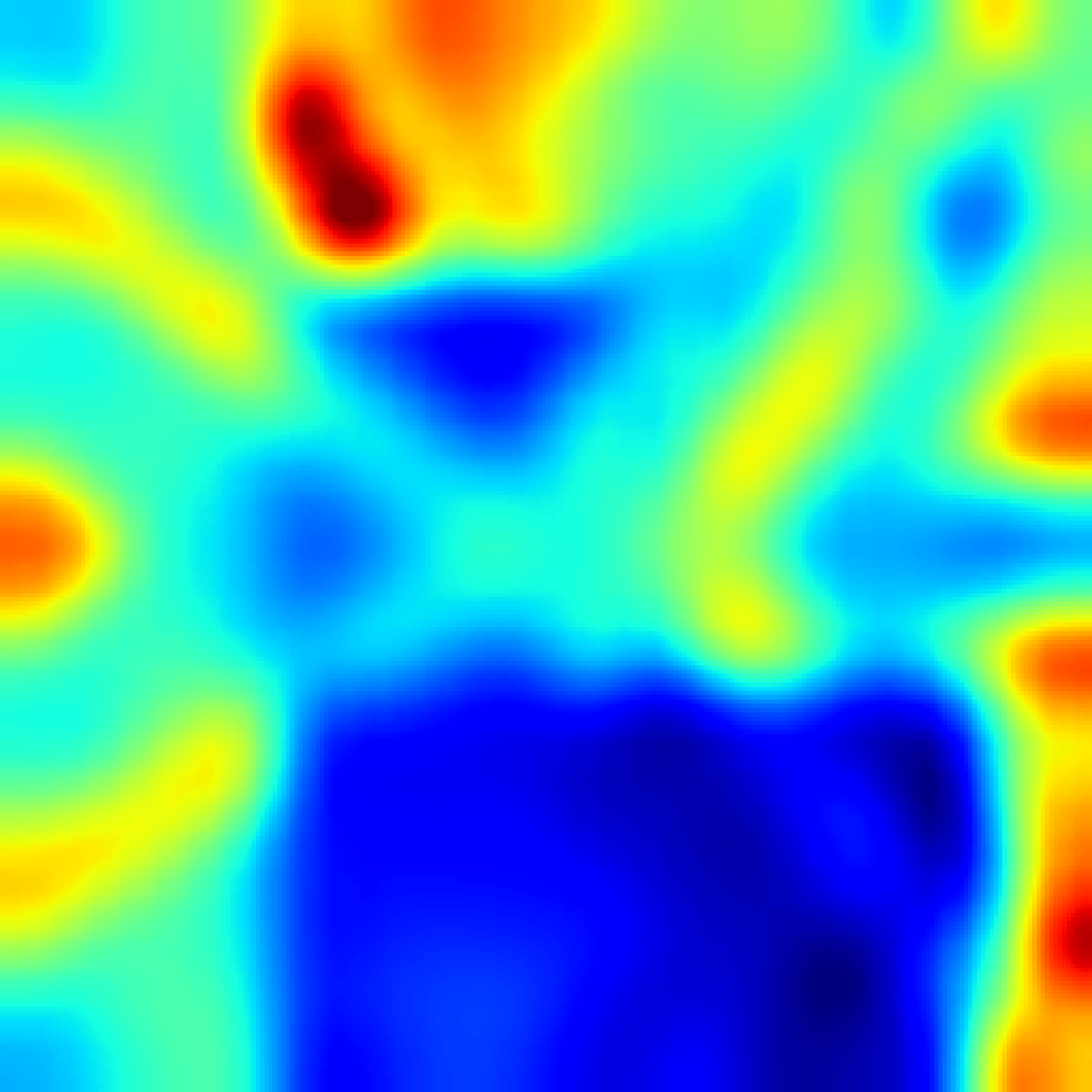}
    \end{minipage}

    \begin{minipage}[b]{.12\linewidth}
        \centering
        \includegraphics[height=2.1cm,width=2.1cm]{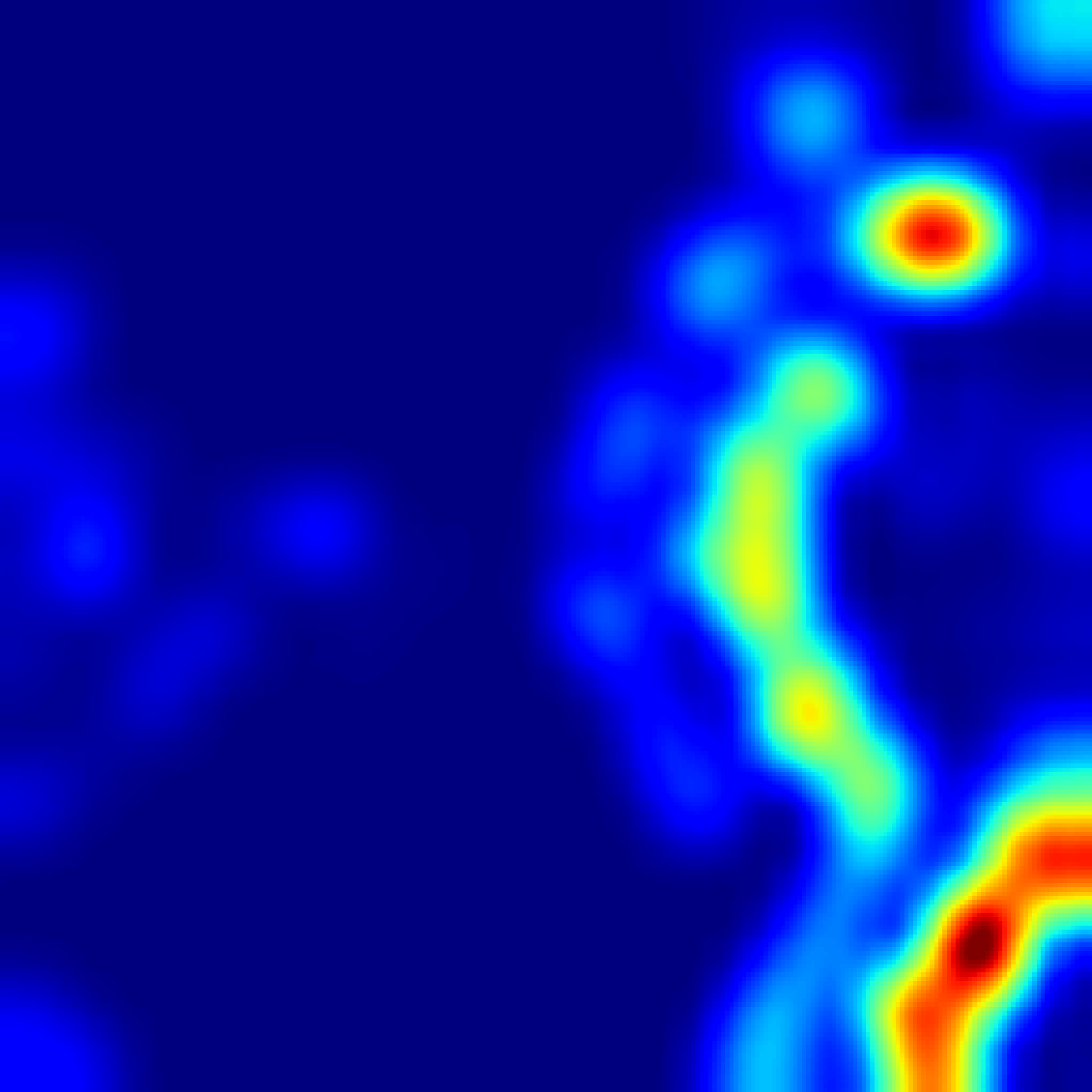}
    \end{minipage}
    
}\\
\subfigure
{
    \begin{minipage}[b]{.12\linewidth}
        \centering
        \includegraphics[height=2.1cm,width=2.1cm]{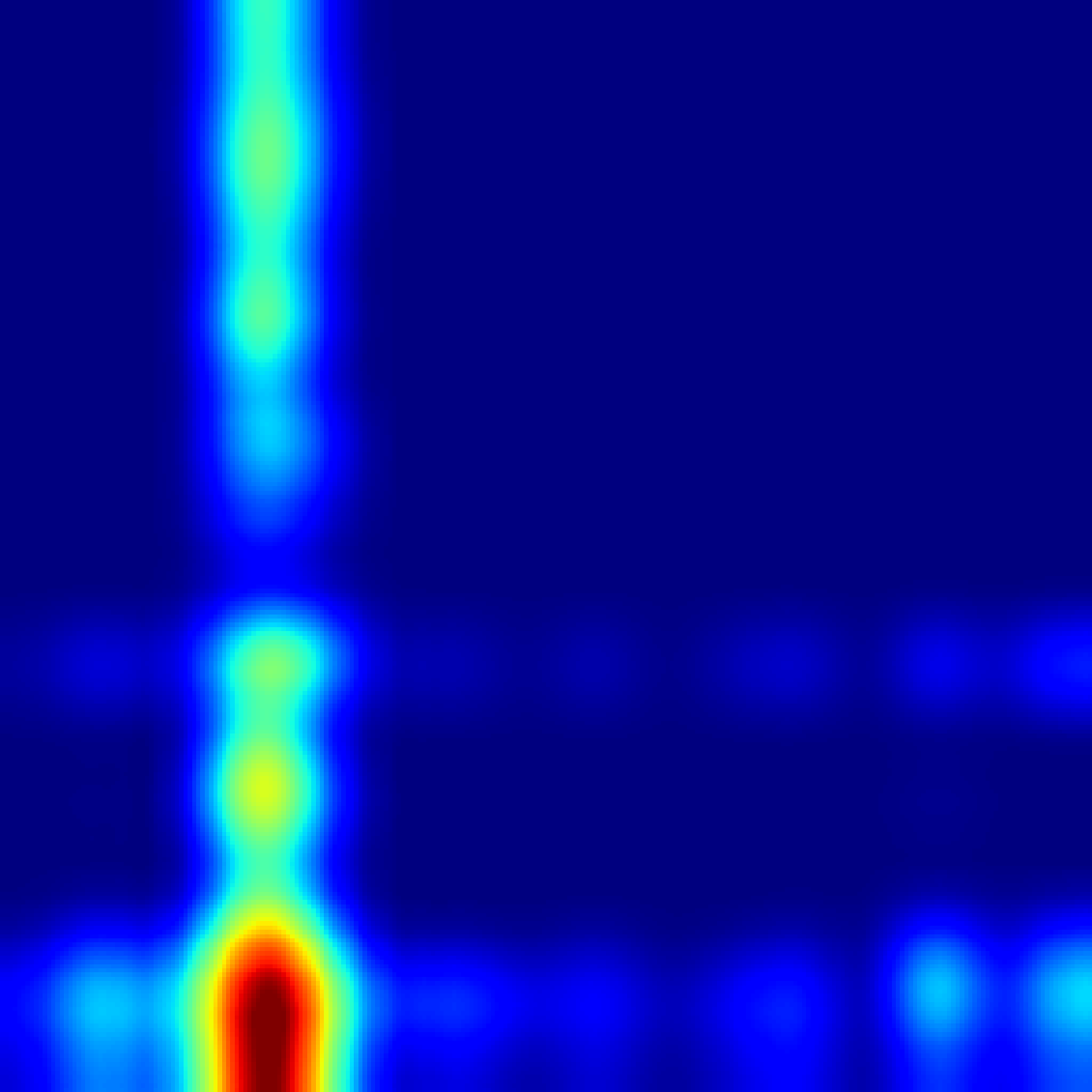}
    \end{minipage}

    \begin{minipage}[b]{.12\linewidth}
        \centering
        \includegraphics[height=2.1cm,width=2.1cm]{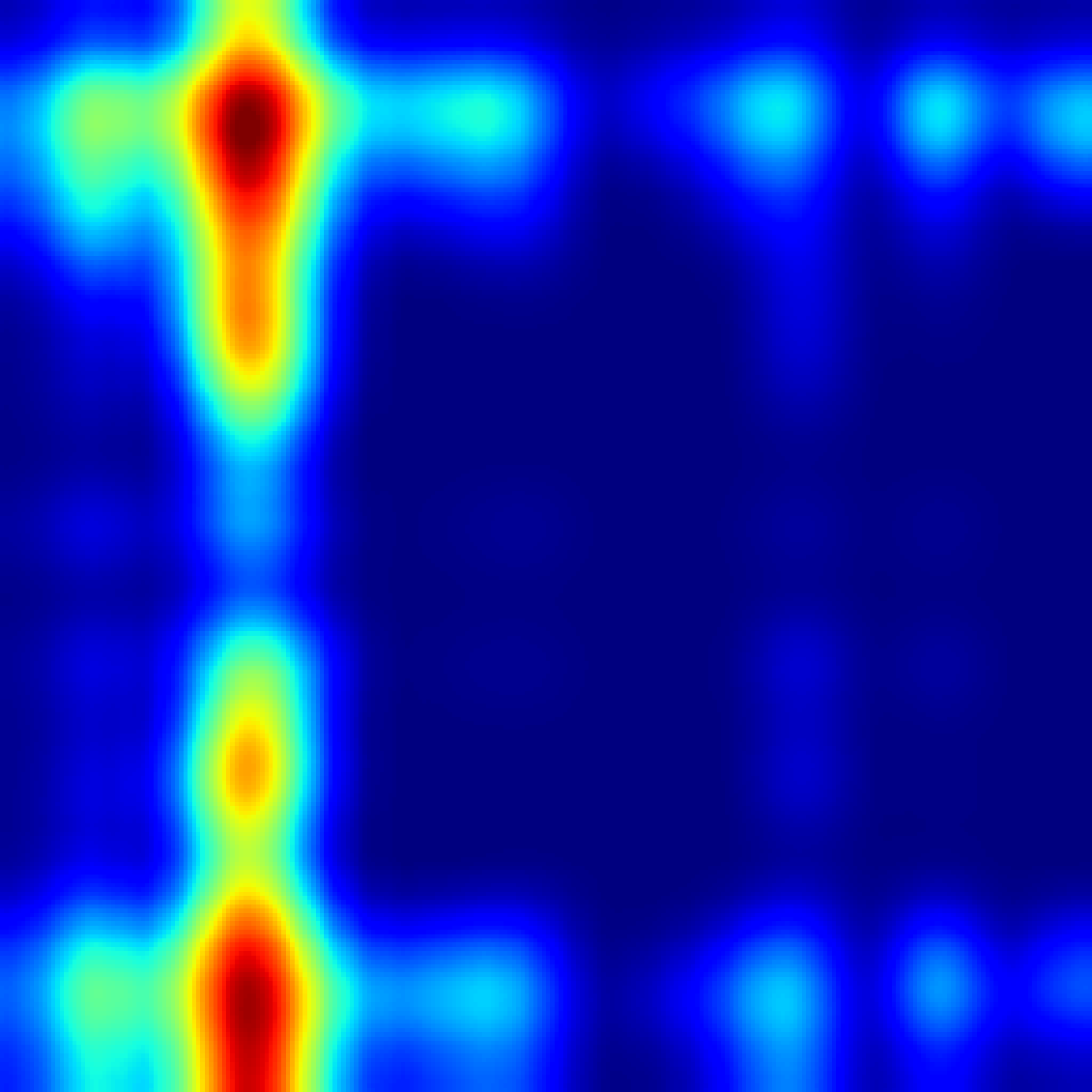}
    \end{minipage}

    \begin{minipage}[b]{.12\linewidth}
        \centering
        \includegraphics[height=2.1cm,width=2.1cm]{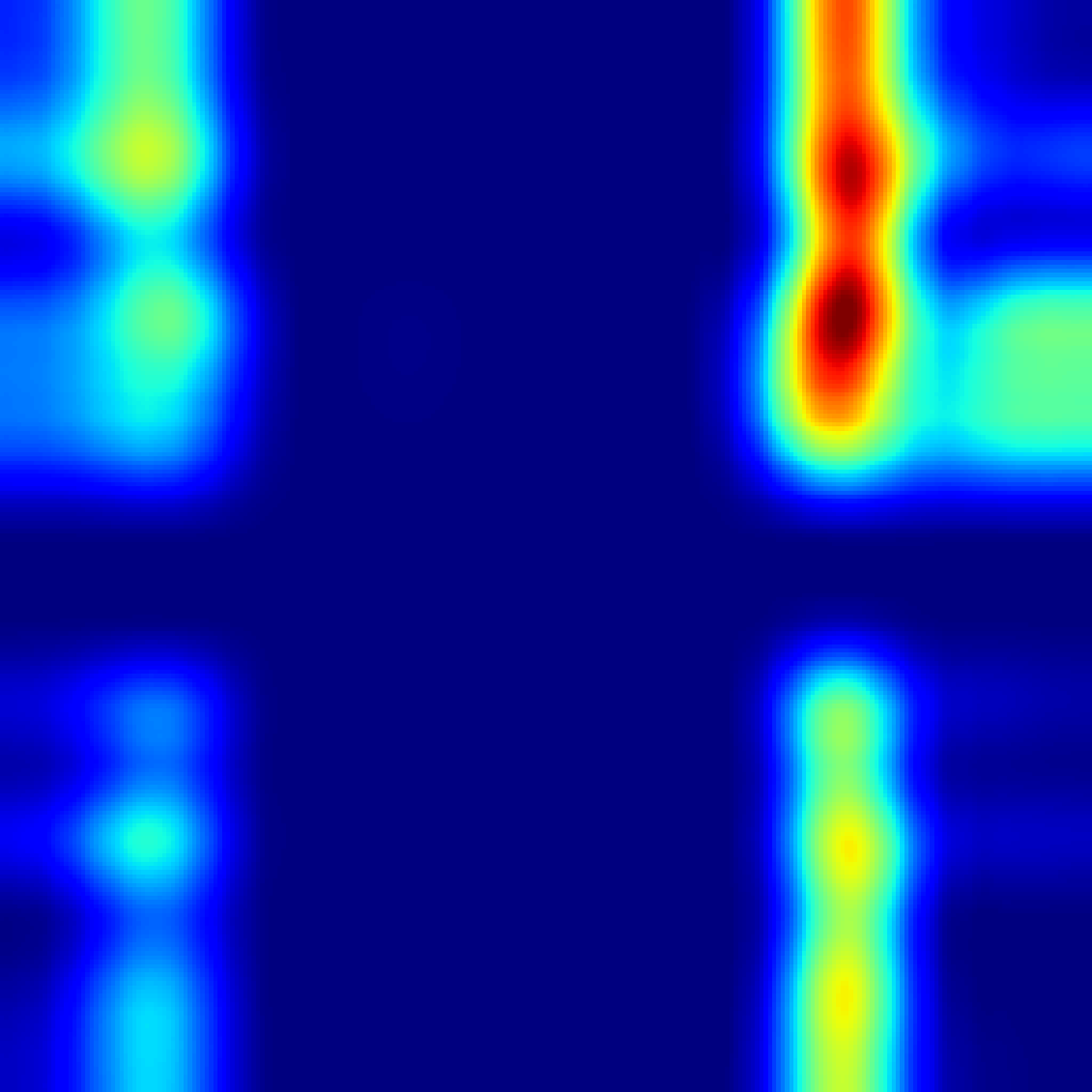}
    \end{minipage}

    \begin{minipage}[b]{.12\linewidth}
        \centering
        \includegraphics[height=2.1cm,width=2.1cm]{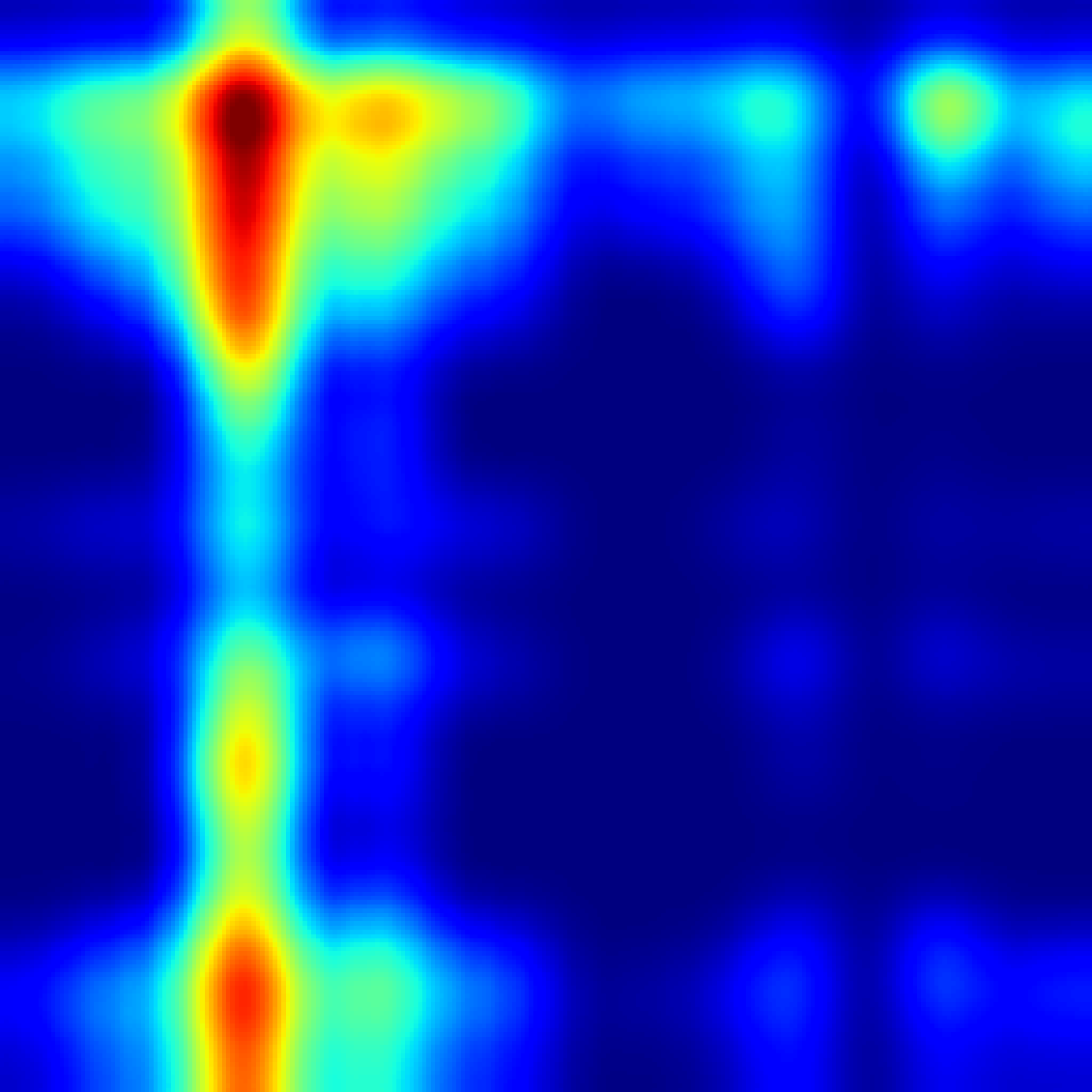}
    \end{minipage}
    
    \begin{minipage}[b]{.12\linewidth}
        \centering
        \includegraphics[height=2.1cm,width=2.1cm]{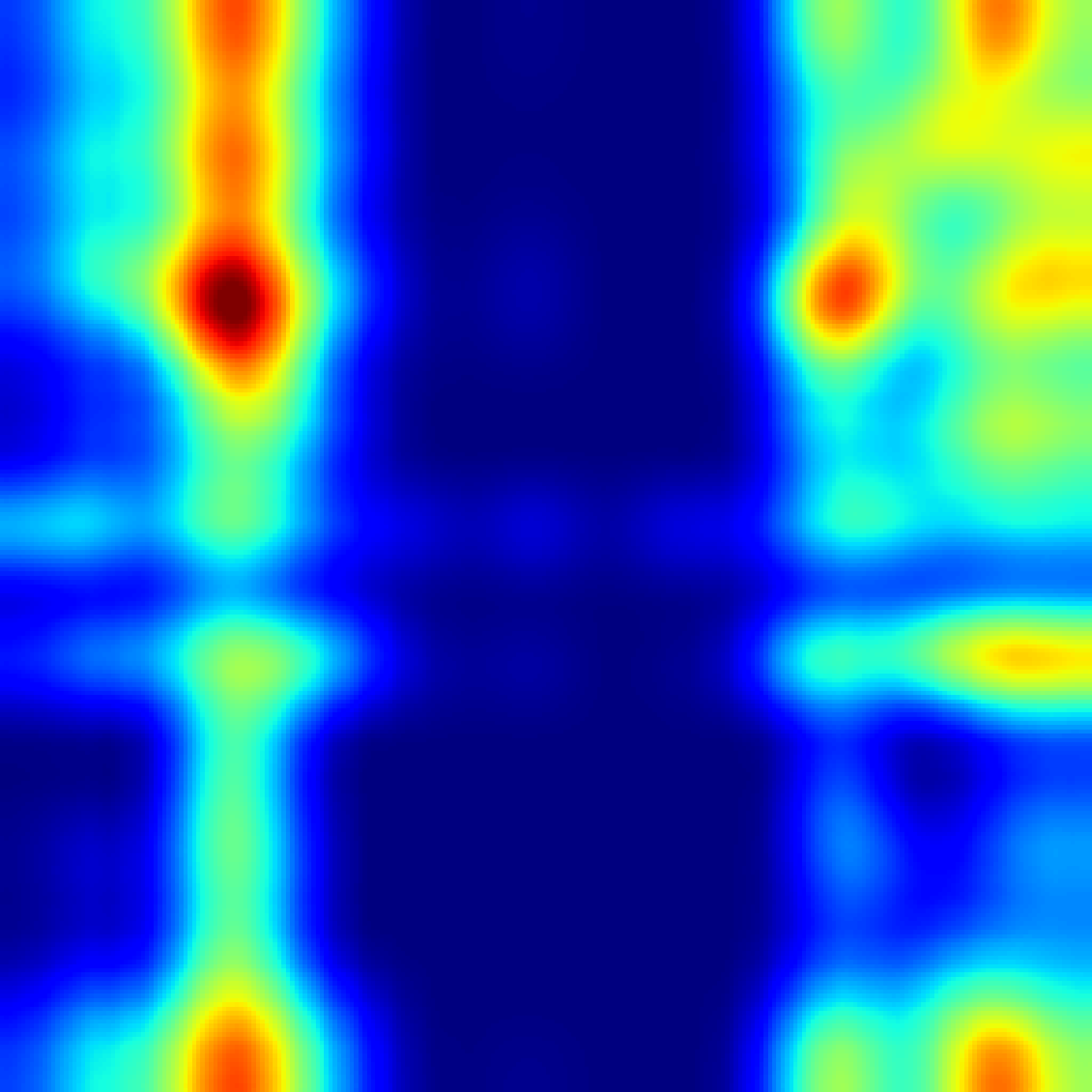}
    \end{minipage}

    \begin{minipage}[b]{.12\linewidth}
        \centering
        \includegraphics[height=2.1cm,width=2.1cm]{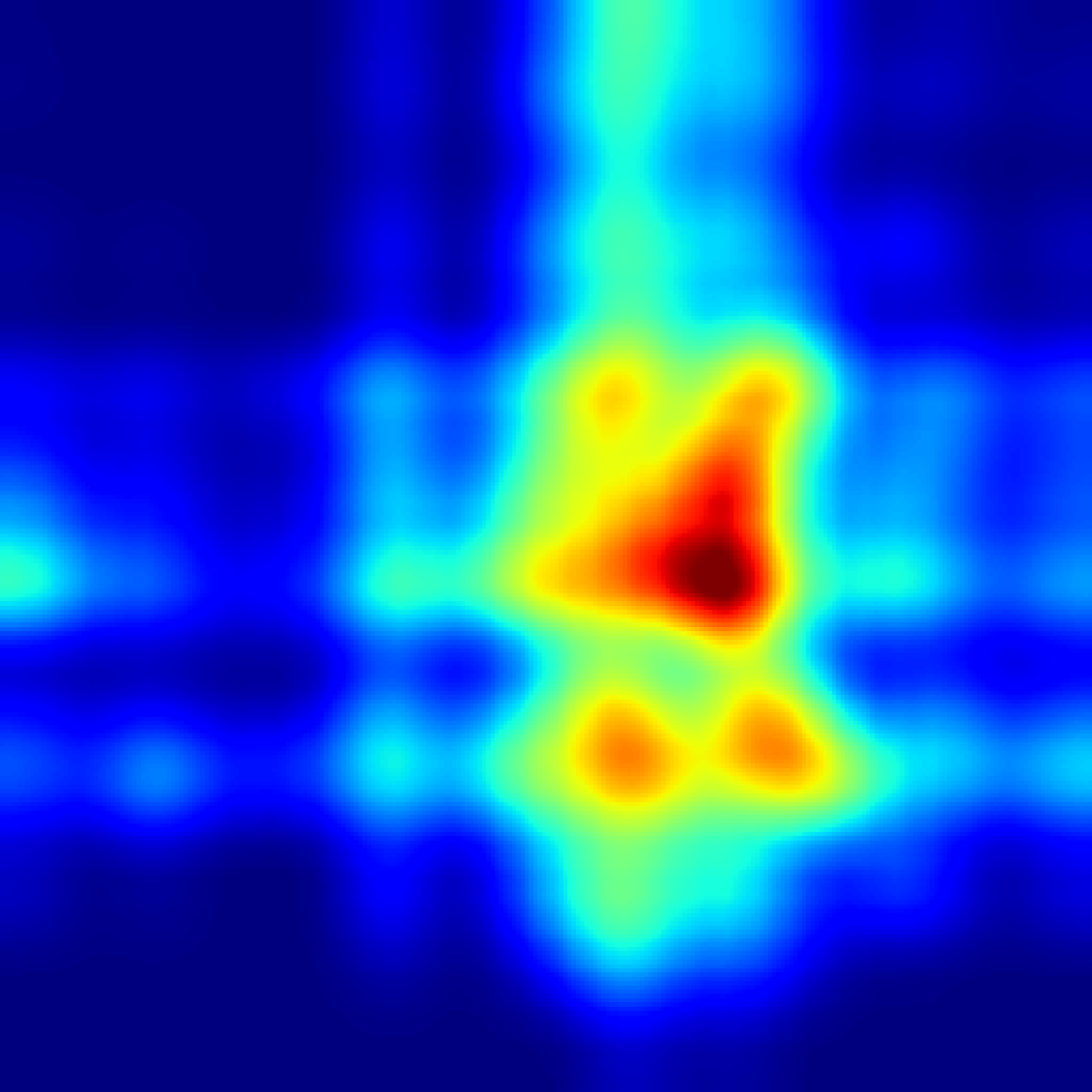}
    \end{minipage}

    \begin{minipage}[b]{.12\linewidth}
        \centering
        \includegraphics[height=2.1cm,width=2.1cm]{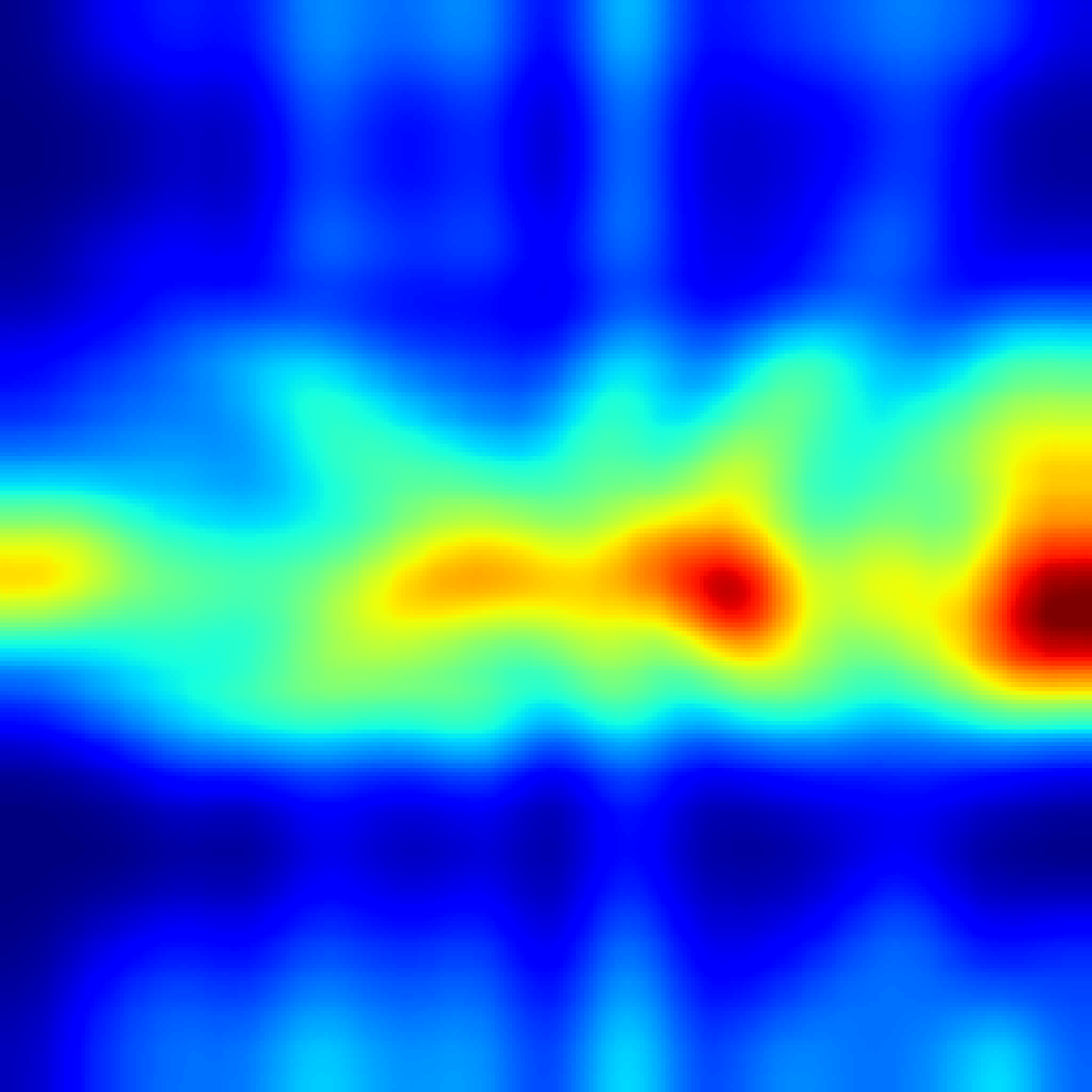}
    \end{minipage}

    \begin{minipage}[b]{.12\linewidth}
        \centering
        \includegraphics[height=2.1cm,width=2.1cm]{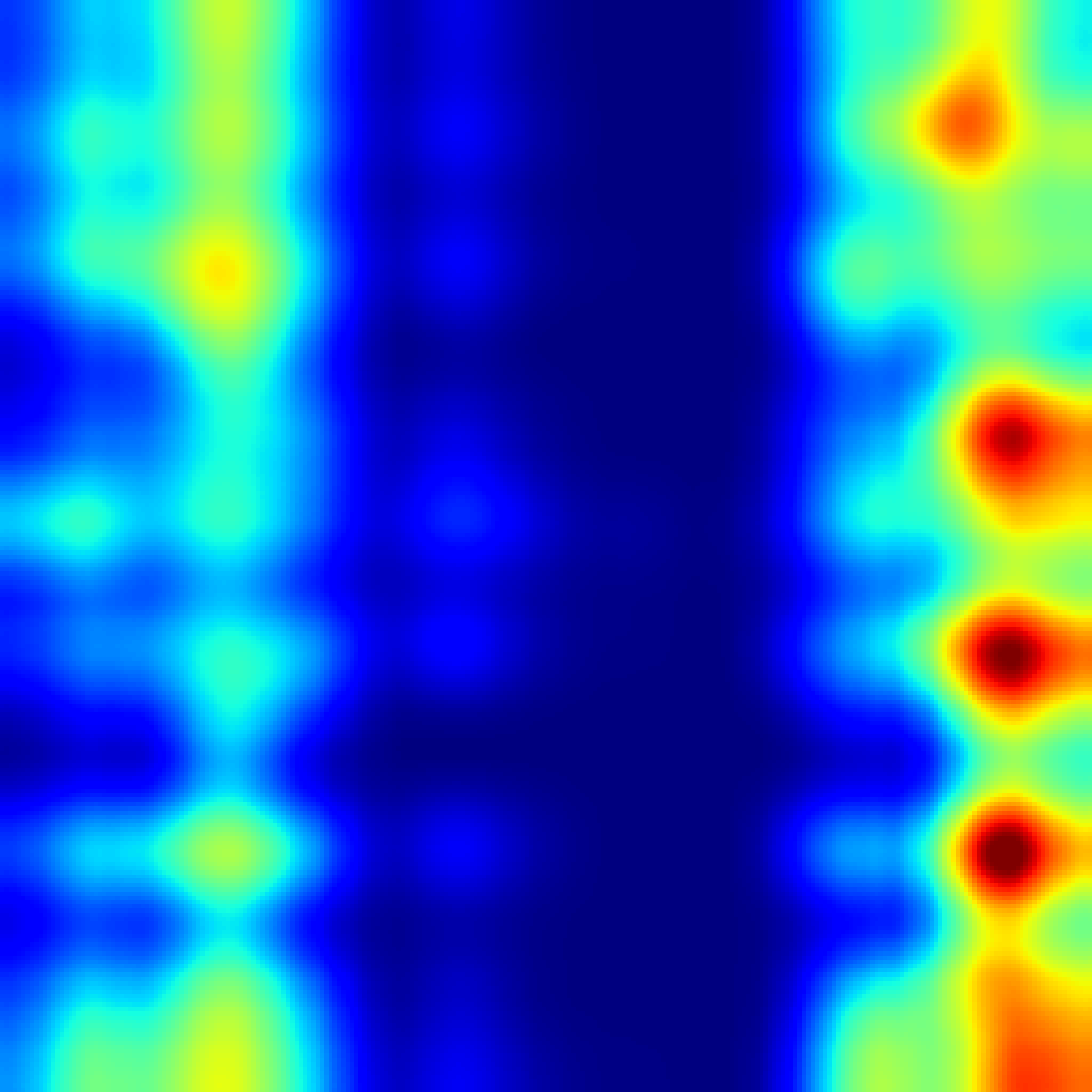}
    \end{minipage}
}
\caption{This figure compares the attention maps of the original example and its transformed examples. The first and second rows represent the spatial and spectral dimensions, respectively. The first four columns are the ResNet-18 model, corresponding to the original example and three transformation examples randomly obtained through our proposed transformation method; the last four columns are the same but from the VGG-11 model.} 
\label{attention map}
\end{figure*}

Fig. \ref{attention map} shows the comparison between the attention heatmaps of the original example and the transformed examples using the proposed 3D structure-invariant strategy. It can be observed that the model has different attention regions for the images before and after transformation. The use of multiple transformed examples enables the model to focus on a broader range of spectral information during the generation of adversarial examples.

\begin{algorithm}[t]
\small
\caption{Ours Adversarial Attack Method}
\label{adversarial_algorithm}

\begin{algorithmic}[0]
\STATE \textbf{Input:} HSI example $x$, substitute model $f$, loss function $L$, transformation $T\left( \right)$, perturbation $\delta$, perturbation budget $\varepsilon$, step size $\alpha$, decay factor $\mu$, momentum term $m_i$, gradients $g_i$, number of copies $N$, iteration steps $I$.
\STATE \textbf{Output:} HSI adversarial example $x^{adv}$.
\end{algorithmic}

\begin{algorithmic}[1]
\STATE  $\delta_0 = \text{zero}(C \times H \times W)$, $x_0^{adv} = x$, $g_0 = 0$, $m_0 = 0$
\FOR {$i = 0 \to I-1$} 
    \STATE Construct a set $X^{adv}_{T}$ of $N$ transformed images using $T()$:
    \STATE \hspace*{2em} $X^{adv}_{T} = T(x + \delta_i)* N$
    \STATE Calculate the channel weights according to \eqref{equ3};
    \STATE Calculate weighted feature divergence loss according to \eqref{equ4};
    \STATE Calculate the cross-entropy classification loss according to \eqref{equ5}; 
    \STATE Then resulting in the total loss:
    \STATE \hspace*{2em} $L = L_F + \eta * L_A$
    \STATE Calculate the average gradient of $N$ copies:
    \STATE \hspace*{2em}$g_i^{ave} = \frac{1}{N} \left(\sum_{j=0}^{N} (\nabla_{\delta} L (x_j^{adv}, x))\right), (x_j^{adv} \in X^{adv}_{T})$
    \STATE Update the momentum:
    \STATE \hspace*{2em}$m_{i+1} = \mu \cdot m_i + \frac{g_i^{ave}}{||g_i^{ave}||_1} $
    \STATE Update perturbations:
    \STATE \hspace*{2em}$\delta_i = {Clip_\varepsilon} \{ \delta_i + \alpha \cdot {sign}(m_{i+1}) \}$
    \STATE Update adversarial examples: $x_{i+1}^{adv} = Clip_{x,\varepsilon} \{ x + \delta_i \}$
\ENDFOR
\STATE \textbf{output} $ x^{adv} = x_I^{adv} $
\end{algorithmic}
\end{algorithm}

\subsection{Weighted Feature Divergence Loss}
Current research on adversarial examples for HSI classification rarely considers the image feature maps extracted by the model. However, these features are transferable between models with different architectures \cite{21}. Relying exclusively on information from the classification-layer is inadequate for effectively disrupting the deep-level features of the examples.

Based on this, we introduce the distance between the intermediate feature maps of the original and adversarial examples as the primary guiding loss, and refer to it as the weighted feature divergence loss. By increasing the difference between the feature maps of the original and adversarial examples, our purpose is to effectively perturb the features within adversarial examples so that they deviate from the original feature representations. As there are several layers in the model to extract different level features, we take the feature map of a certain layer for feature loss calculation.
 
To guide the perturbations to suppress features associated with the true class, it first enlarges the feature values of the original example by a fixed factor, ensuring that the adversarial example's feature values are lower. Next, it computes the variance of each channel within the feature map to assign different weights to different channels, thereby destroying the features that have a greater impact on HSI classification. We detail the key points of the weighted feature divergence loss calculation as follows.

\begin{figure*}[t]
\centering
\subfigure[raw]
{
    \begin{minipage}[b]{.2\linewidth}
        \centering
        \includegraphics[height=3cm,width=3cm]{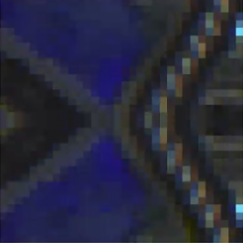}
    \end{minipage}}
\subfigure[channel 10 (0.1213)]
{
    \begin{minipage}[b]{.2\linewidth}
        \centering
        \includegraphics[height=3cm,width=3cm]{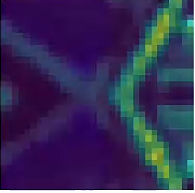}
    \end{minipage}}
\subfigure[channel 28 (0.2106)]
{
     \begin{minipage}[b]{.2\linewidth}
        \centering
        \includegraphics[height=3cm,width=3cm]{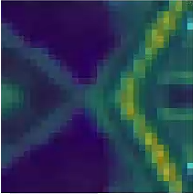}
    \end{minipage}}
\subfigure[channel 50 (0.3260)]
{
    \begin{minipage}[b]{.2\linewidth}
        \centering
        \includegraphics[height=3cm,width=3cm]{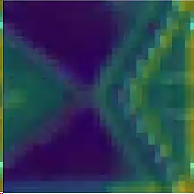}
    \end{minipage}
}
\caption{Feature maps of different channels in the same layer of the VGG-11 model, along with their variances.}
\label{Channel feature map}
\end{figure*}

\subsubsection{Enlarge the Feature Values of Original Examples}
When calculating the feature distance, if we directly increase the difference between the feature maps of the original and adversarial examples, the perturbation during the update process may cause the values in the high-activation regions of the adversarial example's features to either increase or decrease, which will not effectively suppress the features related to the true class. 

To address this issue, we enlarge the original features by a certain factor as shown in Eq.\eqref{equ4}, so that the values in the high-activation regions of the initial adversarial features are significantly lower than those in the original example. In this way, when updating the perturbation based on the weighted feature divergence loss, the values in the high-activation regions of the adversarial example's features related to the true class will be more likely to decrease. 

Although transformation operations are applied to adversarial examples, Ding et al. \cite{55} pointed out that data augmentation operations only alter the spatial positions and areas of different feature components in the feature maps. Thus, neighboring or overlapping regions can still suppress the feature values within adversarial examples. By averaging the gradients of multiple adversarial examples, the perturbation more stably suppresses the features related to the ground-truth label in these adversarial examples, thereby enhancing the effectiveness of the weighted feature divergence strategy.

\subsubsection{Weighted Feature Divergence}
Even features extracted from the same convolutional layer of a model, differences can still be observed between the feature maps of different channels, and the variance of features in each channel also varies, as shown in Fig. \ref{Channel feature map}. When the variance in a specific channel is large, the features extracted by that channel are non-robust features related to the true class \cite{23}, which have a greater impact on the model's classification results.

If we simply averages these channels when computing the distance between features, the differences between the channels are ignored. Therefore, to make the perturbation more effectively disrupt the non-robust features within the adversarial example, we calculated the weights of each channel according to their variance in the feature maps, and give non-robust features a greater influence in the loss. This increases the probability of the adversarial examples deviating from the ground-truth label. 

Let the intermediate layer $l$ of the substitute model $f$ be denoted as $f^l$. $X^l \in {R^{C_l*D_l}}$ represents the feature maps of the original example $x$ at layer $l$, where $C_l$ is the number of channels of the feature maps and $D_l$ is the product of width and height of the feature maps. Let $\sigma _k^{X^l}$ denote the variance of the $k$-th channel in $X^l$, which can be expressed by the following formula:

\begin{equation}
\sigma_k^{X^l} = \frac{1}{D_l}{\sum\limits_{i = 1}^{D_l} {\left({X_{k,i}^l -\mu _k^{X^l}} \right)} ^2}
\label{equ2}
\end{equation}

Here, $\mu _k^{X^l} $is the mean of the $k$-th channel in $X_l$ (i.e., $\mu _k^{X^l} = \frac{1}{D_l}\sum\nolimits_{i = 1}^{{D_l}} {X_{k,i}^l}$), and$X_{k,i}^l$ denotes the $i$-th feature value in the $k$-th channel of $X^l$.
 
Then, the weight of each channel is defined as the proportion of the variance of a single channel to the total variance across all channels, which can be expressed as follows:
\begin{equation}
{W_k} = \frac{\sigma_k^{X^l}} {{\sum\nolimits_{k = 1}^ {C_l} {\sigma _k^{X^l}}}}
\label{equ3}
\end{equation}

When $W_k$ is higher, the model pays more attention to the $k$-th channel.

For each convolutional layer in DNN, convolutional kernels extract various feature maps from the input. There is an overall correlation among features across different channels, such as feature distribution, but Euclidean distance ignores such correlations. In contrast, the Chi-squared distance provides a fair starting point for evaluating the differences between each pair of channels. Let $ A^l \in R ^ {C_l*D_l} $ represents the features of the adversarial example $x_{adv}$ at layer $l$, where $C_l$ is the number of channels and $D_l$ is the product of width and height. With the Chi-squared distance, the weighted feature divergence loss, i.e., the weighted feature maps difference between $x$ and $x_{adv}$, can be expressed as:
\begin{align}
    L_F\left( {x_{adv},x} \right) &= W \cdot d \left(f^l\left(x_{adv}\right),\lambda f^l\left( x \right) \right) \notag &\\
                    & =W \cdot d({A^l},\lambda{X^l}) \label{equ4}&\\
                    & = \sum\limits_{k = 1}^{C_l} {W_k}\frac{M_k}{N_k} \notag &
\end{align}
where
\begin{flalign*}
\left\{
    \begin{aligned}
    W_k &= \frac{\sigma_k^{\lambda X^l}} {{\sum\nolimits_{k = 1}^ {C_l} {\sigma _k^{\lambda X^l}}}}, \notag &\\
    M_k &= \sum_{i=1}^{D_1} (A^l_{k,i} - \lambda  X^l_{k,i})^2, \notag &\\
    N_k &= \sum_{i=1}^{D_1} (|A^l_{k,i}| + |\lambda  X^l_{k,i}| + o).\notag &
    \end{aligned}
    \right.
\end{flalign*}

where $\lambda$ is the scaling factor for enlarging the feature values of the original examples. 

Moreover, due to the pooling layers in DNN models, some information is inevitably lost, and certain intermediate layer feature maps have lower correlation with the true class of the image \cite{54}. This limitation reduces the attack success rate when relying solely on intermediate layer information. To address this, the cross-entropy classification loss which is derived according to the information of the output layer is also adopted, which is defined as:

\begin{equation}
{L_A}\left(x_{adv},x\right) = CE\left(f \left(x_{adv}\right),f(x)\right)
\label{equ5}
\end{equation}

where $CE(\cdot)$ represents the cross-entropy function. Thus, by combining the previous $L_F$ with the cross-entropy classification loss, the total loss of the proposed method is defined as Eq.\eqref{equ6}. 
\begin{equation}
L = {L_F}+ \eta * {L_A}
\label{equ6}
\end{equation}

where $\eta$ is a hyperparameter representing the proportion of the cross-entropy classification loss.


The final optimization formula can be expressed:
\begin{equation}
x_{adv} = \arg \max \left( {L_F} + \eta * {L_A}\right), s.t.\left\| \delta  \right\|_1 \le \varepsilon
\label{equ7}
\end{equation}

We greatly enhanced the diversity of the input samples through the use of 3D structure-invariant transformation techniques. This compels the model's attention to extend from local features to multiple global regions, thereby establishing richer discriminative zones in the feature space. This provides the weighted feature divergence loss with more optimization directions and richer feature information, while the loss itself also reinforces the utilization of the feature space. Together, these two factors strengthen the adversarial transferability.

\begin{figure}[htbp]
\centering
\subfigure[HoustonU 2018]
{
    \begin{minipage}[b]{.9\linewidth}
        \centering     
        \includegraphics[height=1.8cm,width=7.2cm]{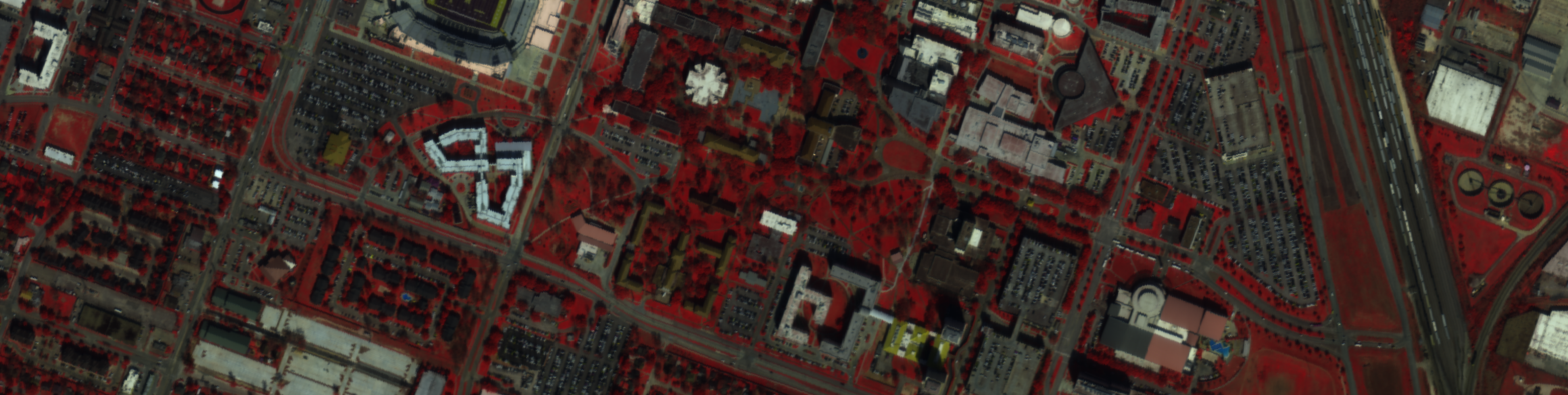} \\
        \includegraphics[height=1.8cm,width=7.2cm]{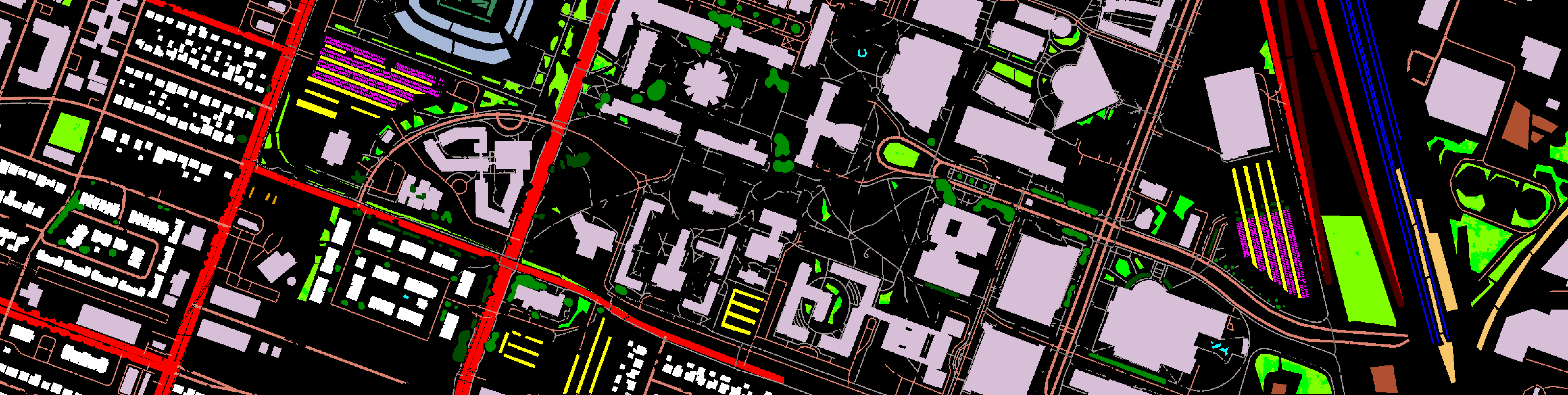} \\
        \includegraphics[height=2.2cm,width=7.1cm]{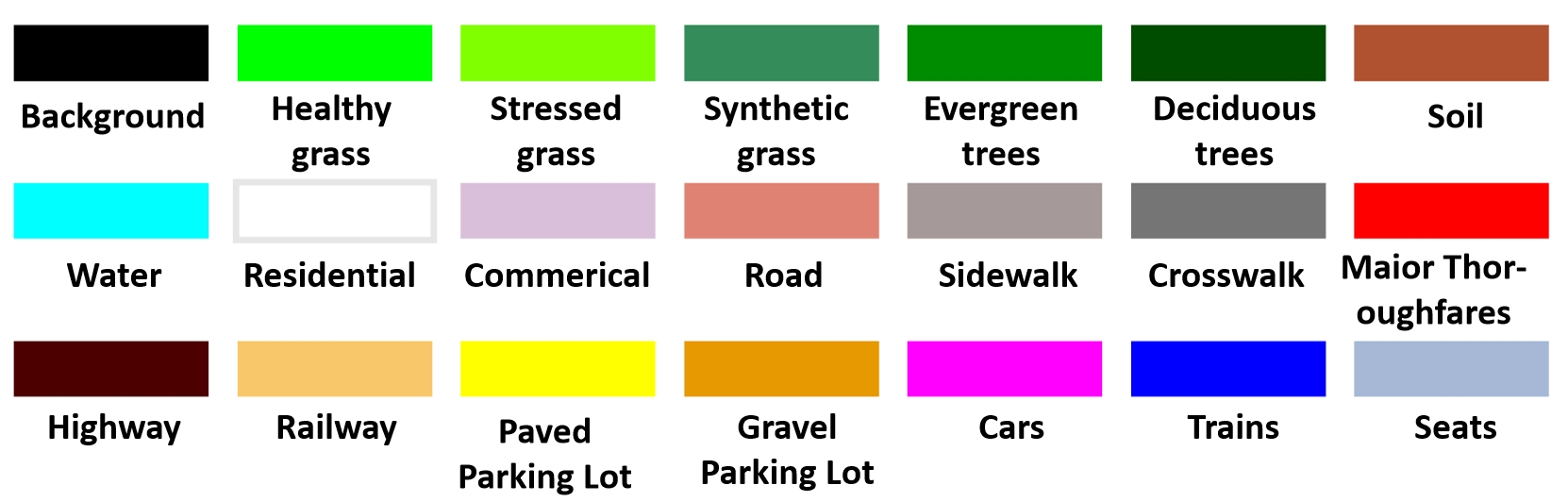}
    \end{minipage}
}
\subfigure[PaviaU]
{
    \begin{minipage}[b]{.25\linewidth}
        \centering
        \includegraphics[height=3cm,width=1.8cm]{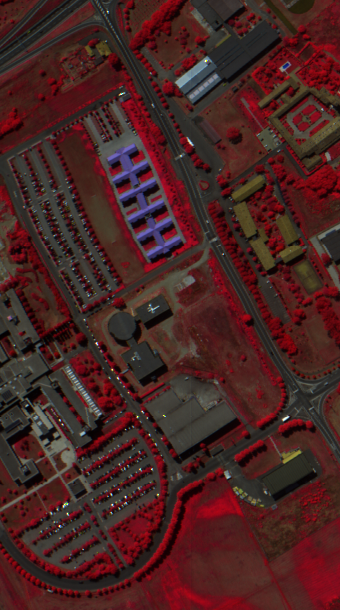}
    \end{minipage}
    \begin{minipage}[b]{.25\linewidth}
        \centering
        \includegraphics[height=3cm,width=1.8cm]{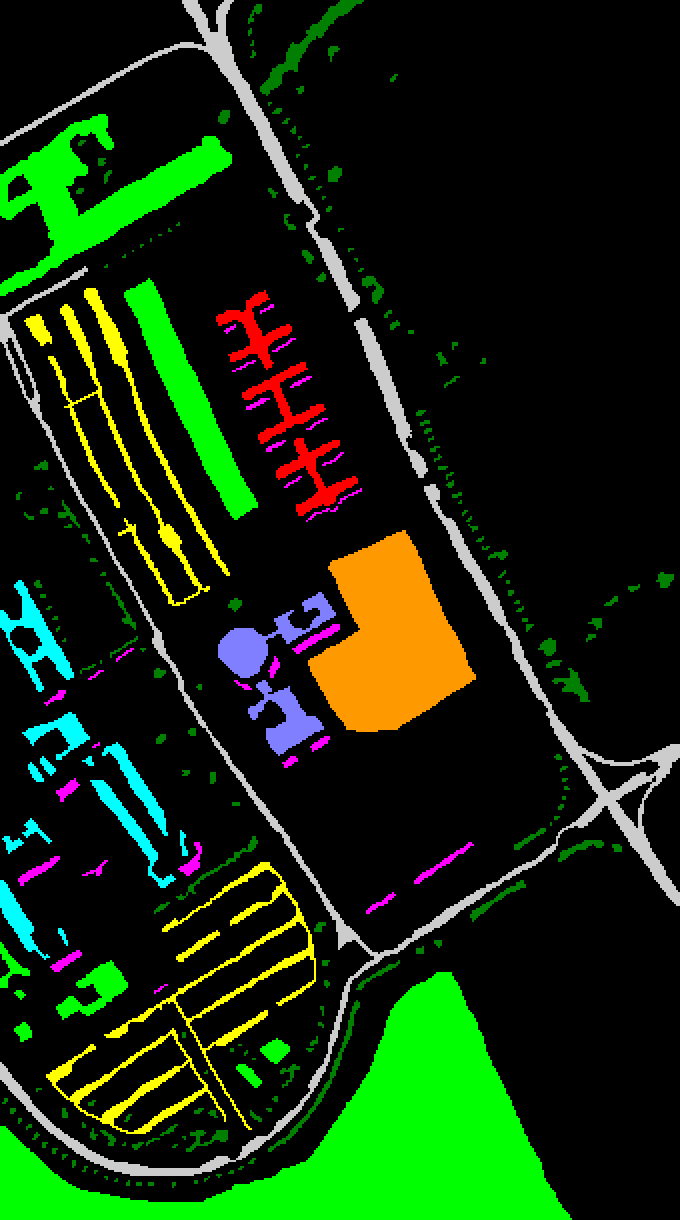}
    \end{minipage}
    \begin{minipage}[b]{.25\linewidth}
        \centering
        \includegraphics[height=3cm,width=2cm]{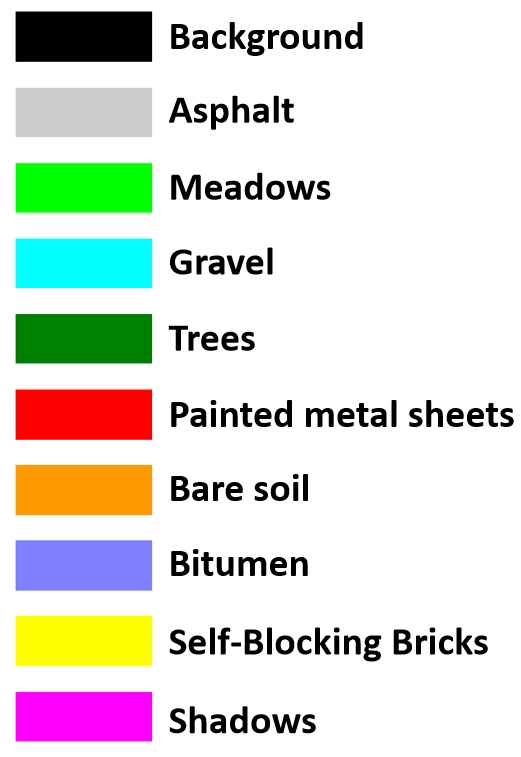}
    \end{minipage}
}
\subfigure[Indian Pines]
{
    \begin{minipage}[b]{.3\linewidth}
        \centering
        \includegraphics[height=2.5cm,width=2.5cm]{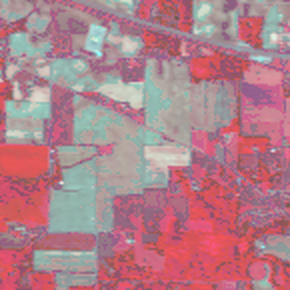}
    \end{minipage}
    \begin{minipage}[b]{.3\linewidth}
        \centering
        \includegraphics[height=2.5cm,width=2.5cm]{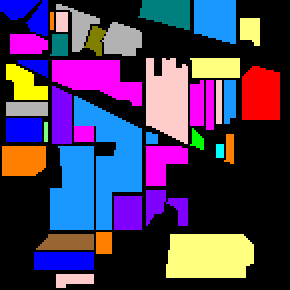}
    \end{minipage}
    \begin{minipage}[b]{.3\linewidth}
        \centering
        \includegraphics[height=3.3cm,width=2.3cm]{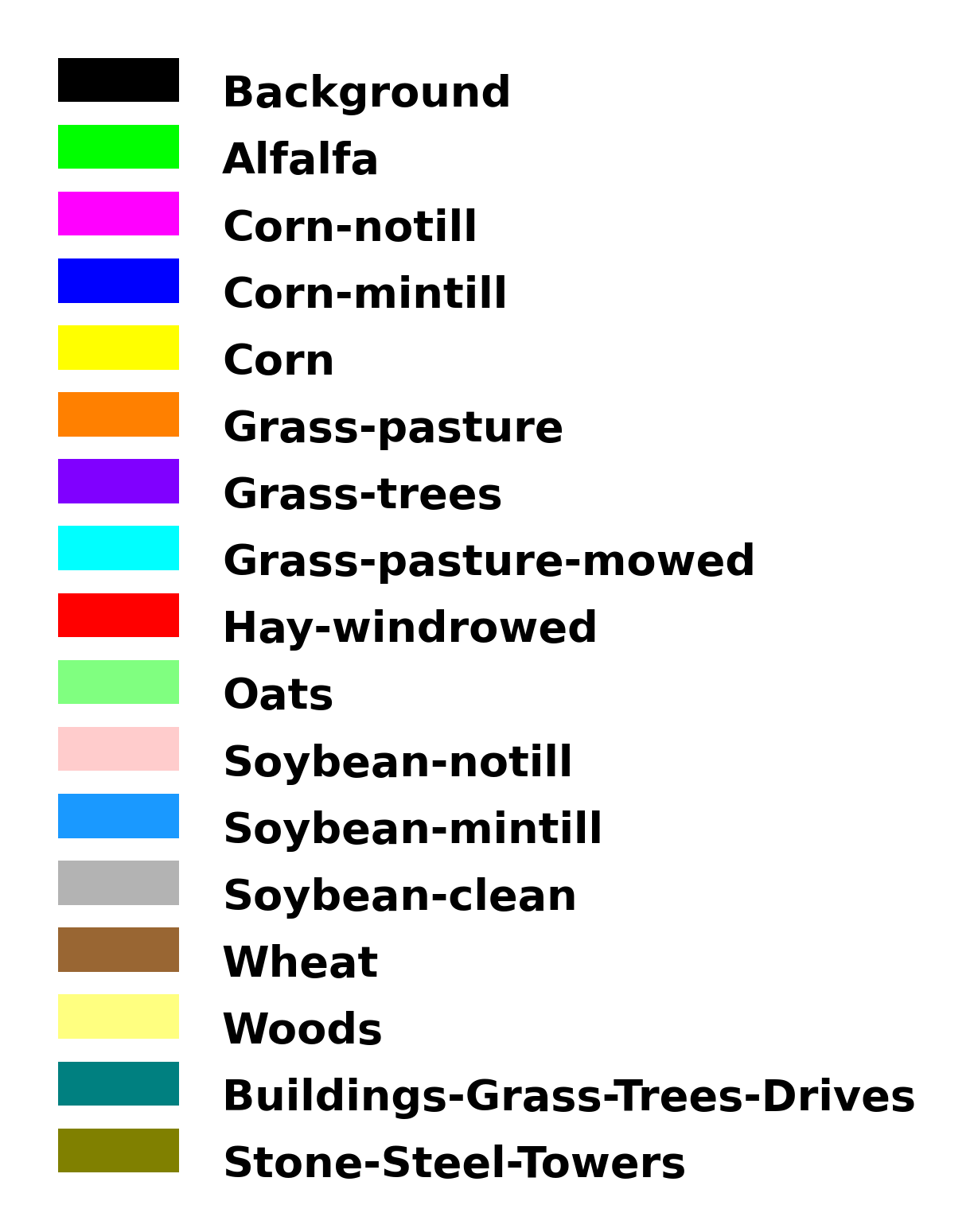}
    \end{minipage}
}
\caption{False color images and ground-truth maps of the (a) HoustonU 2018, (b) PaviaU, and (c) Indian Pines datasets.}
\label{Image dataset}
\end{figure}

\begin{table*}[htbp]
\centering
\caption{Training and testing sets of the HoustonU 2018, PaviaU, and Indian Pines datasets}  
\label{Table dataset}
\scalebox{0.9}{
    \begin{tabular}{cccccc|ccc|cccccc}
    \toprule
    \multicolumn{6}{c|}{\textbf{HoustonU 2018}} & \multicolumn{3}{c|}{\textbf{PaviaU}} &
    \multicolumn{6}{c}{\textbf{Indian Pines}}\\
    \midrule
    Class&Train&Test&Class&Train&Test&Class&Train&Test&Class&Train&Test&Class&Train&Test\\
    \midrule
    1 &5000  &4799  &11  &1000  &24002&1 &900 &5731 & 1 & 20 & 26 & 11 & 1200 & 1255\\
    2 &10000 &22502 &12  &100   &516  &2 &900 &17749 & 2 &700  & 728 & 12 & 300 &293 \\
    3 &500   &184   &13  &10000 &36358&3 &900 &1199 & 3 & 400 & 430 & 13 & 100 &105 \\
    4 &10000 &3588  &14  &5000  &4849 &4 &900 &164  & 4 & 120 & 117 & 14 & 630 &635 \\
    5 &3000  &2048  &15  &5000  &1937 &5 &900 &445  & 5 & 240 & 243 & 15 & 200 &183 \\
    6 &3000  &1516  &16  &10000 &1475 &6 &900 &4129 & 6 & 360 & 370 & 16 & 50 & 43\\
    7 &200   &66    &17  &100   &49   &7 &900 &430  & 7 & 15 &  13 \\
    8 &10000 &29762 &18  &3000  &3578 &8 &900 &2782 & 8 & 230 & 248 \\
    9 &10000 &213684&19  &3000  &2365 &9 &900 &47   & 9  & 10 & 10 \\
    10 &10000&35810 &20  &3000  &3824 &  &    &     & 10 & 500 & 472 \\
    \bottomrule
    \end{tabular}
}
\end{table*}

\section{Experiments}
In this section, we detail the implementation details and experimental results of our method on three public datasets. First, we introduce the dataset information, parameter settings, and comparison methods. Second, we evaluate the transferability of the adversarial examples on black-box models and their robustness against two defense methods. Subsequently, we discuss the impact of multiple parameters on experimental results. Finally, we analyze a set of visualization results.

\subsection{Experimental Setup}

\textit{1) Datasets:} Experiments were conducted on three widely used public datasets: HoustonU 2018, PaviaU, and Indian Pines as shown in Fig. \ref{Image dataset}.

\textit{HoustonU 2018:} This dataset was captured by HSI CASI 1500 sensor in the Houston University area on February 16, 2017. It has a spatial resolution of 1 meter, a spatial size of 601 × 2384 pixels, 48 spectral bands, and 20 available classes of interest. The dataset can be obtained by applying through the official website.

\textit{PaviaU:} This dataset was collected by ROSIS-03 satellite over the Pavia University area. It has a spatial resolution of 1.3 meters, a spatial size of 610 × 304 pixels, and includes 115 spectral bands. Prior to the experiments, 12 noisy spectral bands were removed, leaving 103 usable spectral bands and 9 available classes for classification.

\textit{Indian Pines:} This dataset was collected over the Indianapolis area of Indiana, USA, using the AVIRIS sensor. It has a spatial resolution of 20 meters, a spatial size of 145 × 145 pixels, and includes 200 spectral bands, and 16 available classes of interest.

\textit{2) Training examples:} For each pixel, a neighborhood window with a spatial size of 32 × 32 is applied to construct the HSIs examples. The detailed number of training and testing examples for the three datasets are provided in Table \ref{Table dataset}.

\textit{3) Networks:} We use five pretrained classification models to verify the transferability of the attack: VGG-11, ResNet-18, VGG-19, Inception v3 (Inc-v3), and Inception Resnet V2 (IncRes-v2). Consistent with the work of SS \cite{19}, VGG-11 and ResNet-18 served as substitute models to generate adversarial examples, and other models are used to test transferability.

\textit{4) Attack Methods:} We use FGSM \cite{24} and MI-FGSM \cite{27} as the baseline attack methods, and combine our method with baseline attack methods, namely ours-FGSM and ours-MI-FGSM. Other advanced comparison methods are SS \cite{19}, SIA \cite{20}, and BSR \cite{39}, which are also combined with the baseline methods. We compare our method with these approaches to demonstrate its effectiveness. In addition, we also choose two common defense methods based on image processing, which are random noise addition and spectral smoothing filtering, to test the robustness of the attack.

\begin{table}[htbp]
\centering
\caption{Intermediate layers used for feature maps extraction. "Block" refers to the basic blocks in ResNet-18. "ReLU-k" denotes the $k$-th ReLU layer in VGG-11}  
\label{Table Model layer}
\scalebox{1}{
    \begin{tabular}{c|cccc}
    \toprule
    Model   & Layer1 & Layer2 & Layer3 & Layer4 \\
    \midrule
    ResNet-18 &Block-2 &Block-4 &Block-6 &Block-8\\
    VGG-11    &ReLU-1  &ReLU-2  &ReLU-4  &ReLU-6 \\
    \bottomrule
    \end{tabular}
}
\end{table}

\begin{table*}[htbp]
    \renewcommand\arraystretch{0.6}
    \centering
    \caption{Classification accuracy (\%) of adversarial attacks on five trained networks on the HoustonU 2018 dataset. The adversarial examples are crafted on ResNet-18. The maximum perturbation budget is set to 0.01 and 0.03, respectively}  
    \label{Table HUresnet}
    \scriptsize   
    \begin{tabular}{cccccccccccc}
    
    \toprule
     Attack &   & \multicolumn{2}{c}{ResNet-18} & \multicolumn{2}{c}{VGG-11} & \multicolumn{2}{c}{VGG-19}& \multicolumn{2}{c}{Inc-V3}& \multicolumn{2}{c}{IncRes-V2}\\
    \cline{2-12}
     
      Method &Perturbation & \multirow{2}*{0.01} & \multirow{2}*{0.03}  &\multirow{2}*{0.01} &\multirow{2}*{0.03} &\multirow{2}*{0.01} &\multirow{2}*{0.03} &\multirow{2}*{0.01} &\multirow{2}*{0.03} &\multirow{2}*{0.01} &\multirow{2}*{0.03}\\
       &budget\\

\midrule
    \multirow{3}*{NONE}& OA & 98.42& 98.42 &98.67 &98.67 &98.66 &98.66 &98.61 &98.61&98.63 &98.63 \\
   \multicolumn{1}{c}{} & AA &98.52 &98.52 & 98.35 & 98.35 & 98.61 & 98.61 & 98.62 & 98.62 & 98.81 & 98.81 \\
    \multicolumn{1}{c}{}& Kappa & 97.67 & 97.67 &98.04 &98.04 &98.02 &98.02 &97.94 &97.94 &97.97 &97.97 \\
\midrule[1.4pt]
    \multirow{3}*{FGSM\cite{24}}& OA & 29.70 & 19.19 & 94.71 & 61.09 & 94.16 & 62.60 & 94.50 & 61.05 & 90.99 & 56.36 \\
   \multicolumn{1}{c}{} & AA & 16.24 & 4.83 & 92.91 & 28.62 & 94.27 & 30.46 & 89.69 & 27.03 & 82.39 & 20.54 \\
    \multicolumn{1}{c}{}& Kappa & 8.79 & -0.72 & 92.20 & 37.83 & 91.38 & 41.02 & 91.86 & 35.38 & 86.75 & 33.92 \\
 \midrule
    \multirow{3}*{SS-FGSM\cite{19}}& OA & \textbf{15.90} & 4.20 & 95.24 & 69.58 & 94.06 & 67.85 & 94.98 & 70.72 & 91.33 & 60.96 \\
   \multicolumn{1}{c}{} & AA & \textbf{5.22} & 0.43 & 94.32 & 50.71 & 94.48 & 57.30 & 91.29 & 45.44 & 86.31 & 34.86 \\
    \multicolumn{1}{c}{}& Kappa & -3.10 & -10.90 & 92.98 & 55.38 & 91.25 & 53.10 & 92.58 & 55.89 & 87.28 & 44.72 \\
\midrule
    \multirow{3}*{SIA-FGSM\cite{20}}& OA & 17.60 & \textbf{2.71} & 89.61 & 49.37 & 88.57 & 49.00 & 88.39 & 55.64 & 81.80 & 42.79 \\
   \multicolumn{1}{c}{} & AA & 6.42 & \textbf{0.43} & 86.11 & 27.91 & \textbf{90.27} & 35.36 & 80.49 & 27.79 & 68.27 & 20.00 \\
    \multicolumn{1}{c}{}& Kappa & \textbf{-3.22} & -11.93 & 84.55 & 28.26 & 83.25 & 28.80 & 82.55 & 31.71 & 72.96 & 20.40 \\
\midrule
    \multirow{3}*{BSR-FGSM\cite{39}}& OA & 41.21 & 14.53 & 90.41 & 51.60 & 90.52 & 53.09 & 87.60 & 53.78 & 82.86 & 40.97 \\
   \multicolumn{1}{c}{} & AA & 22.11 & 4.61 & 83.94 & 22.97 & 91.47 & 28.34 & 76.46 & 21.57 & 66.72 & \textbf{15.94} \\
    \multicolumn{1}{c}{}& Kappa & 17.55 & -8.48 & 85.53 & 25.68 & 85.92 & 29.06 & 81.02 & 24.91 & 74.21 & \textbf{13.59} \\
\midrule
    \multirow{3}*{ours-FGSM}& OA & 26.61 & 7.58 & \textbf{88.89} & \textbf{44.67} & \textbf{88.44} & \textbf{44.47} & \textbf{86.40} & \textbf{53.07} & \textbf{80.23} & \textbf{40.11} \\
   \multicolumn{1}{c}{} & AA & 9.30 & 0.70 & \textbf{81.80} & \textbf{21.02} & 90.31 & \textbf{27.83}& \textbf{72.22} & \textbf{20.74}& \textbf{62.49} & 16.85 \\
    \multicolumn{1}{c}{}& Kappa & -0.05 & \textbf{-13.04} & \textbf{83.35} & \textbf{19.67} & \textbf{83.11} & \textbf{21.05} & \textbf{79.17} & \textbf{24.41} &\textbf{70.28} & 14.59 \\
\midrule[1.4pt]
    \multirow{3}*{MI-FGSM\cite{27}}& OA & \textbf{10.60}& \textbf{1.75} & 94.06 & 57.16 & 92.63 & 54.04 & 93.89 & 59.96 & 89.02 & 48.37 \\
   \multicolumn{1}{c}{} & AA & \textbf{3.16} & \textbf{0.56} & 93.04 & 37.11 & 93.40 & 43.57 & 89.93 & 33.58 & 81.88 & 26.67 \\
    \multicolumn{1}{c}{}& Kappa & \textbf{-5.48} & -10.94 & 91.24 & 38.65 & 89.17 & 35.26 & 90.97 & 39.50 & 83.89 & 30.05 \\
\midrule
    \multirow{3}*{SS-MI-FGSM\cite{19}}& OA & 13.15 & 4.50 & 93.34 & 53.35 & 91.82 & 53.42 & 92.89 & 57.84 & 87.77 & 47.09 \\
   \multicolumn{1}{c}{} & AA & 3.96 & 0.57 & 91.39 & 30.77 & 92.76 & 34.16 & 86.63 & 31.83 & 77.15 & 23.41 \\
    \multicolumn{1}{c}{}& Kappa & -4.40 & -10.40 & 90.17 & 31.90 & 87.95 & 32.69 & 89.46 & 34.76 & 82.03 & 25.79 \\
\midrule
    \multirow{3}*{SIA-MI-FGSM\cite{20}}& OA & 15.49 & 2.73 & 85.99 & 37.46 & 84.93 & 35.17 & 85.01 & 42.54 & 77.50 & 32.46 \\
   \multicolumn{1}{c}{} & AA & 5.52 & 0.71 & 79.47 & 15.09 & 87.30 & 16.57 & 73.84 & 16.15 & 61.52 & 12.13 \\
    \multicolumn{1}{c}{}& Kappa & -4.80 & -12.58 & 79.23 & 13.43 & 78.00 & 12.00 & 77.45 & 14.74 & 68.85 & 8.96 \\
\midrule
    \multirow{3}*{BSR-MI-FGSM\cite{39}}& OA & 34.96 & 15.72 & 85.99 & 45.66 & 86.23 & 46.14 & 82.49 & 45.73 & 76.79 & 36.37 \\
    \multicolumn{1}{c}{} & AA & 14.99 & 4.37 & 74.24 & 16.15 & 86.75 & 16.35 & 65.58 & 13.81 & 54.76 & 10.60 \\
    \multicolumn{1}{c}{}& Kappa & 9.47 & -8.37 & 78.81 & 16.27 & 79.54 & 18.26 & 73.01 & 12.50 & 65.05 & 6.52 \\
\midrule
    \multirow{3}*{ours-MI-FGSM}& OA & 25.15 & 9.11 & \textbf{81.73} & \textbf{33.58} & \textbf{82.22} & \textbf{31.93} &\textbf{79.95} & \textbf{41.36} & \textbf{72.17} & \textbf{30.58} \\
   \multicolumn{1}{c}{} & AA & 7.80 & 0.84 & \textbf{70.62} & \textbf{11.32} & \textbf{85.83} & \textbf{11.73} & \textbf{62.74} & \textbf{12.20} & \textbf{51.63} & \textbf{10.10} \\
    \multicolumn{1}{c}{}& Kappa & -2.00 & \textbf{-13.73} & \textbf{72.84} &\textbf{5.04} & \textbf{74.68} &\textbf{4.60}& \textbf{69.30} & \textbf{8.89} & \textbf{58.66} & \textbf{2.81} \\
\bottomrule[1.4pt]
    \end{tabular}
\end{table*}

\textit{5) Parameters:} In our work, two smaller disturbance budgets are selected for experiments, which are 0.01 and 0.03 respectively. Some parameters of the SS \cite{19} method follow the settings of the original paper: the spatial superpixel size is set to four spatial pixels, and the spectral clustering size is set to 20. The number of external iterations $T$=20, $\mu$=1, the parameter $t$ is set to 1. To ensure fairness with other methods, the main parameters of our method are established as follows: the spatial division parameter is set to 3, consistent with SIA \cite{20}, and the spectral division parameter is also set to 3. The iteration steps is set to 20, the number of transformed example copies is 10 and the cross-entropy classification loss weight coefficient $\eta$ is set to 0.03. For the substitute model, one layer is selected for feature maps extraction from shallow to deep layers (1/2/3/4, as shown in Table \ref{Table Model layer}). In our experiments, both models selected layer 3. The original feature enlargement factor $\lambda$ is set to 1.2, the momentum factor for MI is set to 1, the step size is 2/255, and the batch size is 100. The experiments are carried out with 13th Gen Intel Core i5-13600KF CPUs with 32 GB of RAM, and one NVIDIA GeForce RTX 4090 GPU.

\begin{table*}[htbp]
\renewcommand\arraystretch{0.6}
\centering
\caption{Classification accuracy (\%) of adversarial attacks on five trained networks on the HoustonU 2018 dataset. The adversarial examples are crafted on VGG-11. The maximum perturbation budget is set to 0.01 and 0.03, respectively}  
    \label{Table HUVGG-11}
    \scriptsize
    \begin{tabular}{cccccccccccc}
    \toprule
     Attack &   & \multicolumn{2}{c}{VGG-11} & \multicolumn{2}{c}{ResNet-18} & \multicolumn{2}{c}{VGG-19}& \multicolumn{2}{c}{Inc-V3}& \multicolumn{2}{c}{IncRes-V2}\\
    \cline{2-12}
    
      Method &Perturbation & \multirow{2}*{0.01} & \multirow{2}*{0.03}  &\multirow{2}*{0.01} &\multirow{2}*{0.03} &\multirow{2}*{0.01} &\multirow{2}*{0.03} &\multirow{2}*{0.01} &\multirow{2}*{0.03} &\multirow{2}*{0.01} &\multirow{2}*{0.03}\\
       &budget\\
\midrule
    \multirow{3}*{NONE}& OA &98.67 &98.67 & 98.42& 98.42 &98.66 &98.66 &98.61 &98.61&98.63 &98.63 \\
   \multicolumn{1}{c}{} & AA & 98.35 & 98.35 &98.52 &98.52  & 98.61 & 98.61 & 98.62 & 98.62 & 98.81 & 98.81 \\
    \multicolumn{1}{c}{}& Kappa &98.04 &98.04 & 97.67 & 97.67 &98.02 &98.02 &97.94 &97.94 &97.97 &97.97 \\
\midrule[1.4pt]
    \multirow{3}*{FGSM\cite{24}}& OA & 66.60 & 59.61 & 92.25 & 69.17 & 88.71 & 66.96 & 92.57 & 67.54 & 91.07 & 64.73 \\
   \multicolumn{1}{c}{} & AA & 35.66 & 16.57 & 88.68 & 39.13 & 85.41 & 30.00 & 86.18 & 29.50 & 81.14 & 24.14 \\
    \multicolumn{1}{c}{}& Kappa & 52.25 & 39.71 & 88.71 & 54.55 & 83.39 & 49.37 & 89.06 & 48.83 & 86.91 & 45.97 \\
\midrule
    \multirow{3}*{SS-FGSM\cite{19}}& OA & 53.90 & 39.83 & 92.85 & 68.72 & 88.22 & 60.55 & 92.43 & 69.96 & 90.48 & 67.30 \\
   \multicolumn{1}{c}{} & AA & 21.30 & 9.57 & 82.27 & 41.70 & 85.88 & 36.20 & 81.33 & 40.96 & 73.92 & 36.21 \\
    \multicolumn{1}{c}{}& Kappa & 33.76 & 15.96 & 89.44 & 53.26 & 82.52 & 41.34 & 88.69 & 54.03 & 85.77 & 51.18 \\
\midrule
    \multirow{3}*{SIA-FGSM\cite{20}}& OA & \textbf{28.04} & \textbf{6.70} & 80.48 & 40.50 & \textbf{70.34} & 33.51 & 75.41 & 46.29 & 70.48 & 39.51 \\
   \multicolumn{1}{c}{} & AA & \textbf{10.30} & \textbf{0.70} & 63.98 & 21.11 & \textbf{60.78} & 16.88 & 54.22 & 21.36 & 49.54 & 15.12 \\
    \multicolumn{1}{c}{}& Kappa & \textbf{2.60} & -16.08 & 71.23 & 14.67 & \textbf{56.93} & 6.65 & 62.87 & 18.85 & 56.32 & 11.81 \\
\midrule
    \multirow{3}*{BSR-FGSM\cite{39}}& OA & 55.16 & 26.72 & 82.64 & 48.57 & 79.03 & 49.92 & 75.80 & 50.50 & 71.29 & 43.68 \\
    \multicolumn{1}{c}{} & AA & 36.47 & 6.18 & 62.82 & 25.09 & 68.71 & 25.16 & 52.20 & 22.62 & 46.96 & 16.54 \\
    \multicolumn{1}{c}{}& Kappa & 33.66 & -6.08 & 73.71 & 22.07 & 68.63 & 22.52 & 62.41 & 21.86 & 56.16 & 13.24 \\
\midrule
    \multirow{3}*{ours-FGSM}& OA & 40.25 & 15.20 & \textbf{79.94} & \textbf{31.27} & 73.44 & \textbf{26.71} & \textbf{72.51} & \textbf{37.82}& \textbf{68.43} & \textbf{31.82}\\
   \multicolumn{1}{c}{} & AA & 19.32 & 1.40 & \textbf{61.78} & \textbf{12.86} & 64.47 & \textbf{9.62} & \textbf{50.83} & \textbf{13.35} & \textbf{44.34} & \textbf{9.93} \\
    \multicolumn{1}{c}{}& Kappa & 9.86 & \textbf{-20.08} &\textbf{69.76} & \textbf{-4.37} & 60.62 & \textbf{-10.52} &\textbf{56.90} & \textbf{-0.14 }& \textbf{51.35} & \textbf{-5.80} \\
\midrule[1.4pt]
    \multirow{3}*{MI-FGSM\cite{27}}& OA & 51.96 & 48.22 & 91.40 & 63.98 & 84.86 & 56.20 & 90.96 & 64.42 & 88.79 & 62.69 \\
   \multicolumn{1}{c}{} & AA & 16.15 & 10.43 & 80.02 & 35.41 & 80.92 & 25.55 & 78.50 & 31.52 & 71.25 & 27.21 \\
    \multicolumn{1}{c}{}& Kappa & 30.36 & 25.16 & 87.32 & 46.86 & 77.48 & 35.71 & 86.50 & 46.78 & 83.27 & 45.18 \\
\midrule
    \multirow{3}*{SS-MI-FGSM\cite{19}}& OA & 52.92 & 44.61 & 90.02 & 61.79 & 82.91 & 55.76 & 89.35 & 63.97 & 87.31 & 61.58 \\
   \multicolumn{1}{c}{} & AA & 19.23 & 10.67 & 76.15 & 31.68 & 76.35 & 25.26 & 74.12 & 32.38 & 69.36 & 28.17 \\
    \multicolumn{1}{c}{}& Kappa & 32.09 & 21.97 & 85.29 & 43.50 & 74.57 & 35.15 & 84.07 & 45.82 & 81.09 & 43.05 \\
\midrule
    \multirow{3}*{SIA-MI-FGSM\cite{20}}& OA & \textbf{22.83} & \textbf{6.60}& 76.03 & 24.30 & \textbf{64.66} & \textbf{19.92} & 71.91 & \textbf{26.67} & 66.09 & \textbf{24.31} \\
   \multicolumn{1}{c}{} & AA & \textbf{8.33 }& \textbf{0.68} & 59.83 & 11.16 & 53.93 & 7.38 & 50.80 & \textbf{9.64} & 46.04 & 6.71 \\
    \multicolumn{1}{c}{}& Kappa & \textbf{-1.72} & -14.37 & 65.14 & -0.61 & 49.19 & -3.99 & 58.12 & 0.68 & 50.53 & -1.82 \\
\midrule
    \multirow{3}*{BSR-MI-FGSM\cite{39}}& OA & 46.05 & 23.97 & 76.95 & 33.27 & 72.74 & 36.16 & 70.94 & 34.45 & 64.82 & 30.70 \\
    \multicolumn{1}{c}{} & AA & 26.83 & 4.21 & 56.39 & 13.27 & 57.97 & 13.13 & 46.40 & 11.29 & 41.43 & 7.39 \\
    \multicolumn{1}{c}{}& Kappa & 21.53 & -7.40 & 65.48 & 3.75 & 59.31 & 7.53 & 55.38 & 4.27 & 47.31 & -0.36 \\
\midrule
    \multirow{3}*{ours-MI-FGSM}& OA & 32.04 & 16.21 & \textbf{72.21} & \textbf{23.54} & 64.83 & 21.88 & \textbf{66.17} & 28.55 & \textbf{60.77} & 25.29 \\
   \multicolumn{1}{c}{} & AA & 11.80 & 1.49 & \textbf{54.01} & \textbf{9.30} & \textbf{52.61} & \textbf{6.57} & \textbf{43.54} & 9.64 & \textbf{38.67} & \textbf{6.67} \\
    \multicolumn{1}{c}{}& Kappa & -1.09 &\textbf{-18.60} & \textbf{58.48} & \textbf{-10.45} & \textbf{48.11} & \textbf{-13.94} & \textbf{47.49} & \textbf{-8.09} & \textbf{40.57} & \textbf{-10.99} \\
\bottomrule[1.4pt]
    \end{tabular}
\end{table*}

\begin{table*}[htbp]
\renewcommand\arraystretch{0.6}
\centering
\caption{Classification accuracy (\%) of adversarial attacks on five trained networks on the PaviaU dataset. The adversarial examples are crafted on ResNet-18. The maximum perturbation budget is set to 0.01 and 0.03, respectively}  
    \label{Table PU resnet18}
    \scriptsize  
    \begin{tabular}{cccccccccccc}
    
    \toprule
    Attack &   & \multicolumn{2}{c}{ResNet-18} & \multicolumn{2}{c}{VGG-11} & \multicolumn{2}{c}{VGG-19}& \multicolumn{2}{c}{Inc-V3}& \multicolumn{2}{c}{IncRes-V2}\\
    \cline{2-12}

      Method &Perturbation & \multirow{2}*{0.01} & \multirow{2}*{0.03}  &\multirow{2}*{0.01} &\multirow{2}*{0.03} &\multirow{2}*{0.01} &\multirow{2}*{0.03} &\multirow{2}*{0.01} &\multirow{2}*{0.03} &\multirow{2}*{0.01} &\multirow{2}*{0.03}\\
       &budget\\
\midrule
    \multirow{3}*{NONE}& OA &99.96 &99.96 &99.85 &99.85 &99.95 &99.95 &99.81 &99.81 &99.94 &99.94 \\
    \multicolumn{1}{c}{} & AA &99.97 &99.97 &99.88 &99.88 &99.96 &99.96 &99.81 &99.81 &99.92 &99.92  \\
    \multicolumn{1}{c}{}& Kappa &99.94 &99.94 &99.84 &99.84 &99.92 &99.92 &99.80 &99.80 &99.91 &99.91 \\
\midrule[1.4pt]
    \multirow{3}*{FGSM\cite{24}}& OA & 39.11 & 12.49 & 78.77 & 35.22 & 80.91 & 56.91 & 81.86 & 53.48 & 95.74 & 62.70 \\
   \multicolumn{1}{c}{} & AA & 58.24 & 31.82 & 89.91 & 51.05 & 93.88 & 64.57 & 94.86 & 64.56 & 98.83 & 82.32 \\
    \multicolumn{1}{c}{}& Kappa & 27.87 & 1.14 & 70.97 & 20.34 & 74.25 & 44.18 & 75.73 & 42.67 & 93.92 & 53.69 \\
\midrule
    \multirow{3}*{SS-FGSM\cite{19}}& OA & 20.67 & 2.18 & 51.83 & 18.74 & 62.30 & 33.55 & 75.92 & 48.05 & 89.06 & 52.22 \\
   \multicolumn{1}{c}{} & AA & 43.09 & 15.62 & 82.37 & 43.70 & 84.41 & 55.66 & 93.70 & 62.61 & 97.51 & 76.86 \\
    \multicolumn{1}{c}{}& Kappa & 6.39 & \textbf{-14.26} & 40.51 & 3.96 & 52.91 & 19.74 & 68.93 & 36.93& 84.96 & 43.12 \\
\midrule
    \multirow{3}*{SIA-FGSM\cite{20}}& OA & \textbf{16.92} &\textbf{0.72} & 44.02 & 15.91 & 57.20 & 35.87 & 73.08 & 43.05 & \textbf{88.27} & 48.24 \\
   \multicolumn{1}{c}{} & AA & \textbf{39.40} & \textbf{7.05} & 75.34 & 42.25 & 84.63 & 62.72 & 91.17 & 60.98 & \textbf{97.23} & 73.37 \\
    \multicolumn{1}{c}{}& Kappa & \textbf{1.09} & -11.98 & 30.87 & 1.96 & 46.23 & 24.24 & 65.13 & 31.60 & \textbf{83.93} & 38.30 \\
\midrule
    \multirow{3}*{BSR-FGSM\cite{39}} & OA & 26.34 & 4.00 & 42.83 & 13.87 & 59.79 & 35.52 & 74.44 & 43.84 & 88.67 & 47.08 \\ 
    \multicolumn{1}{c}{} & AA & 50.98 & 21.49 & 74.12 & 38.12 & 87.06 & 63.96 & 92.91 & 62.37 & 97.40 & 74.78 \\ 
    \multicolumn{1}{c}{} & Kappa & 12.49 & -7.16 & 29.78 & 0.24 & 49.96 & 24.73 & 67.11 & 32.56 & 84.45 & 37.52 \\ 

\midrule
    \multirow{3}*{ours-FGSM}& OA & 20.83 & 1.84 & \textbf{41.04} & \textbf{9.44} & \textbf{55.43} & \textbf{29.53} & \textbf{72.06}& \textbf{40.55} & 88.82 & \textbf{46.26} \\
   \multicolumn{1}{c}{} & AA & 43.96 & 12.11 & \textbf{72.77} & \textbf{32.16} & \textbf{83.28} & \textbf{50.42} & \textbf{90.25} &\textbf{57.14} & 97.27 & \textbf{72.30} \\
    \multicolumn{1}{c}{}& Kappa & 5.35 & -11.31 & \textbf{27.58} & \textbf{-6.73} & \textbf{43.66} & \textbf{16.47} & \textbf{63.53} & \textbf{28.14} & 84.62 & \textbf{36.12} \\
\midrule[1.4pt]
    \multirow{3}*{MI-FGSM\cite{27}}& OA & 20.48 & 1.83 & 55.87 & 24.54 & 65.41 & 40.76 & 75.60 & 46.88 & 89.25 & 52.18 \\
   \multicolumn{1}{c}{} & AA & 41.91 & 13.70 & 83.56 & 47.76 & 90.12 & 57.82 & 93.35 & 61.96 & 97.46 & 75.71 \\
    \multicolumn{1}{c}{}& Kappa & 7.89 & -10.20 & 45.73 & 11.91 & 56.98 & 28.49 & 68.59 & 36.93 & 85.20 & 43.11 \\
\midrule
    \multirow{3}*{SS-MI-FGSM\cite{19}}& OA & 22.63 & 8.04 & 54.06 & 17.85 & 64.55 & 33.97 & 76.19 & 45.80 & 89.13 & 50.98 \\
   \multicolumn{1}{c}{} & AA & 44.99 & 26.41 & 84.52 & 42.50 & 90.47 & 53.85 & 93.93 & 61.42 & 97.52 & 73.17 \\
    \multicolumn{1}{c}{}& Kappa & 9.40 & -6.46 & 43.67 & 4.47 & 56.16 & 21.19 & 69.29 & 35.37 & 85.05 & 41.90 \\
\midrule
    \multirow{3}*{SIA-MI-FGSM\cite{20}}& OA & \textbf{18.16} & \textbf{0.84} & 43.95 & 12.10 & 56.06 & 30.55 & 72.28 & 40.48 & \textbf{88.23} & 48.06 \\
   \multicolumn{1}{c}{} & AA & \textbf{40.21} &\textbf{7.51} & 75.30 & 35.89 & 83.49 & 49.09 & 90.37 & 57.54 & 97.13 & \textbf{71.91} \\
    \multicolumn{1}{c}{}& Kappa &\textbf{2.66} & -12.14 & 30.77 & -3.84 & 44.47 & 15.64 & 63.98 & 28.96 & \textbf{83.86} & 38.54 \\
\midrule
    \multirow{3}*{BSR-MI-FGSM\cite{39}} & OA & 26.63 & 2.94 & 43.10 & 11.33 & 58.42 & 33.04 & 73.65 & 42.42 & 88.66 & 50.73 \\ 
    \multicolumn{1}{c}{} & AA & 51.85 & 18.06 & 74.37 & 36.59 & 85.75 & 54.01 & 92.23 & 59.97 & 97.37 & 76.96 \\ 
    \multicolumn{1}{c}{} & Kappa & 12.84 & -7.89 & 30.03 & -3.40 & 47.92 & 19.64 & 65.98 & 30.78 & 84.43 & 41.88 \\ 
\midrule
    \multirow{3}*{ours-MI-FGSM}& OA & 20.52 & 2.08 & \textbf{40.47} & \textbf{6.18} & \textbf{54.40} & \textbf{25.69} & \textbf{71.08} &\textbf{37.06} & 88.57 & \textbf{45.90} \\
   \multicolumn{1}{c}{} & AA & 43.75 & 12.55 & \textbf{72.62} & \textbf{28.51} & \textbf{82.44} & \textbf{45.06} & \textbf{89.32} & \textbf{54.23}& \textbf{97.11} & 71.95 \\
    \multicolumn{1}{c}{}& Kappa & 5.16 & \textbf{-12.23} & \textbf{27.09} & \textbf{-11.27} & \textbf{42.23} & \textbf{9.89} & \textbf{62.10} & \textbf{23.81} & 84.28 & \textbf{35.83} \\
\bottomrule[1.4pt]
    \end{tabular}
\end{table*}

\begin{table*}[htbp]
\renewcommand\arraystretch{0.6}
\centering
\caption{Classification accuracy (\%) of adversarial attacks on five trained networks on the PaviaU dataset. The adversarial examples are crafted on VGG-11. The maximum perturbation budget is set to 0.01 and 0.03, respectively}  
    \label{Table PU VGG-11}
    \scriptsize   
    \begin{tabular}{cccccccccccc}
    \toprule
    Attack &   & \multicolumn{2}{c}{VGG-11} & \multicolumn{2}{c}{ResNet-18} & \multicolumn{2}{c}{VGG-19}& \multicolumn{2}{c}{Inc-V3}& \multicolumn{2}{c}{IncRes-V2}\\
    \cline{2-12}
    
       Method &Perturbation & \multirow{2}*{0.01} & \multirow{2}*{0.03}  &\multirow{2}*{0.01} &\multirow{2}*{0.03} &\multirow{2}*{0.01} &\multirow{2}*{0.03} &\multirow{2}*{0.01} &\multirow{2}*{0.03} &\multirow{2}*{0.01} &\multirow{2}*{0.03}\\
       &budget\\
    
\midrule
    \multirow{3}*{NONE}& OA &99.85 &99.85 &99.96 &99.96 &99.95 &99.95 &99.81 &99.81 &99.94 &99.94 \\
    \multicolumn{1}{c}{} & AA  &99.88 &99.88 &99.97 &99.97 &99.96 &99.96 &99.81 &99.81 &99.92 &99.92  \\
    \multicolumn{1}{c}{}& Kappa &99.84 &99.84 &99.94 &99.94  &99.92 &99.92 &99.80 &99.80 &99.91 &99.91 \\
\midrule[1.4pt]
    \multirow{3}*{FGSM\cite{24}}& OA & 32.18 & 12.14 & 70.14 & 37.18 & 66.34 & 50.74 & 83.09 & 60.64 & 93.62 & 55.37 \\
   \multicolumn{1}{c}{} & AA & 46.52 & 28.52 & 86.92 & 63.42 & 86.27 & 66.21 & 94.89 & 74.77 & 98.46 & 77.26 \\
    \multicolumn{1}{c}{}& Kappa & 14.02 & -4.08 & 60.05 & 24.10 & 55.30 & 37.76 & 77.05 & 49.53 & 91.00 & 45.98 \\
\midrule
    \multirow{3}*{SS-FGSM\cite{19}}& OA & 19.18 & 8.72 & 60.23 & 34.14 & 60.50 & 40.97 & 80.75 & 62.27 & 91.41 & 57.49 \\
   \multicolumn{1}{c}{} & AA & 37.98 & 17.83 & 83.82 & 62.08 & 86.04 & 65.64 & 95.51 & 78.27 & 98.03 & 80.21 \\
    \multicolumn{1}{c}{}& Kappa & 2.08 & -8.65 & 49.20 & 20.98 & 49.60 & 27.48 & 74.59 & 53.01 & 88.04 & 49.33 \\
\midrule
    \multirow{3}*{SIA-FGSM\cite{20}}& OA &\textbf{13.22} & 0.79 & 46.69 & 20.92 & 52.92 & 32.26 & 73.51 & 49.01 & 87.22 & 53.06 \\
   \multicolumn{1}{c}{} & AA & \textbf{32.77} & 6.06 & 69.30 & 42.60 & 80.84 & 50.55 & 90.93 & 64.39 & 96.96 & 71.35 \\
    \multicolumn{1}{c}{}& Kappa & \textbf{-3.44} & -16.61 & 33.82 & 4.28 & 40.04 & 16.13 & 65.34 & 36.71 & 82.59 & 43.97 \\
\midrule
    \multirow{3}*{BSR-FGSM\cite{39}} & OA & 29.64 & 9.42 & 52.75 & 21.57 & 58.59 & 37.31 & 75.34 & 46.65 & 88.00 & 53.18 \\ 
    \multicolumn{1}{c}{} & AA & 45.86 & 21.75 & 70.27 & 40.46 & 83.40 & 55.76 & 91.45 & 62.56 & 97.20 & 72.96 \\ 
    \multicolumn{1}{c}{} & Kappa & 12.61 & -5.69 & 40.25 & 5.02 & 46.58 & 21.85 & 67.68 & 33.81 & 83.59 & 43.67 \\ 

\midrule
    \multirow{3}*{ours-FGSM}& OA & 13.47 & \textbf{0.04} & \textbf{41.82}& \textbf{11.21} & \textbf{49.93} & \textbf{18.46} & \textbf{71.61} & \textbf{38.51} & \textbf{86.95} & \textbf{43.16} \\
   \multicolumn{1}{c}{} & AA & 35.82 & \textbf{1.49} & \textbf{66.05} & \textbf{35.73} & \textbf{80.24} & \textbf{38.41} & \textbf{88.67} & \textbf{55.40} & \textbf{96.82} & \textbf{63.88} \\
    \multicolumn{1}{c}{}& Kappa & -2.89 & \textbf{-19.47} & \textbf{28.90} & \textbf{-4.93} & \textbf{37.06} & \textbf{0.88} & \textbf{62.78} & \textbf{23.53} & \textbf{82.23} & \textbf{31.86} \\
\midrule[1.4pt]
    \multirow{3}*{MI-FGSM\cite{27}}& OA & 11.56 & 9.70 & 61.46 & 38.16 & 61.18 & 42.46 & 79.79 & 59.40 & 87.75 & 55.70 \\
   \multicolumn{1}{c}{} & AA & 27.09 & 12.77 & 82.00 & 60.71 & 85.29 & 63.10 & 95.21 & 75.01 & 97.19 & 79.02 \\
    \multicolumn{1}{c}{}& Kappa & -4.85 & -7.65 & 50.48 & 23.58 & 50.05 & 27.60 & 73.42 & 46.81 & 83.27 & 47.53 \\
\midrule
    \multirow{3}*{SS-MI-FGSM\cite{19}}& OA & 15.62 & 9.25 & 57.43 & 25.34 & 59.06 & 31.70 & 78.86 & 50.26 & 89.30 & 50.47 \\
   \multicolumn{1}{c}{} & AA & 35.29 & 22.38 & 82.42 & 52.71 & 85.32 & 57.06 & 94.88 & 68.43 & 97.55 & 73.40 \\
    \multicolumn{1}{c}{}& Kappa & -1.07 & -7.83 & 46.22 & 11.20 & 47.91 & 16.83 & 72.29 & 38.59 & 85.27 & 41.18 \\
\midrule
    \multirow{3}*{SIA-MI-FGSM\cite{20}}& OA & 10.35 & 0.86 & 44.37 & 19.62 & 51.85 & 24.02 & 71.67 & 41.19 & \textbf{85.59} & 47.46 \\
   \multicolumn{1}{c}{} & AA & \textbf{27.02} & 6.87 & 66.51 & 38.58 & 80.00 & 42.29 & 88.07 & 57.06 & 96.46 & 65.00 \\
    \multicolumn{1}{c}{}& Kappa & -6.81 & -16.26 & 31.06 & 2.34 & 38.78 & 7.27 & 62.88 & 27.51 & \textbf{80.51} & 37.49 \\
\midrule
    \multirow{3}*{BSR-MI-FGSM\cite{39}} & OA & 23.05 & 7.58 & 51.73 & 24.82 & 58.67 & 33.61 & 72.83 & 42.21 & 87.91 & 47.89 \\ 
    \multicolumn{1}{c}{} & AA & 39.99 & 21.58 & 66.61 & 39.45 & 82.77 & 49.62 & 87.74 & 56.66 & 96.97 & 64.06 \\ 
    \multicolumn{1}{c}{} & Kappa & 6.62 & -6.66 & 37.92 & 6.77 & 46.00 & 17.26 & 64.02 & 27.72 & 83.44 & 36.22 \\ 
\midrule
    \multirow{3}*{ours-MI-FGSM}& OA & \textbf{8.98} & \textbf{0.04} & \textbf{37.40} & \textbf{9.01} & \textbf{48.65} & \textbf{11.73} & \textbf{70.01} & \textbf{31.90} & 85.83 & \textbf{37.98} \\
   \multicolumn{1}{c}{} & AA & 29.32 & \textbf{1.56 }& \textbf{61.14} & \textbf{33.61} & \textbf{79.04} & \textbf{31.88} & \textbf{85.12} & \textbf{50.42} & \textbf{96.26} & \textbf{57.57} \\
    \multicolumn{1}{c}{}& Kappa & \textbf{-7.72} & \textbf{-19.16} & \textbf{23.88 }& \textbf{-7.38} & \textbf{35.56} & \textbf{-6.22} &\textbf{60.69} & \textbf{16.04} & 80.77 & \textbf{25.27} \\
\bottomrule[1.4pt]
    \end{tabular}
\end{table*}

\subsection{Transferability Experiments}
In this section, we evaluate the performance of the proposed method on five pretrained classification models. The evaluation metrics include Overall Accuracy (OA), Average Accuracy (AA), and the Kappa coefficient. Lower values of these metrics indicate a lower recognition accuracy of the model on adversarial examples, thereby reflecting better adversarial attack performance. In the following experimental analysis, we use the decrease in overall accuracy to reflect the increase in attack success rate. The experimental results on the HoustonU 2018 dataset are presented in Tables \ref{Table HUresnet} and \ref{Table HUVGG-11}. Results on the PaviaU dataset are shown in Tables \ref{Table PU resnet18} and \ref{Table PU VGG-11}, while those from the Indian Pines dataset are shown in Tables \ref{Table IP resnet18} and \ref{Table IP VGG-11}. The experimental results represent the averages of multiple trials to reduce the influence of randomness introduced by random operations.

\begin{table*}[htbp]
\renewcommand\arraystretch{0.6}
\centering
\caption{Classification accuracy (\%) of adversarial attacks on five trained networks on the Indian Pines dataset. The adversarial examples are crafted on ResNet-18. The maximum perturbation budget is set to 0.01 and 0.03, respectively}  
\label{Table IP resnet18}
    \scriptsize   
    \begin{tabular}{cccccccccccc}
    \toprule
    Attack &   & \multicolumn{2}{c}{ResNet-18} & \multicolumn{2}{c}{VGG-11} & \multicolumn{2}{c}{VGG-19}& \multicolumn{2}{c}{Inc-V3}& \multicolumn{2}{c}{IncRes-V2}\\
    \cline{2-12}

       Method &Perturbation & \multirow{2}*{0.01} & \multirow{2}*{0.03}  &\multirow{2}*{0.01} &\multirow{2}*{0.03} &\multirow{2}*{0.01} &\multirow{2}*{0.03} &\multirow{2}*{0.01} &\multirow{2}*{0.03} &\multirow{2}*{0.01} &\multirow{2}*{0.03}\\
       &budget\\
\midrule
    \multirow{3}*{NONE}& OA & 99.92  & 99.92 & 99.92  & 99.92 & 99.75  & 99.75 & 99.94 & 99.94 & 99.73 & 99.73\\
    \multicolumn{1}{c}{} & AA  & 99.95 & 99.95 &99.70 &99.70 &99.50  & 99.50 & 99.93 & 99.93 & 99.76 & 99.76\\
    \multicolumn{1}{c}{}& Kappa & 99.92  & 99.92  & 99.92 & 99.92 & 99.71 & 99.71 & 99.93 & 99.93 & 99.69 & 99.69\\
\midrule[1.4pt]
    \multirow{3}*{FGSM\cite{24}} & OA & 91.21 & 15.10 & 99.88 & 93.89 & 99.59 & 94.74 & 99.94 & 94.57 & 99.50 & 92.46 \\ 
    \multicolumn{1}{c}{} & AA & 89.41 & 17.85 & 99.19 & 92.30 & 99.15 & 95.61 & 99.93 & 97.70 & 99.72 & 92.89 \\ 
    \multicolumn{1}{c}{} & Kappa & 89.98 & 4.72 & 99.87 & 93.01 & 99.54 & 94.02 & 99.93 & 93.83 & 99.43 & 91.39 \\ 
\midrule
    \multirow{3}*{SS-FGSM\cite{19}} & OA & \textbf{88.91} & \textbf{3.90} & 99.88 & 86.71 & 99.69 & 91.54 & 99.94 & 90.07 & 99.54 & 90.99 \\ 
    \multicolumn{1}{c}{} & AA & \textbf{88.32} & \textbf{6.53} & 99.19 & \textbf{88.97} & 99.22 & 92.40 & 99.93 & 94.65 & 99.73 & \textbf{87.82} \\ 
    \multicolumn{1}{c}{} & Kappa & \textbf{87.36} & \textbf{-8.70} & 99.87 & 85.81 & 99.65 & 90.26 & 99.93 & 88.61 & 99.47 & 89.64 \\ 
\midrule
    \multirow{3}*{SIA-FGSM\cite{20}} & OA & 99.87 & 5.14 & 90.51 & 91.34 & 99.58 & 92.85 & 99.94 & 90.79 & 99.59 & 92.86 \\ 
    \multicolumn{1}{c}{} & AA & 98.57 & 8.21 & 90.79 & 91.11 & 98.66 & 93.51 & 99.93 & 95.94 & 99.55 & 92.26 \\ 
    \multicolumn{1}{c}{} & Kappa & 99.85 & -4.60 & 89.21 & 90.11 & 99.52 & 92.19 & 99.93 & 89.54 & 99.54 & 91.83 \\ 
\midrule
    \multirow{3}*{BSR-FGSM\cite{39}} & OA & 98.18 & 23.13 & 99.90 & 88.53 & 99.71 & 90.52 & 99.94 & 85.11 & 99.59 & 89.78 \\ 
    \multicolumn{1}{c}{} & AA & 98.97 & 35.05 & 99.07 & 93.25 & 99.45 & 94.85 & 99.93 & 94.03 & 99.70 & 91.02 \\ 
    \multicolumn{1}{c}{} & Kappa & 97.93 & 12.72 & 99.89 & 86.85 & 99.67 & 89.22 & 99.93 & 83.14 & 99.54 & 88.29 \\ 
\midrule
    \multirow{3}*{ours-FGSM} & OA & 96.46 & 18.01 & 99.90 & \textbf{86.20} & 99.65 & \textbf{88.21} & 99.92 & \textbf{82.73} & 99.63 & \textbf{89.04} \\ 
    \multicolumn{1}{c}{} & AA & 94.95 & 20.05 & 99.22 & 91.85 & 99.21 & \textbf{91.35} & 99.92 & \textbf{93.85} & 99.74 & 91.29 \\ 
    \multicolumn{1}{c}{} & Kappa & 95.97 & 10.86 & 99.89 & \textbf{84.53} & 99.60 & \textbf{86.63} & 99.91 & \textbf{80.63} & 99.58 & \textbf{87.73} \\ 
\midrule[1.4pt]
    \multirow{3}*{MI-FGSM\cite{24}} & OA &\textbf{85.45} & \textbf{4.79} & 99.88 & 91.07 & 99.58 & 92.10 & 99.94 & 90.20 & 99.50 & 90.70 \\ 
    \multicolumn{1}{c}{} & AA &\textbf{85.47} & \textbf{7.23} & 99.19 & 92.19 & 99.15 & 95.03 & 99.93 & 95.50 & 99.70 & 90.54 \\ 
    \multicolumn{1}{c}{} & Kappa &\textbf{83.43} & \textbf{-6.36} & 99.87 & 89.75 & 99.52 & 90.99 & 99.93 & 88.85 & 99.43 & 89.35 \\ 
\midrule
    \multirow{3}*{SS-MI-FGSM\cite{19}} & OA & 89.08 & 10.69 & 99.88 & 87.86 & 99.65 & 92.04 & 99.94 & 92.71 & 99.58 & 92.93 \\ 
    \multicolumn{1}{c}{} & AA & 88.44 & 15.29 & 99.19 & 91.06 & 99.18 & 95.40 & 99.93 & 95.94 & 99.77 & 93.72 \\ 
    \multicolumn{1}{c}{} & Kappa & 87.56 & -0.22 & 99.87 & 86.00 & 99.60 & 90.88 & 99.93 & 91.64 & 99.52 & 91.87 \\ 
\midrule
   \multirow{3}*{SIA-MI-FGSM\cite{20}} & OA & 89.81 & 6.51 & 99.85 & 90.87 & 99.56 & 92.25 & 99.94 & 88.93 & 99.61 & 92.26 \\ 
    \multicolumn{1}{c}{} & AA & 89.83 & 10.05 & 98.55 & 90.83 & 98.64 & 93.84 & 99.93 & 95.64 & 99.79 & 91.48 \\ 
    \multicolumn{1}{c}{} & Kappa & 88.43 & -3.08 & 99.82 & 89.62 & 99.49 & 91.20 & 99.93 & 87.50 & 99.56 & 91.17 \\ 
\midrule
    \multirow{3}*{BSR-MI-FGSM\cite{39}} & OA & 97.80 & 24.32 & 99.88 & 87.08 & 99.69 & 88.63 & 99.94 & \textbf{80.88 }& 99.58 & 87.97 \\ 
    \multicolumn{1}{c}{} & AA & 97.93 & 37.06 & 98.59 & 93.46 & 99.42 & 94.69 & 99.93 & \textbf{93.46} & 99.68 & 93.25 \\ 
    \multicolumn{1}{c}{} & Kappa & 97.49 & 14.58 & 99.87 & 85.33 & 99.65 & 87.16 & 99.93 & \textbf{78.59 }& 99.52 & 86.30 \\ 
\midrule
    \multirow{3}*{ours-MI-FGSM} & OA & 96.56 & 19.34 & 99.88 & \textbf{85.44} & 99.61 & \textbf{87.57} & 99.92 & 83.08 & 99.67 & \textbf{87.19} \\ 
    \multicolumn{1}{c}{} & AA & 94.92 & 21.57 & 98.59 & \textbf{90.11} & 99.18 &\textbf{92.49} & 99.92 & 94.21 & 99.77 & \textbf{90.04} \\ 
    \multicolumn{1}{c}{} & Kappa & 96.08 & 12.32 & 99.87 & \textbf{83.67}& 99.56 & \textbf{86.29} & 99.91 & 81.07 & 99.63 & \textbf{85.63} \\ 
\bottomrule[1.4pt]
    \end{tabular}
\end{table*}

\begin{table*}[htbp]
\renewcommand\arraystretch{0.6}
\centering
\caption{Classification accuracy (\%) of adversarial attacks on five trained networks on the Indian Pines dataset. The adversarial examples are crafted on VGG-11. The maximum perturbation budget is set to 0.01 and 0.03, respectively}  
\label{Table IP VGG-11}
    \scriptsize   
    \begin{tabular}{cccccccccccc}
    \toprule
    Attack &   & \multicolumn{2}{c}{VGG-11} & \multicolumn{2}{c}{ResNet-18} & \multicolumn{2}{c}{VGG-19}& \multicolumn{2}{c}{Inc-V3}& \multicolumn{2}{c}{IncRes-V2}\\
    \cline{2-12}

       Method &Perturbation & \multirow{2}*{0.01} & \multirow{2}*{0.03}  &\multirow{2}*{0.01} &\multirow{2}*{0.03} &\multirow{2}*{0.01} &\multirow{2}*{0.03} &\multirow{2}*{0.01} &\multirow{2}*{0.03} &\multirow{2}*{0.01} &\multirow{2}*{0.03}\\
       &budget\\
\midrule
    \multirow{3}*{NONE}& OA & 99.92 & 99.92 & 99.92 & 99.92 & 99.75 & 99.75 & 99.94 & 99.94 & 99.73 & 99.73\\
    \multicolumn{1}{c}{} & AA  & 99.70 & 99.70  & 99.95 & 99.95 & 99.50  & 99.50 & 99.93 & 99.93 & 99.76 & 99.76\\
    \multicolumn{1}{c}{}& Kappa & 99.92 & 99.92 & 99.92 & 99.92 & 99.71 & 99.71 & 99.93 & 99.93 & 99.69 & 99.69\\
\midrule[1.4pt]
    \multirow{3}*{FGSM\cite{24}} & OA  & \textbf{95.81} & 46.71 & 99.83 & 97.16 & 99.75 & 99.00 & 99.94 & 97.58 & 99.63 & 93.64 \\
    \multicolumn{1}{c}{} & AA  & 92.64 & 51.44 & 99.17 & 97.59 & 99.03 & 98.45 & 99.93 & 99.11 & 99.71 & 90.26 \\
    \multicolumn{1}{c}{} & Kappa & \textbf{95.21} & 40.28 & 99.80 & 96.77 & 99.71 & 98.85 & 99.93 & 97.25 & 99.58 & 92.77 \\
\midrule
    \multirow{3}*{SS-FGSM\cite{19}} & OA  & 95.88 & 25.90 & 99.85 & 90.24 & 99.73 & 95.79 & 99.94 & 97.29 & 99.71 & 97.26 \\
    \multicolumn{1}{c}{} & AA  &\textbf{92.09} & 29.46 & 99.65 & 92.96 & 99.01 & 96.78 & 99.93 & 98.18 & 99.75 & 94.53 \\
    \multicolumn{1}{c}{} & Kappa & 95.30 & 16.52 & 99.82 & 88.81 & 99.69 & 95.19 & 99.93 & 96.90 & 99.67 & 96.87 \\
\midrule
    \multirow{3}*{SIA-FGSM\cite{20}} & OA & 97.33 & \textbf{23.62} & 99.77 & 84.70 & 99.63 & 91.20 & 99.94 & 88.45 & 99.54 & 87.30 \\ 
    \multicolumn{1}{c}{} & AA & 92.71 &\textbf{27.52} & 99.12 & 91.01 & 98.70 & 93.40 & 99.93 & 95.01 & 99.66 & 91.22 \\ 
    \multicolumn{1}{c}{} & Kappa & 96.95 & \textbf{14.65} & 99.74 & 82.53 & 99.58 & 89.98 & 99.93 & 86.86 & 99.47 & 85.53 \\ 
\midrule
    \multirow{3}*{BSR-FGSM\cite{39}} & OA & 99.85 & 83.70 & 99.88 & 94.17 & 99.73 & 98.48 & 99.94 & 98.21 & 99.69 & 91.13 \\ 
    \multicolumn{1}{c}{} & AA & 98.77 & 79.41 & 99.91 & 97.41 & 99.47 & 98.88 & 99.93 & 99.32 & 99.74 & 95.94 \\ 
    \multicolumn{1}{c}{} & Kappa & 99.82 & 81.67 & 99.87 & 93.40 & 99.69 & 98.26 & 99.93 & 97.96 & 99.65 & 89.97 \\ 
\midrule
    \multirow{3}*{ours-FGSM} & OA & 98.49 & 28.65 & 99.79 & \textbf{69.73} & 99.63 & \textbf{86.10} & 99.94 & \textbf{77.44} & 99.54 &\textbf{83.34} \\ 
    \multicolumn{1}{c}{} & AA & 94.76 & 33.78 & 99.16 & \textbf{78.78} & 98.68 & \textbf{89.98} & 99.93 & \textbf{90.82} & 99.64 & \textbf{83.95} \\ 
    \multicolumn{1}{c}{} & Kappa & 98.28 & 19.10 & 99.76 & \textbf{65.53} & 99.58 & \textbf{84.18} & 99.93 & \textbf{74.50} & 99.47 & \textbf{80.93} \\ 
\midrule[1.4pt]
    \multirow{3}*{MI-FGSM\cite{24}} & OA & \textbf{78.04} & 15.68 & 99.79 & 92.83 & 99.71 & 95.28 & 99.94 & 94.86 & 99.56 & 90.99 \\
    \multicolumn{1}{c}{} & AA & \textbf{72.87} & 10.32 & 99.15 & 94.50 & 98.77 & 95.50 & 99.93 & 97.41 & 99.67 & 88.82 \\
    \multicolumn{1}{c}{} & Kappa & \textbf{74.91} & 5.63  & 99.76 & 91.81 & 99.67 & 94.63 & 99.93 & 94.13 & 99.49 & 89.76 \\
\midrule
    \multirow{3}*{SS-MI-FGSM\cite{19}} & OA & 95.54 & 46.04 & 99.83 & 77.17 & 99.75 & 92.50 & 99.94 & 91.96 & 99.63 & 97.68 \\
    \multicolumn{1}{c}{} & AA & 91.86 & 54.03 & 99.17 & 87.76 & 99.03 & 96.12 & 99.93 & 96.78 & 99.71 & 92.93 \\
    \multicolumn{1}{c}{} & Kappa & 94.90 & 39.40 & 99.80 & 74.36 & 99.71 & 91.46 & 99.93 & 90.86 & 99.58 & 97.35 \\
\midrule
    \multirow{3}*{SIA-MI-FGSM\cite{20}} & OA & 94.07 & \textbf{14.08} & 99.67 & 80.79 & 99.58 & 89.07 & 99.92 & 83.36 & 99.46 & 82.22 \\ 
    \multicolumn{1}{c}{} & AA & 89.22 & \textbf{16.00} & 99.07 & 89.91 & 98.64 & 91.81 & 99.93 & 92.75 & 99.13 & 85.58 \\ 
    \multicolumn{1}{c}{} & Kappa & 93.21 & \textbf{4.48} & 99.63 & 78.26 & 99.52 & 87.58 & 99.91 & 81.20 & 99.38 & 79.79 \\ 
\midrule
    \multirow{3}*{BSR-MI-FGSM\cite{39}} & OA & 99.75 & 58.68 & 99.87 & 88.24 & 99.69 & 94.39 & 99.94 & 92.34 & 99.61 & 80.32 \\ 
    \multicolumn{1}{c}{} & AA & 98.22 & 62.38 & 99.89 & 95.03 & 99.44 & 97.38 & 99.93 & 97.53 & 99.69 & 89.49 \\ 
    \multicolumn{1}{c}{} & Kappa & 99.71 & 54.61 & 99.85 & 86.81 & 99.65 & 93.64 & 99.93 & 91.36 & 99.56 & 78.05 \\ 
\midrule
    \multirow{3}*{ours-MI-FGSM} & OA & 96.79 & 21.99 & 99.73 & \textbf{67.49} & 99.63 & \textbf{83.46} & 99.92 & \textbf{72.43} & 99.48 & \textbf{77.20} \\ 
    \multicolumn{1}{c}{} & AA & 93.04 & 24.91 & 99.13 & \textbf{75.89} & 98.68 & \textbf{87.86} & 99.93 & \textbf{88.54} & 99.62 & \textbf{75.50} \\ 
    \multicolumn{1}{c}{} & Kappa & 96.33 & 11.80 & 99.69 & \textbf{63.16} & 99.58 & \textbf{81.22} & 99.91 & \textbf{68.99} & 99.40 & \textbf{73.56} \\ 
\bottomrule[1.4pt]
    \end{tabular}
\end{table*}

\textit{HoustonU 2018 Dataset:} The experimental results are shown in Tables \ref{Table HUresnet} and \ref{Table HUVGG-11}. Adversarial examples were generated using the substitute models ResNet-18 and VGG-11, respectively. It can be seen that across all methods and models, the attack effect of the perturbation budget of 0.03 is almost significantly better than the perturbation budget of 0.01, indicating that smaller perturbation budget pose greater challenges. When combined with MI-FGSM, our method achieved comparable or even better improvements at a budget of 0.01 compared to 0.03. This indicates that it can effectively enhance attack performance even under smaller perturbation budgets. The attack performance of our method across all models is significantly better than that of the baseline method. In terms of overall accuracy (OA), when the perturbation budget of 0.01, the average improvement was 14.06\%, while for a budget of 0.03, the average improvement was 25.5\%. Compared to the SIA method on four black-box models, the attack success rate increased by an average of 2.44\% at $\varepsilon$ = 0.01 and 3.28\% at $\varepsilon$ = 0.03. However, it can be observed that the white-box attack success rate decreased significantly compared to SIA. Data from ablation experiments indicate that this is caused by spectral-wise block transformations and weighted feature divergence loss. The possible reason is that the use of multi-dimensional block transformations and feature loss alter the adversarial example optimization process. Our method not only focuses on the classification boundary of the substitute model but also pays more attention to the change of feature distribution. Consequently, this may lead to a more dispersed disturbance in the input space, thereby diminishing the effectiveness of attacks on the substitute model and resulting in a poorer performance in white-box attacks.

\textit{PaviaU Dataset:} The experimental results are shown in Tables \ref{Table PU resnet18} and \ref{Table PU VGG-11}. We present the results of adversarial examples generated on the ResNet-18 and VGG-11 models. Unlike the HoustonU 2018 dataset, attacking black-box models on this dataset is more challenging due to its higher number of spectral bands. On this dataset, our method largely retained the attack success rate on white-box models while significantly improving baseline attack transferability. Moreover, the performance improvement on the PaviaU dataset was slightly better than that on the HoustonU 2018 dataset. Compared to the baseline method across all models, the attack success rate of this method increased by an average of 13.12\% at $\varepsilon$ = 0.01 and 18.07\% at $\varepsilon$ = 0.03. Compared to the SIA method across four black-box models, the attack success rate increased by an average of 2.05\% at $\varepsilon$ = 0.01 and 7.31\% at $\varepsilon$ = 0.03. Compared to ResNet-18, the adversarial examples generated when VGG-11 is used as a substitute model show greater improvements in attack performance. This is due to differences in model structure and feature maps extraction mechanisms, which lead to varying attack effects.

\begin{table}[htbp]
\centering
\caption{Overall accuracy (\%) of HSI adversarial examples under added random noise conditions. The adversarial examples for the three datasets are generated using the VGG-11 model, with the perturbation budget set to 0.03 and the intensity of the random noise set to 0.1}  
\label{Table Defense noise}
\scalebox{0.9}{
    \begin{tabular}{c|ccccc}
    \toprule
      \textbf{PaviaU} &VGG-11  & ResNet-18 & VGG-19 & Inc-V3 &IncRes-V2  \\
\midrule
    FGSM & 13.09  & 37.85  & 60.61 &57.84 & \textbf{46.20}\\
    SS-FGSM & 9.23  & 35.42 & 65.10 & 62.35  & 53.65 \\
    SIA-FGSM  & 0.89  & 21.79 & 54.87  & 50.19 & 52.15 \\
    BSR-FGSM & 11.81 & 21.80 & 52.28 & 47.64& 51.01 \\
    ours-FGSM & \textbf{0.05}  & \textbf{11.54}  & \textbf{45.71} & \textbf{39.78} & 50.30 \\
\midrule
    MI-FGSM  & 9.79  & 39.82 & 69.73 & 59.66  & 48.11 \\
    SS-MI-FGSM  & 9.80  & 26.47  & 53.80  & 49.90  & 53.14 \\
    SIA-MI-FGSM  & 1.04 & 20.55 & 46.31 & 43.06  & 49.68 \\
    BSR-MI-FGSM & 8.49 & 25.74 & 42.39 & 42.74 & 47.83\\
    ours-MI-FGSM  & \textbf{0.09} & \textbf{9.21}  & \textbf{34.55}  & \textbf{32.82} & \textbf{45.16} \\
\midrule[1.4pt]
    \textbf{HoustonU 2018}  &VGG-11  & ResNet-18 & VGG-19 & Inc-V3 &IncRes-V2  \\
    \midrule
    FGSM  & 59.96 & 69.17 & 67.32  & 67.36  & 63.14 \\
    SS-FGSM & 40.31 & 69.02 & 61.85  & 70.40  & 66.35 \\
    SIA-FGSM & \textbf{7.15}  & 40.63  & 33.75  & 45.64  & 37.29 \\
    BSR-FGSM & 27.44 & 47.30 & 49.50 & 48.75 & 39.57 \\
    ours-FGSM & 15.75 & \textbf{32.02}  & \textbf{26.79} & \textbf{37.79} & \textbf{30.14} \\
\midrule
    MI-FGSM & 48.38 & 65.16 & 57.25  & 65.74 & 63.51 \\
    SS-MI-FGSM  & 45.26 & 63.06 & 56.78 & 65.07  & 61.74 \\
    SIA-MI-FGSM  & \textbf{7.19}  & 25.10  & \textbf{20.17}  & \textbf{26.87} & \textbf{23.60} \\
    BSR-MI-FGSM  & 24.73 & 32.91 & 36.07 & 34.01 & 28.86\\
    ours-MI-FGSM  & 17.13  & \textbf{24.22}  & 21.96 & 28.60  & 24.23 \\
\midrule[1.4pt] 
    \textbf{Indian Pines} &VGG-11 & ResNet-18 & VGG-19 & Inc-V3 & IncRes-V2   \\
\midrule
    FGSM & 79.20 & 96.68 & 98.36 & 96.06 & 75.63 \\ 
    SS-FGSM & 70.18 & 88.85 & 95.92 & 95.52 & 68.61 \\ 
    SIA-FGSM & 40.61 & 83.42 & 89.80 & 84.73 & 66.27 \\ 
    BSR-FGSM & 90.38 & 94.20 & 96.79 & 96.83 & 78.82 \\ 
    ours-FGSM & \textbf{39.24} &\textbf{67.55}&\textbf{83.46} & \textbf{71.45} & \textbf{55.22} \\ 
\midrule
    MI-FGSM & 60.96 & 91.90 & 95.32 & 94.09 & 71.94 \\ 
    SS-MI-FGSM & 67.55 & 76.03 & 92.81 & 89.29 & 68.46 \\ 
    SIA-MI-FGSM & \textbf{35.93} & 78.49 & 86.05 & 78.20 & 62.39 \\ 
    BSR-MI-FGSM & 73.68 & 86.63 & 90.78 & 89.83 & 71.16 \\ 
    ours-MI-FGSM & 37.11 & \textbf{65.31} & \textbf{79.26} & \textbf{64.48} & \textbf{51.86} \\ 
    \bottomrule[1.4pt]
    \end{tabular}
}
\end{table}

\begin{table}[t]
\centering
\caption{Overall accuracy (\%) of HSI adversarial examples under spectral filtering conditions. The adversarial examples for the three datasets are generated using the VGG-11 model, with the perturbation budget set to 0.03 and the filter size range set to 7}  
\label{Table Defense filtering}
\scalebox{0.9}{
    \begin{tabular}{c|ccccc}
    \toprule
    \textbf{PaviaU} &VGG-11 & ResNet-18 & VGG-19 & Inc-V3 & IncRes-V2   \\
\midrule
    FGSM  & 14.01 & 38.85 & 47.46 & 64.86  & 65.03 \\
    SS-FGSM  & 9.43  & 35.78  & 44.46  & 64.88  & 64.31 \\
    SIA-FGSM  & 3.20  & 22.52  & 40.94  & 57.32  & 62.36 \\
    BSR-FGSM  & 15.29 & 21.98 & 44.87 & 56.73 & 61.14\\
    ours-FGSM  & \textbf{1.10}  & \textbf{12.27} & \textbf{29.29}  & \textbf{49.33}  & \textbf{54.24} \\
\midrule
    MI-FGSM  & 10.84  & 40.79 & 47.31  & 65.24  & 66.78 \\
    SS-MI-FGSM  & 9.85  & 27.54  & 36.26  & 54.27  & 54.42 \\
    SIA-MI-FGSM & 1.95  & 20.59  & 36.04  & 51.47  & 57.75 \\
    BSR-MI-FGSM  & 12.82 & 25.64 & 43.36 & 51.99 & 59.02\\
    ours-MI-FGSM & \textbf{1.09} & \textbf{9.42}  & \textbf{21.13}  & \textbf{42.96}  & \textbf{48.66} \\
\midrule[1.4pt]
    
    \textbf{HoustonU 2018} &VGG-11 & ResNet-18 & VGG-19 & Inc-V3 & IncRes-V2   \\
\midrule
    FGSM  & 60.21 & 70.78  & 67.79  & 68.71  & 65.91 \\
    SS-FGSM  & 40.11  & 70.59  & 62.84  & 71.39  & 68.28 \\
    SIA-FGSM  & \textbf{7.46} & 42.23  & 36.68  & 48.91  & 41.94 \\
    BSR-FGSM & 30.29 & 49.46 & 51.60 & 51.81 & 45.69\\
    ours-FGSM  & 15.64  & \textbf{32.81}  & \textbf{28.67} & \textbf{39.22}  & \textbf{33.28} \\
\midrule
    MI-FGSM  & 48.29  & 64.97& 57.60  & 65.78  & 63.91 \\
    SS-MI-FGSM & 44.86 & 62.80  & 56.90  & 64.85  & 62.56 \\
    SIA-MI-FGSM & \textbf{7.40}  & 25.15  & \textbf{21.37}  & \textbf{28.27}  & \textbf{26.26} \\
    BSR-MI-FGSM & 24.42 & 33.55 & 37.22 & 35.32 & 32.37\\
    ours-MI-FGSM  & 16.77  & \textbf{24.38} & 22.81  & 29.71  & 26.28 \\
 \midrule[1.4pt] 
    \textbf{Indian Pines} &VGG-11 & ResNet-18 & VGG-19 & Inc-V3 & IncRes-V2   \\
\midrule
    FGSM & 70.27 & 87.57 & 98.03 & 95.25 & 96.85 \\ 
    SS-FGSM & 52.57 & 83.05 & 94.80 & 93.35 & 97.26 \\ 
    SIA-FGSM & \textbf{39.53} & 78.62 & 91.46 & 87.03 & 88.96 \\ 
    BSR-FGSM & 89.80 & 84.19 & 96.35 & 97.53 & 91.55 \\ 
    ours-FGSM & 51.04 & \textbf{67.30} & \textbf{86.97} & \textbf{79.36} & \textbf{84.60} \\ 
\midrule
    MI-FGSM & \textbf{34.42} & 85.99 & 95.34 & 92.75 & 94.82 \\ 
    SS-MI-FGSM & 63.24 & 74.26 & 91.77 & 85.08 & 94.01 \\ 
    SIA-MI-FGSM & 36.18 & 74.84 & 88.48 & 84.60 & 86.32 \\ 
    BSR-MI-FGSM & 78.55 & 76.77 & 92.29 & 91.36 & 83.40 \\ 
    ours-MI-FGSM & 44.55 & \textbf{63.66} & \textbf{83.98} &\textbf{75.71} & \textbf{78.32} \\ 
\bottomrule[1.4pt]
    \end{tabular}
}
\end{table}

\textit{Indian Pines Dataset:} The experimental results are shown in Tables \ref{Table IP resnet18} and \ref{Table IP VGG-11}. First, in contrast to the other two datasets, adversarial examples generated on this dataset with a perturbation budget of $\varepsilon$ = 0.01 exhibit no transferability. We believe that this is because the dataset is crop and soil zoning, with small inter-class differences and highly correlated spectra. Different networks will capture different small spectral segments, resulting in low cosine similarity between input gradient direction of the substitute model and black-box model, making small perturbations insufficient to reliably change the prediction. Consequently, we consider $\varepsilon$ = 0.01 not informative for comparison. At $\varepsilon$ = 0.03, this method improves the black-box attack success rate by approximately 12.1\% on average over the baseline methods and by about 8.7\% over SIA. Compared with the SIA method, the BSR performs better on ResNet-18 but worse on VGG-11, whereas our method is particularly effective when VGG-11 is used as the substitute model.

\subsection{Defense Experiments}
To verify the robustness of the proposed method against common defense strategies, we conducted experiments on three public datasets using two defense methods: adding random noise and spectral filtering. Both defense techniques are commonly used in image processing to mitigate or disrupt adversarial perturbations, thereby improving model robustness. Comparison results are presented in Tables \ref{Table Defense noise} and \ref{Table Defense filtering}.

\textit{Adding random noise:} Before feeding examples into the model, random noise is added to the examples to disrupt the structure of adversarial perturbations, reducing the model’s sensitivity to adversarial examples and thus weakening the effectiveness of adversarial attacks. The budget of the added random noise is set to 0.1, and the results are shown in Table \ref{Table Defense noise}. It can be observed that our proposed method maintains good attack performance even under the random noise defense.

\textit{Spectral filtering:} HSIs contain redundant spectral bands, therefore, local spectral filtering is a challenging defense method. We perform average filtering along the spectral dimension to validate the robustness of the attack method. The filter size is set to 7, and the results are presented in Table \ref{Table Defense filtering}. Our method demonstrates strong adaptability and robustness across different datasets and defense strategies.

\subsection{Ablation Experiment}
In this section, we explore the influence of each component of the generation method on attack performance through ablation experiments. The relevant parameters have been described in the previous section on parameter settings. The experimental results are summarized in Table \ref{Table xiaorong}. The components in the table are explained as follows: SpatialT refers to block partitioning applied only to the spatial dimension; SpectralT refers to block partitioning applied only to the spectral dimension; WFDL-unEnlarge refers to the use of weighted feature divergence loss without enlarging the original example feature values; WFDL refers to the use of weighted feature divergence loss with enlarging of the original example feature values.

\begin{table*}[!b]
\renewcommand\arraystretch{0.9}
    \centering
    \caption{Ablation study of the proposed approach. The adversarial examples are crafted on the VGG-11 model, with the integrated baseline method being FGSM. The data represents the overall accuracy (\%), with the perturbation budget set to 0.03} 
    \label{Table xiaorong}
        \begin{tabular}{cccc|ccccc}
        \toprule
         \multicolumn{4}{c}{Constrain condition}  & \multicolumn{5}{c}{PaviaU}   \\
        \midrule
        SpatialT & SpectralT & WFDL-unEnlarge & WFDL  &VGG-11  & ResNet-18 & VGG-19 & Inc-V3 &IncRes-V2\\
\midrule
        \checkmark & & & & 0.75 & 21.03 & 32.25 & 48.94 & 52.89 \\
         &\checkmark  & & & 0.55 & 16.40 & 23.66 & 45.62 & 48.26 \\
        \checkmark &\checkmark & & & 0.41 & 17.18 & 25.51 & 45.06 & 49.08 \\
        \checkmark & & \checkmark & & 0.51 & 15.54 & 26.35 & 43.73 & 46.64 \\
        \checkmark &\checkmark&\checkmark& & 0.19 & 12.72 & 20.20 & 39.77 & 43.90 \\
        \checkmark &\checkmark &\checkmark &\checkmark & 0.04 & 11.17 & 18.56 & 38.40 & 43.02 \\
\midrule[1.3pt]
        \multicolumn{4}{c}{Constrain condition}  & \multicolumn{5}{c}{HoustonU 2018}   \\
        \midrule
        SpatialT & SpectralT & WFDL-unEnlarge & WFDL &VGG-11  & ResNet-18 & VGG-19 & Inc-V3 &IncRes-V2 \\
        \midrule
        \checkmark & & & & 6.72 & 40.51 & 33.53 & 46.28 & 39.65 \\
        & \checkmark & & & 15.61 & 36.86 & 30.26 & 41.44 & 36.14 \\
        \checkmark & \checkmark & & & 13.41 & 32.88 & 28.83 & 39.28 & 34.27 \\
        \checkmark & & \checkmark & & 16.56 & 38.94 & 32.84 & 44.48 & 36.69 \\
        \checkmark & \checkmark & \checkmark & & 16.56 & 32.63 & 28.16 & 38.72 & 33.17 \\
        \checkmark & \checkmark & \checkmark & \checkmark & 15.18 & 31.27 & 26.71 & 37.79 & 31.87 \\
        \bottomrule
        \end{tabular}
\end{table*}

Although performance slightly varies across different models, the attack effectiveness generally improves with the application of our proposed method. The experimental results show that partitioning the data only along the spectral dimension and applying transformations also reduces the model's classification accuracy. By combining both spatial and spectral dimensions for 3D structure-invariant transformations, the transferability on the two datasets improved by approximately 4\% and 6\%, respectively. In addition, the weighted feature divergence loss, which takes into account the differences between feature channels, effectively improves the adversarial transferability. Furthermore, the operation of enlarging the original feature values when calculating the differences between feature maps further enhances the attack effectiveness of the proposed loss, thereby validating the effectiveness of the method.

\subsection{Parameter Sensitivity Analysis Experiments}
To gain insight into the performance of the method, we further analyze the influence of various hyperparameters on the experimental results. The experiments were conducted mainly on the PaviaU dataset, with VGG-11 selected as the substitute model for generating adversarial examples, while the other models were treated as black-box models. The perturbation budget was set to 0.03.

\begin{figure*}[htbp]
    \centering
     \includegraphics[width=1\textwidth]{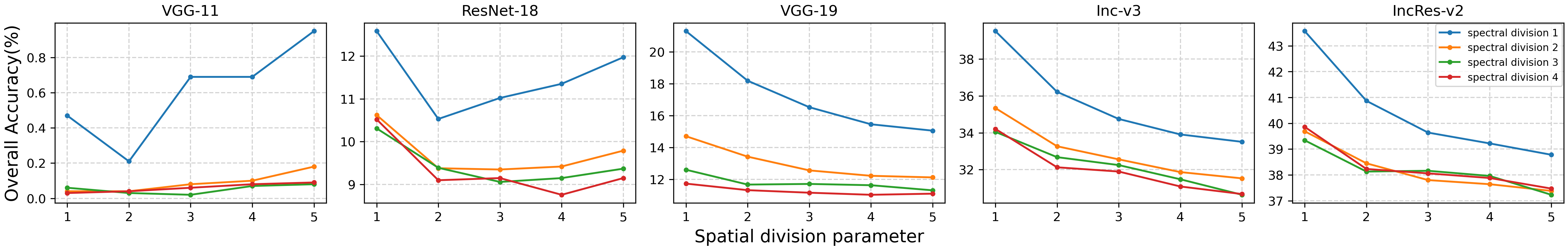}  
    \caption{The impact of different division parameters used in the spatial and spectral dimensions on the results.}
    \label{divide parameters}
\end{figure*}

The influence of division parameters on attack performance is shown in Fig. \ref{divide parameters}. The range for spatial division parameters was set to 1-5, while that for spectral division was set to 1-4. The blue line represents the impact of varying spatial division parameters on overall accuracy when the spectral dimension is not divided. Lines of other colors represent different spectral division parameters. Comparing the four lines across all models, the blue line exhibits significant fluctuation with parameter changes and consistently achieves the highest overall accuracy. The other lines are below, indicating that the operation of data chunking in the spectral dimension can improve the success rate of attacks. The distance between the lines reflects the enhanced attack effectiveness gained from spectral division. In the VGG-11 model, the green line (spectral division parameter 3) reaches its lowest point when the spatial division parameter is also 3, demonstrating the strongest attack effectiveness on the white-box model under this configuration. For other black-box models, most lines decrease as the spatial division parameter increases, this shows that the combination of the two has a good promoting effect. On some models, the lowest accuracy is achieved when the spectral division parameter is 4 and the spatial parameter is 4 or 5. It is observed that the results for spectral division parameters 3 and 4 (green and red lines) are quite similar. Considering that increased division inevitably raises computational complexity, and to balance attack effectiveness with computational efficiency, we sets both the spatial and spectral division parameters to 3.

\begin{figure*}[htbp]
\centering
\subfigure[]
{
    \begin{minipage}[b]{.23\linewidth}
        \centering     
        \includegraphics[scale=0.35]{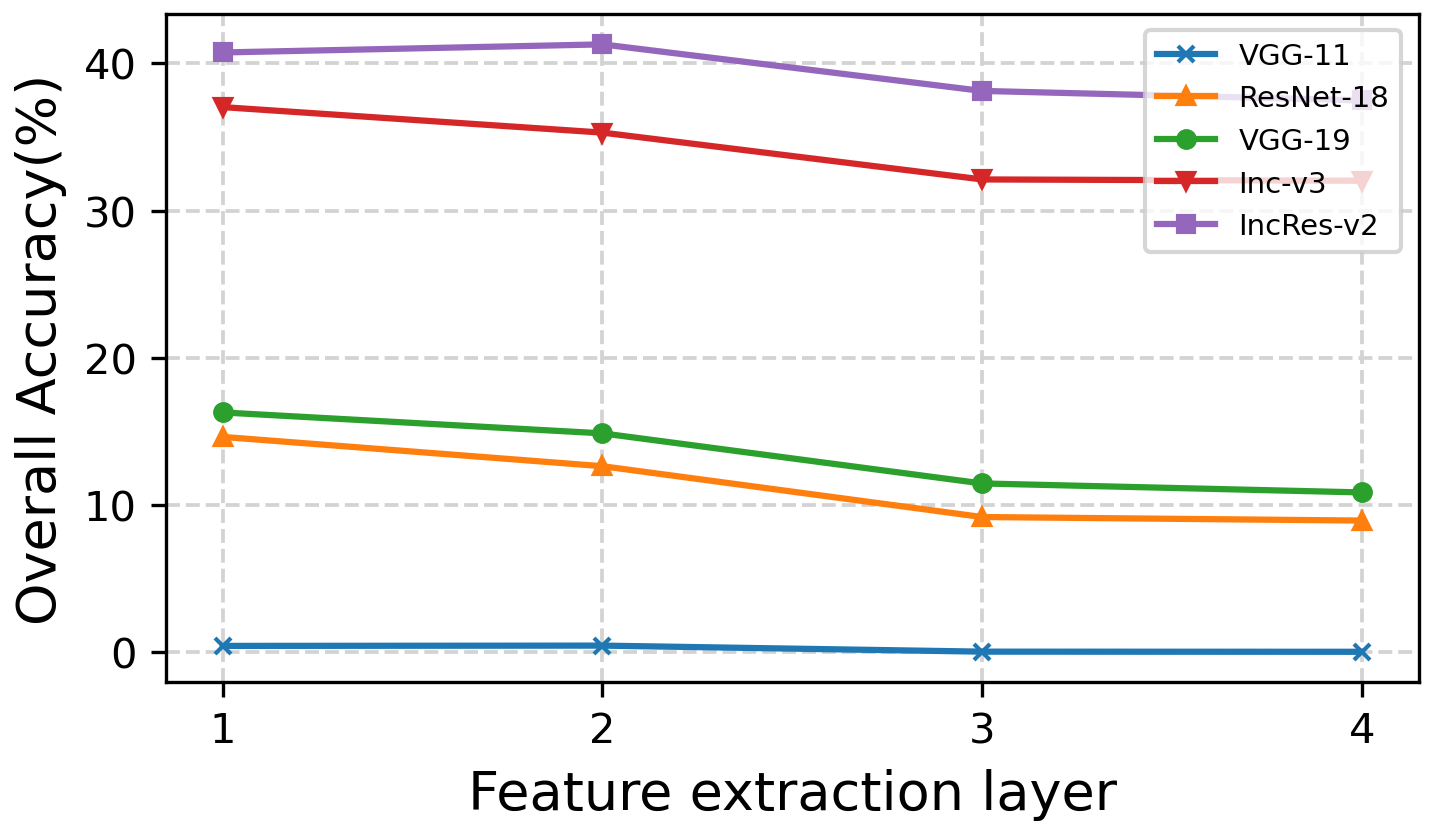}
    \end{minipage}
    \begin{minipage}[b]{.23\linewidth}
        \includegraphics[scale=0.35]{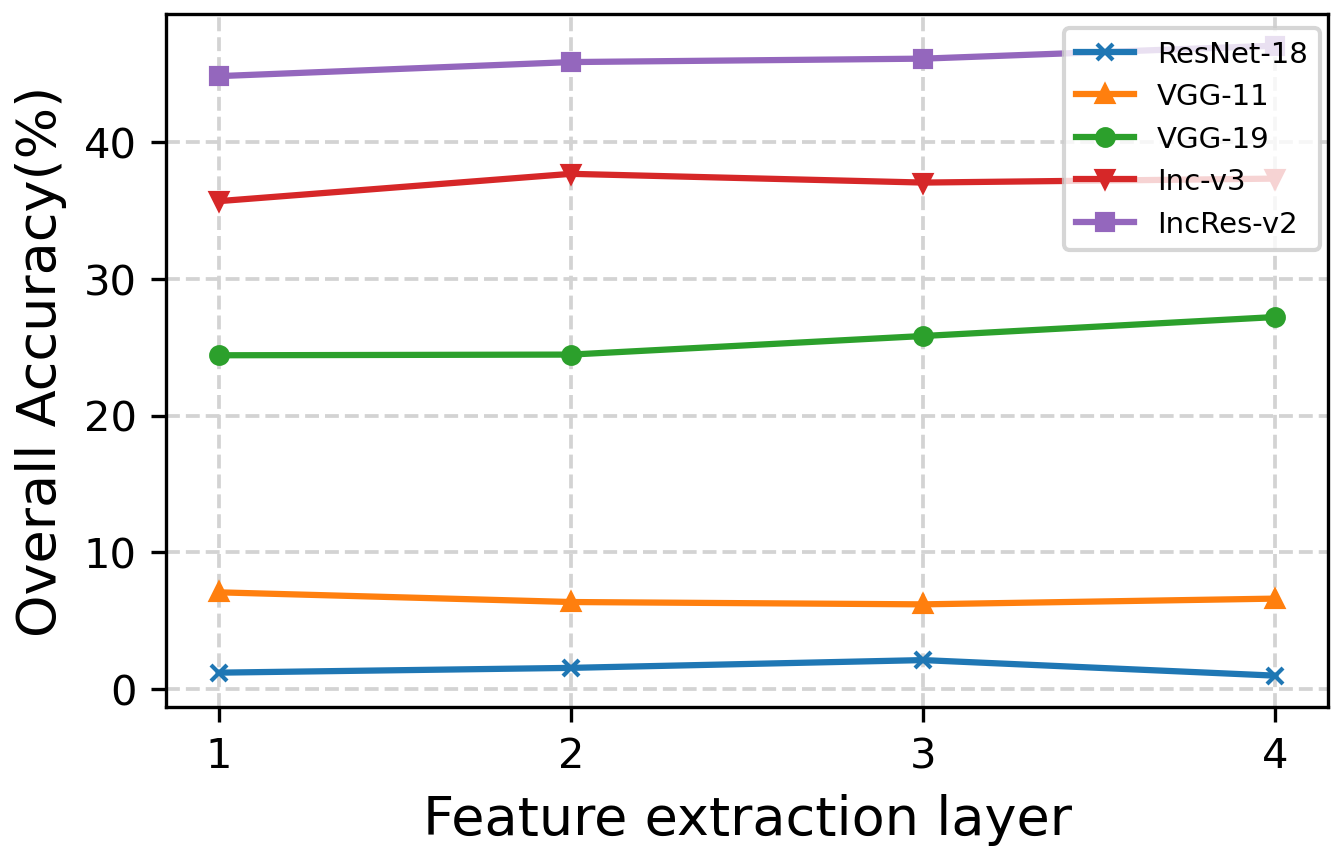} 
    \end{minipage}
    \label{feature layer}
}
\subfigure[]{
    \begin{minipage}[b]{.23\linewidth}
        \centering
        \includegraphics[scale=0.35]{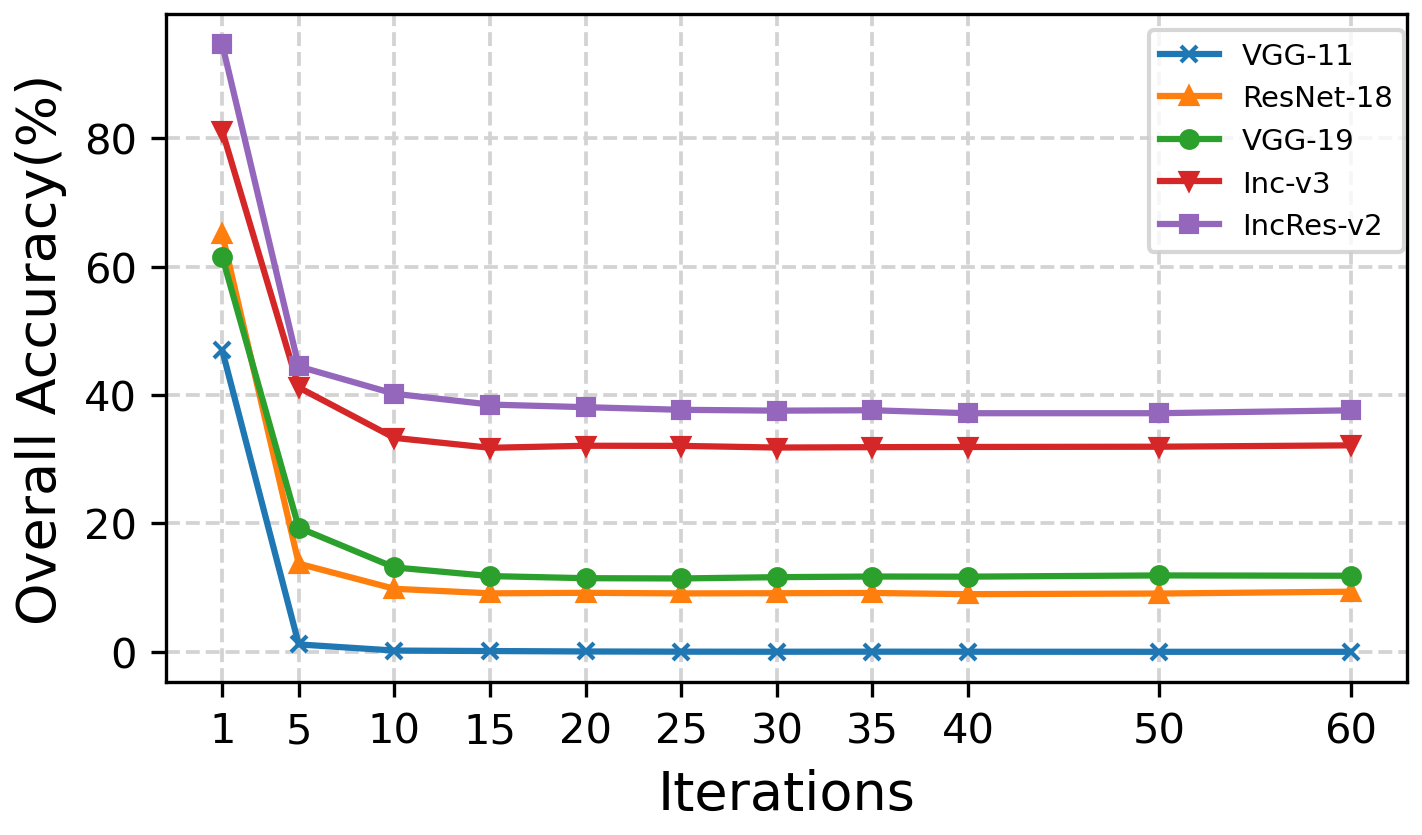}
    \end{minipage}
    \label{iterations}
}
\subfigure[]{
    \begin{minipage}[b]{.23\linewidth}
        \centering
    \includegraphics[scale=0.35]{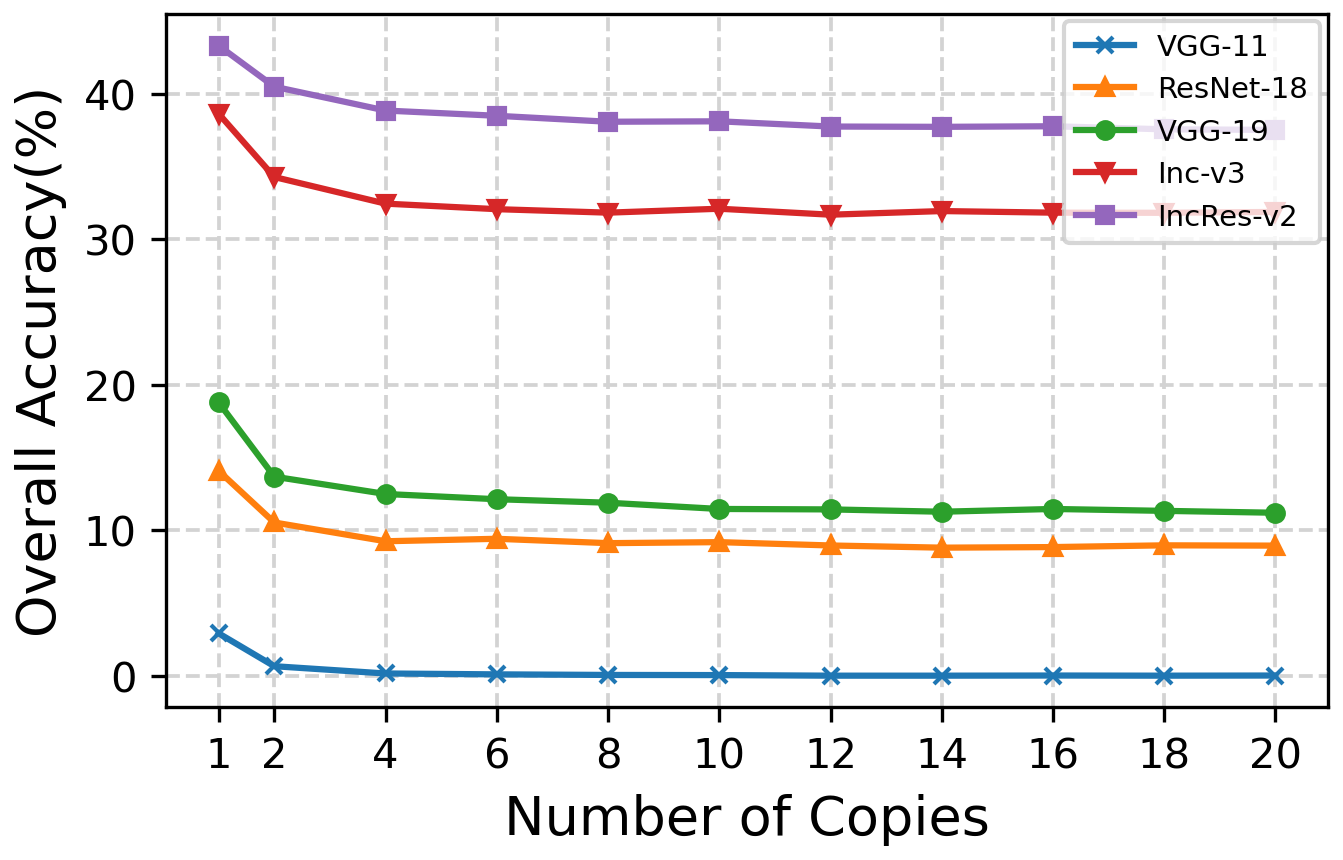}
    \end{minipage}
    \label{Number of Copies}
}
\caption{The impact of (a) the network extraction layer parameter $l$, (b) the number of iterations and (c) the number of copies on the results.}
\label{hyperparameters a}
\end{figure*}

The impact of the feature maps extraction layer $l$ on attack performance is shown in Fig. \ref{feature layer}. The fourth layer of the VGG-11 model demonstrates the most effective attack performance, with classification accuracy showing a declining trend as deeper layer feature maps are used. However, the first layer of the ResNet-18 model has the best attack effect. Notably, using feature maps from other layers in ResNet-18 has a relatively minor impact on classification accuracy. We analyze this discrepancy from the perspective of model architecture: the VGG-11 model utilizes simply stacked convolutional layers, whereas the ResNet-18 model incorporates residual connections (skip connections). Since the VGG-11 model lacks skip connections, the feature maps exhibit greater variation between different layers. Given that the results show that using deeper features generates adversarial examples with stronger transferability, it indicates that these features are more abstract and generalizable. Conversely, in ResNet-18 model, intermediate layer feature maps exhibit greater similarities due to the skip connections. Consequently, the choice of intermediate layer feature maps has a smaller effect on attack performance within ResNet-18. In this paper, based on experimental results across four corresponding layers, we selected the layer 3 as a compromise result for both substitute models to ensure relative fairness.

Fig. \ref{iterations} shows the effect of iteration steps $I$ on attack performance. The iterations range from 1 to 60. When the number of iterations exceeds 20 times, the classification accuracy hardly decreases any more, and the improvement of attack effect is very limited. Therefore, to optimize computational costs while considering the attack's efficacy, we set the number of iterations to 20.

The impact of the number of copies $N $ on attack performance is shown in Fig. \ref{Number of Copies}. The number of transformed copies ranges from 1 to 20. As the number of transformed copies increases, the accuracy decreases most rapidly from 1 to 6, and then gradually stabilizes. This phenomenon indicates that additional transformed examples have little effect on improving attack performance. Therefore, in order to maintain a stable attack effect, we set N to 10.

\begin{figure*}[t]
\centering
\subfigure[]{
    \begin{minipage}[b]{.23\linewidth}
        \centering
        \includegraphics[scale=0.35]{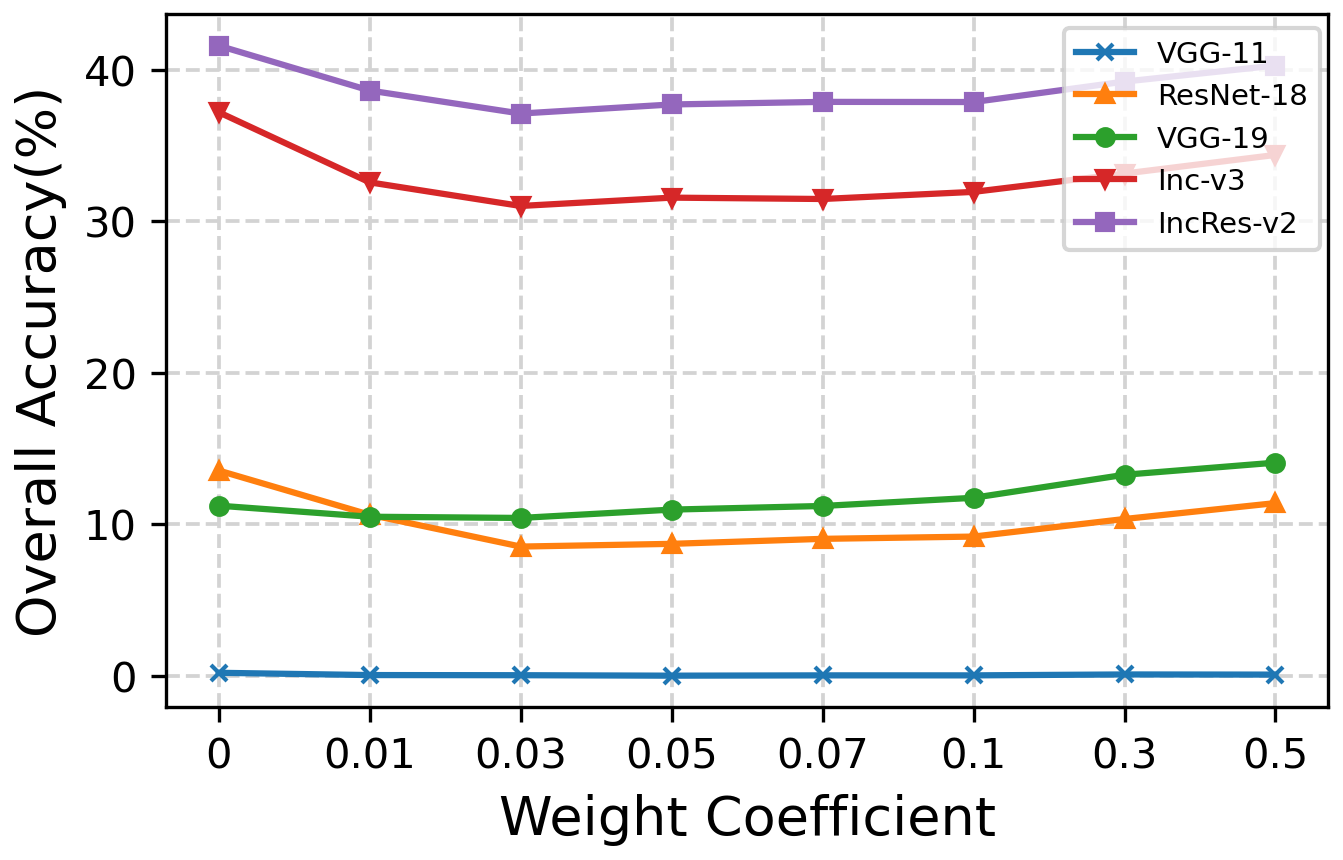}
    \end{minipage}
    \begin{minipage}[b]{.23\linewidth}
        \centering
        \includegraphics[scale=0.35]{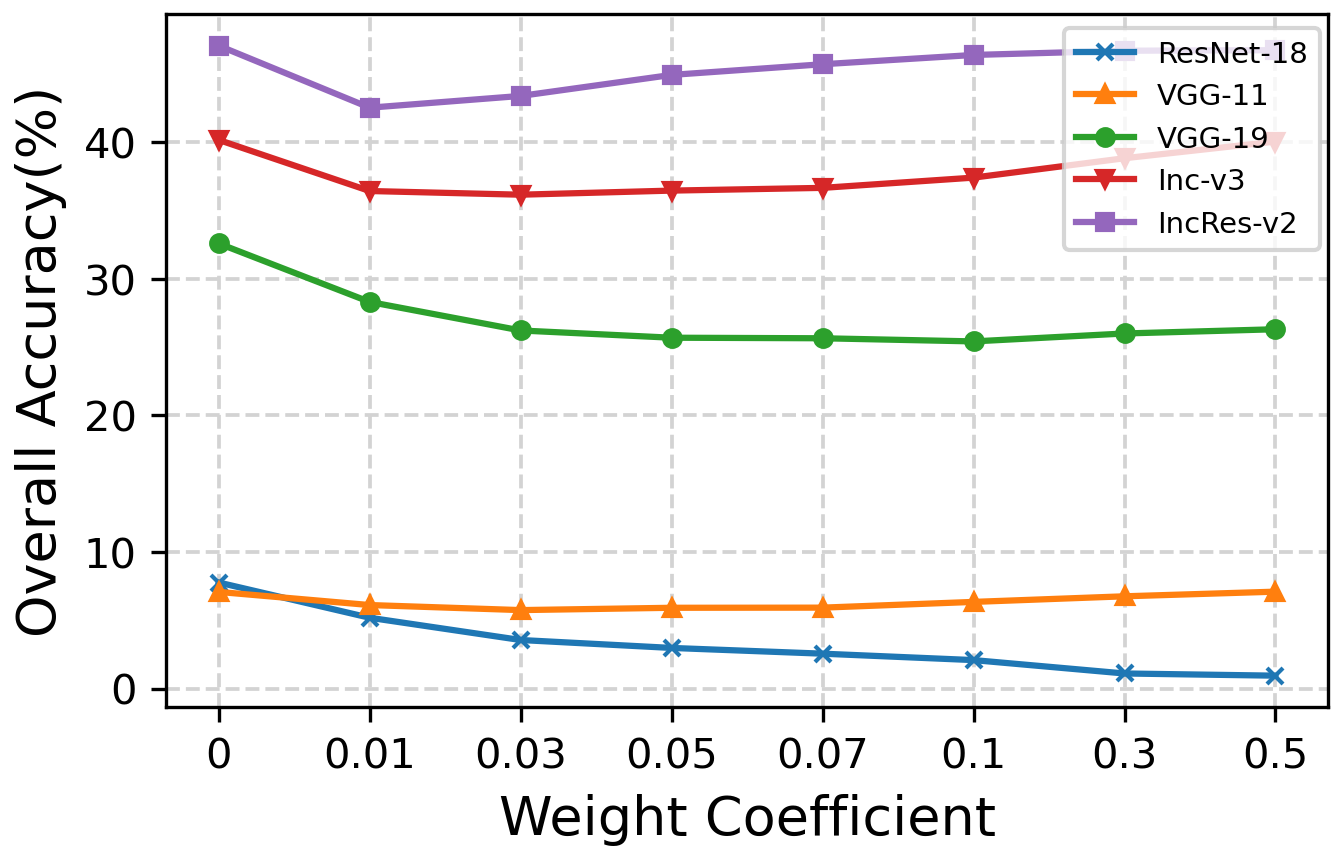} 
    \end{minipage}
    \label{weight coefficient}
}
\subfigure[]{
     \begin{minipage}[b]{.23\linewidth}
        \centering
        \includegraphics[scale=0.35]{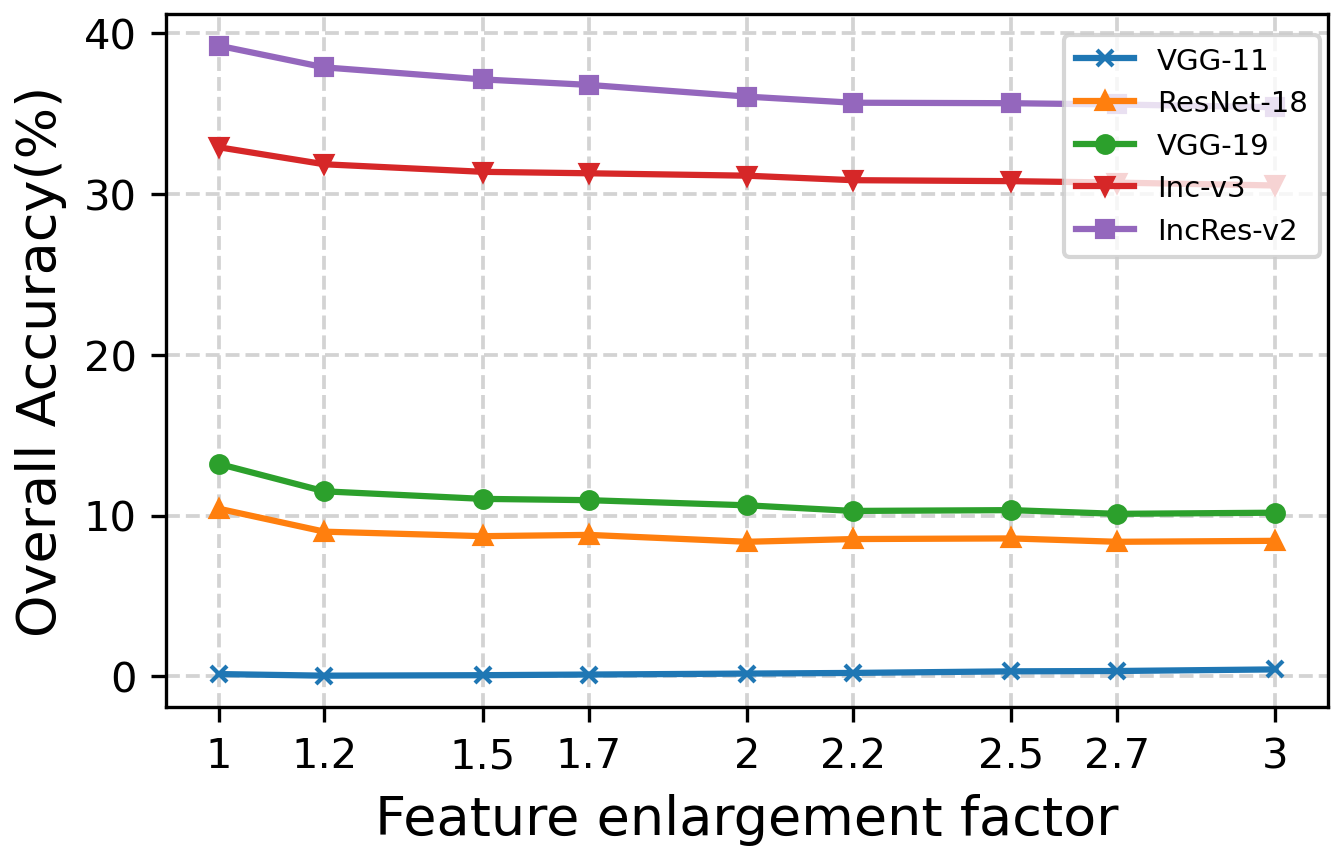}
    \end{minipage}
    \begin{minipage}[b]{.23\linewidth}
        \centering
        \includegraphics[scale=0.35]{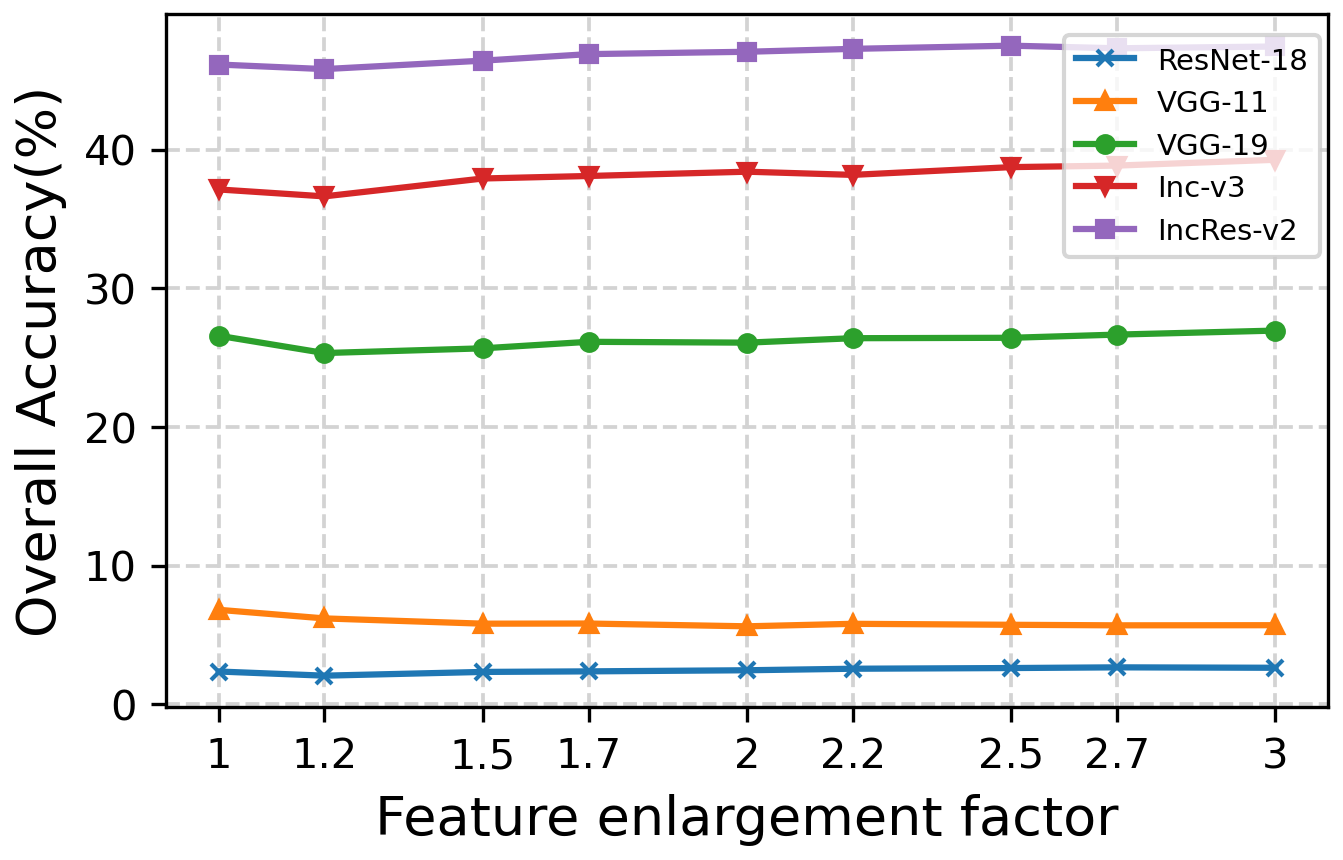}
    \end{minipage}
    \label{enlarge factor}
}
\caption{The impact of (a) the weight coefficient $\eta$ of the cross-entropy classification loss and (b) the enlargement factor $\lambda$ of the original example features on the results.}
\label{canshu b}
\end{figure*}

The impact of the weight coefficient of cross-entropy classification loss $ \eta $ on attack performance is shown in Fig. \ref{weight coefficient}. When $ \eta $ is set to 0, it indicates that the classification information of the output layer is not used. For nearly all models, classification accuracy is the highest at this point but subsequently shows a declining trend, indicating that appropriately incorporating the classification loss can enhance attack effectiveness. Whether using VGG-11 or ResNet-18 as the substitute model, the strongest attack effectiveness against black-box models is achieved when the weight $ \eta $ is set to 0.01 or 0.03. With 0.07 as the boundary, as the weight of the cross-entropy classification loss increases beyond this value, the attack effectiveness against black-box models decreases. Therefore, setting $ \eta $ to 0.03 yields superior experimental results.

The impact of the original feature values enlargement factor $\lambda$ is shown in Fig. \ref{enlarge factor}, with the enlargement factor ranging from 1 to 3. When the VGG-11 model serves as the substitute model, the classification accuracy on the black-box model consistently exhibits a declining trend. The attack performance is best when $\lambda$ is 3. However, for the ResNet-18 model, optimal performance occurs at a magnification factor of 1.2. Enlarging feature values as a method to optimize the perturbation direction, although it has limited impact on improving transferability attacks, presents a new approach. The improvement effect of this method on different substitute models also has differences, demonstrating that the impact of weighted feature divergence loss on results is closely tied to the model's architecture. In this paper, a small and valid value of 1.2 is selected, but the optimal parameter can also be selected according to the information of the substitute model. 

\subsection{Visualization Analysis}
\begin{figure*}[htbp]
\centering
\subfigure{
     \begin{minipage}[b]{.19\linewidth}
        \centering
        \includegraphics[height=3.1cm,width=3.1cm]{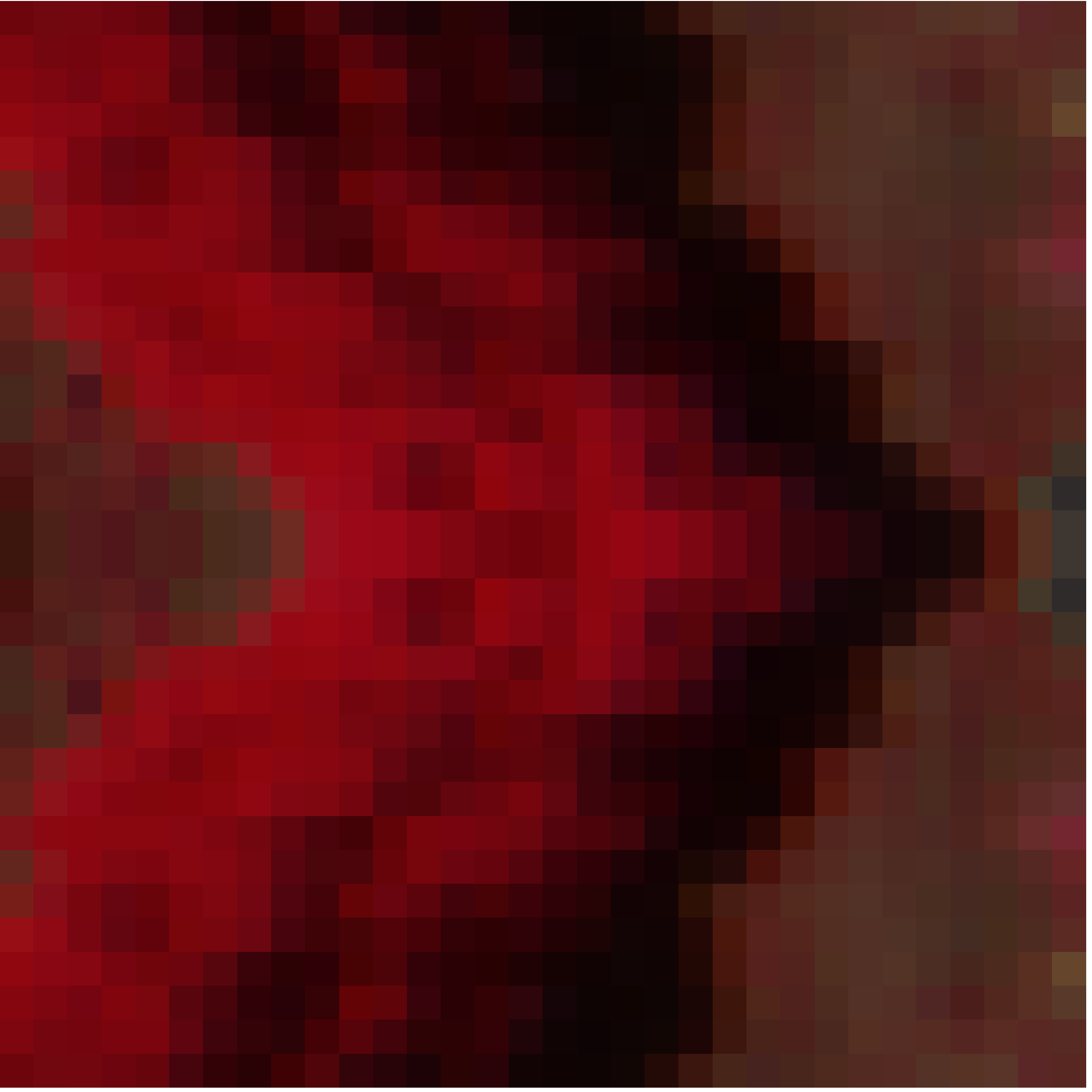}
    \end{minipage}
    \begin{minipage}[b]{.19\linewidth}
        \centering
        \includegraphics[height=3.1cm,width=3.1cm]{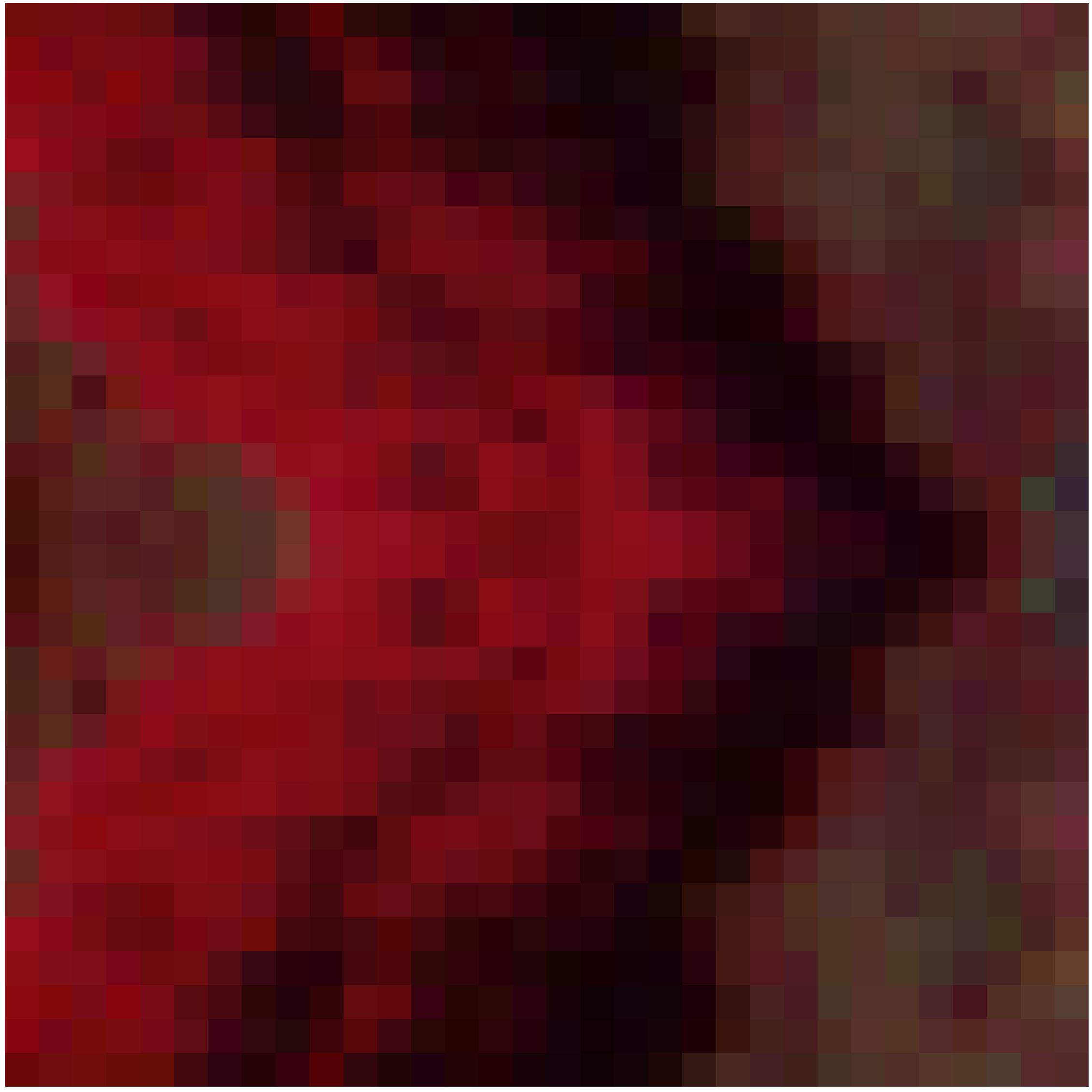}
    \end{minipage}  
    \begin{minipage}[b]{.19\linewidth}
        \centering
        \includegraphics[height=3.1cm,width=3.1cm]{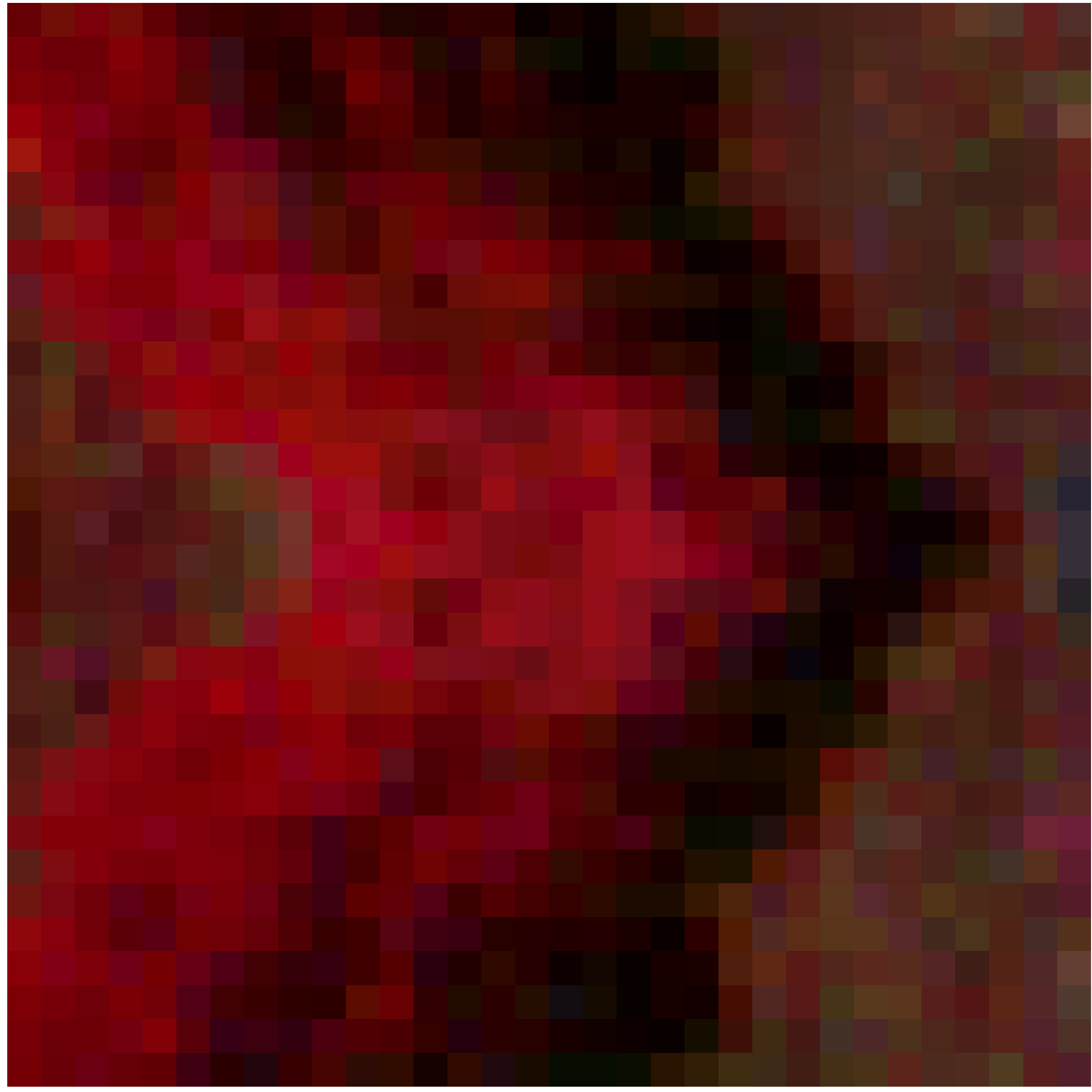}
    \end{minipage}
    \begin{minipage}[b]{.19\linewidth}
        \centering
        \includegraphics[height=3.1cm,width=3.1cm]{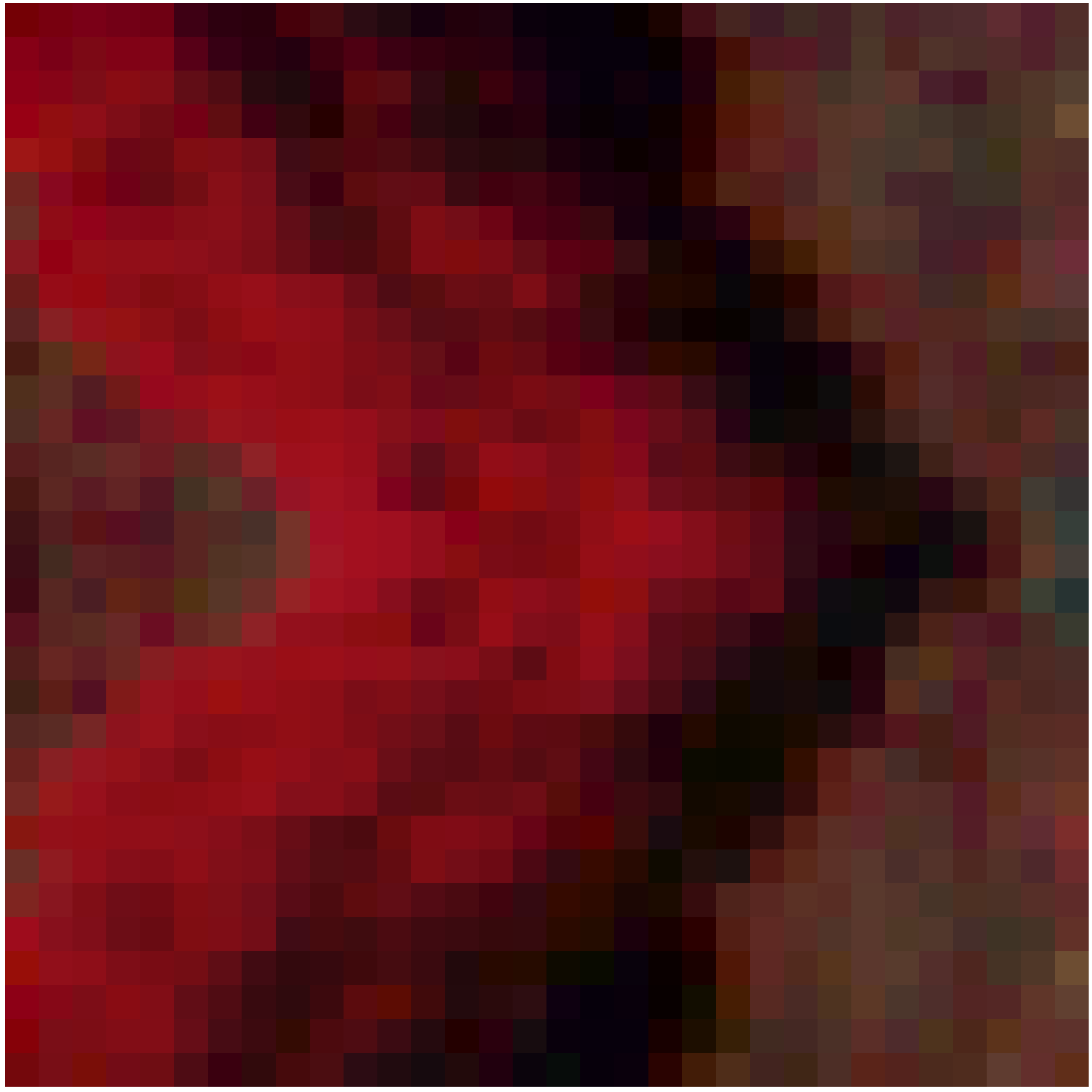}
    \end{minipage}
    \begin{minipage}[b]{.19\linewidth}
        \centering
        \includegraphics[height=3.1cm,width=3.1cm]{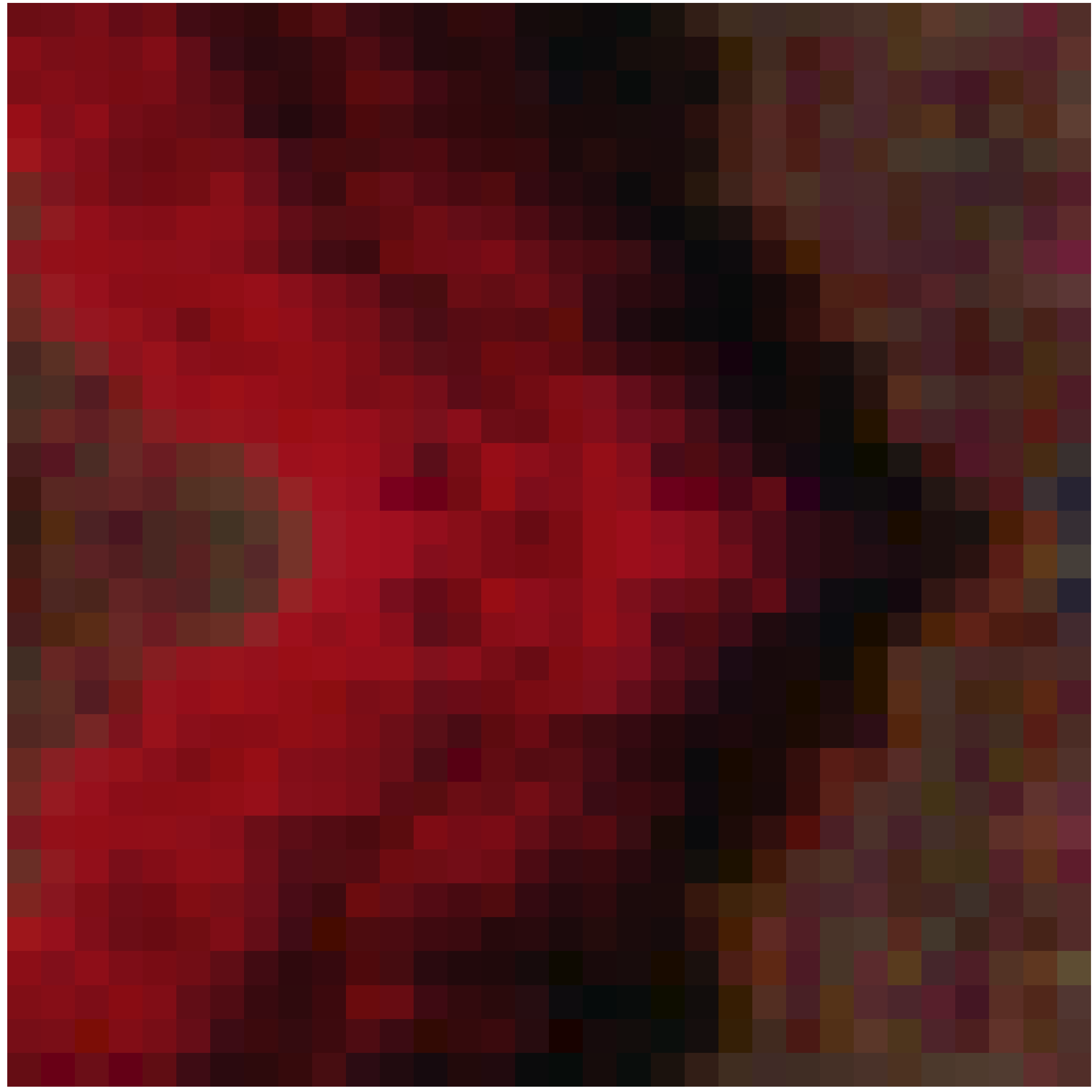}
    \end{minipage}
}\\
\subfigure{
    \begin{minipage}[b]{.19\linewidth}
        \centering
        \includegraphics[height=2.8cm,width=3.1cm]{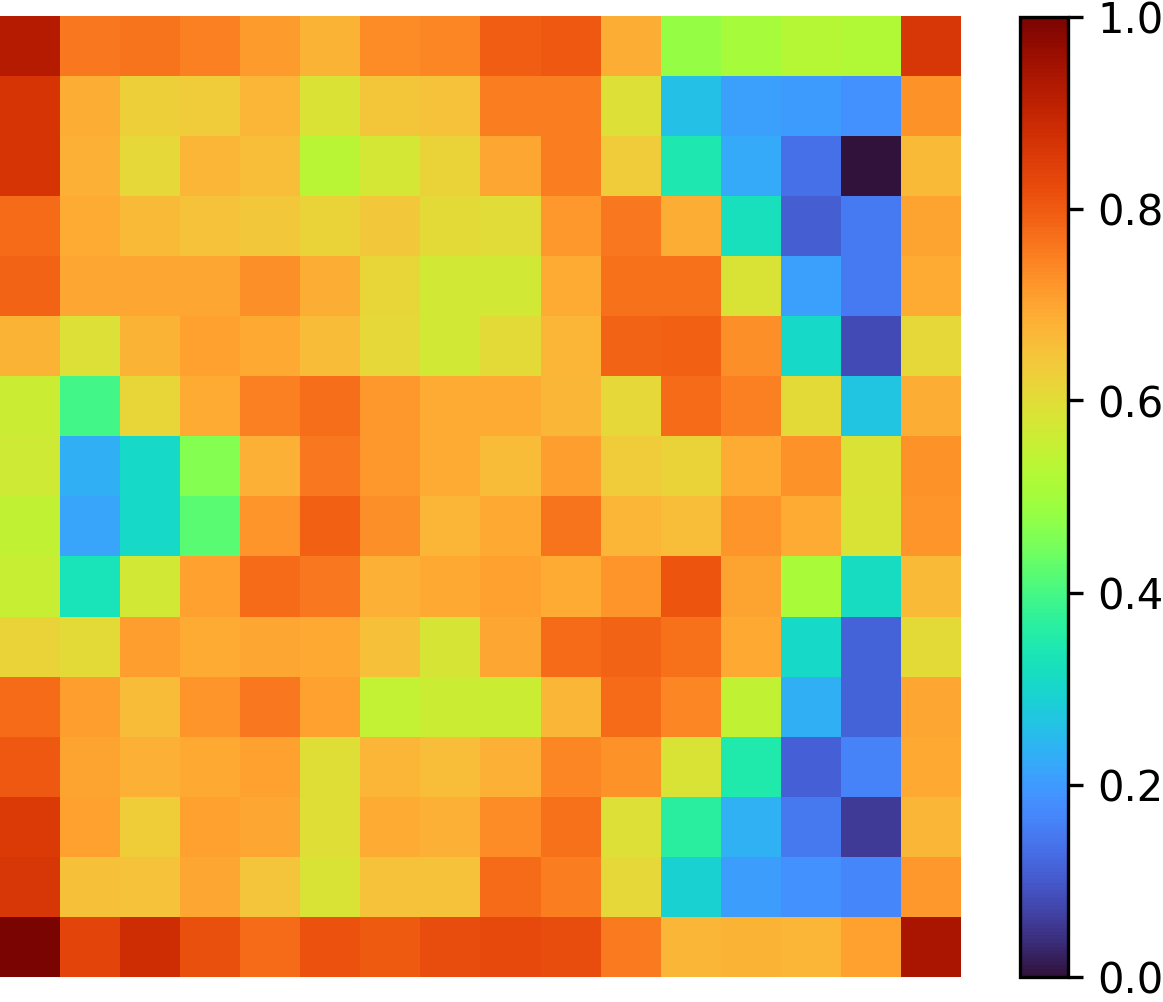}
    \end{minipage}
     \begin{minipage}[b]{.19\linewidth}
        \centering
        \includegraphics[height=2.8cm,width=3.1cm]{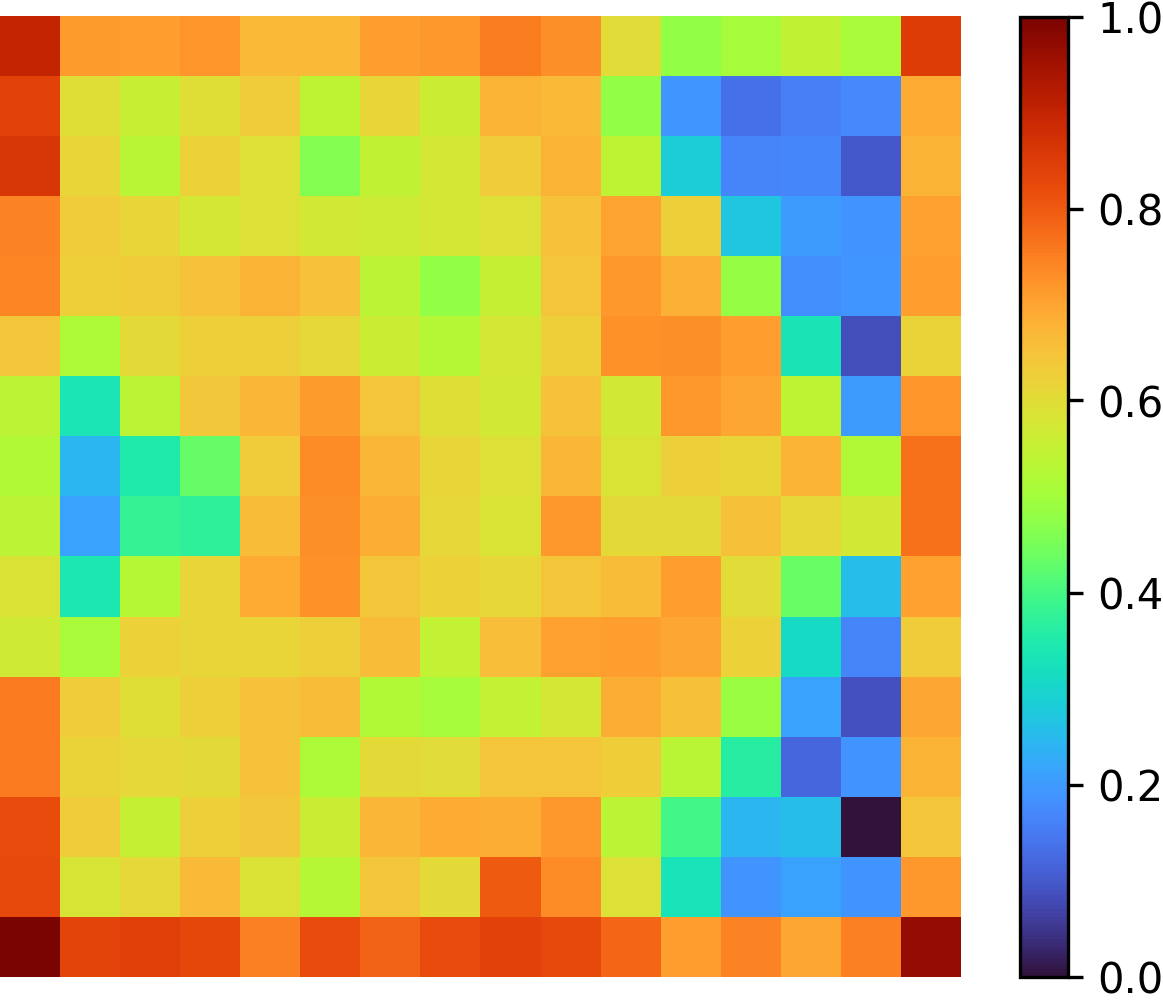}
    \end{minipage}
    \begin{minipage}[b]{.19\linewidth}
        \centering
        \includegraphics[height=2.8cm,width=3.1cm]{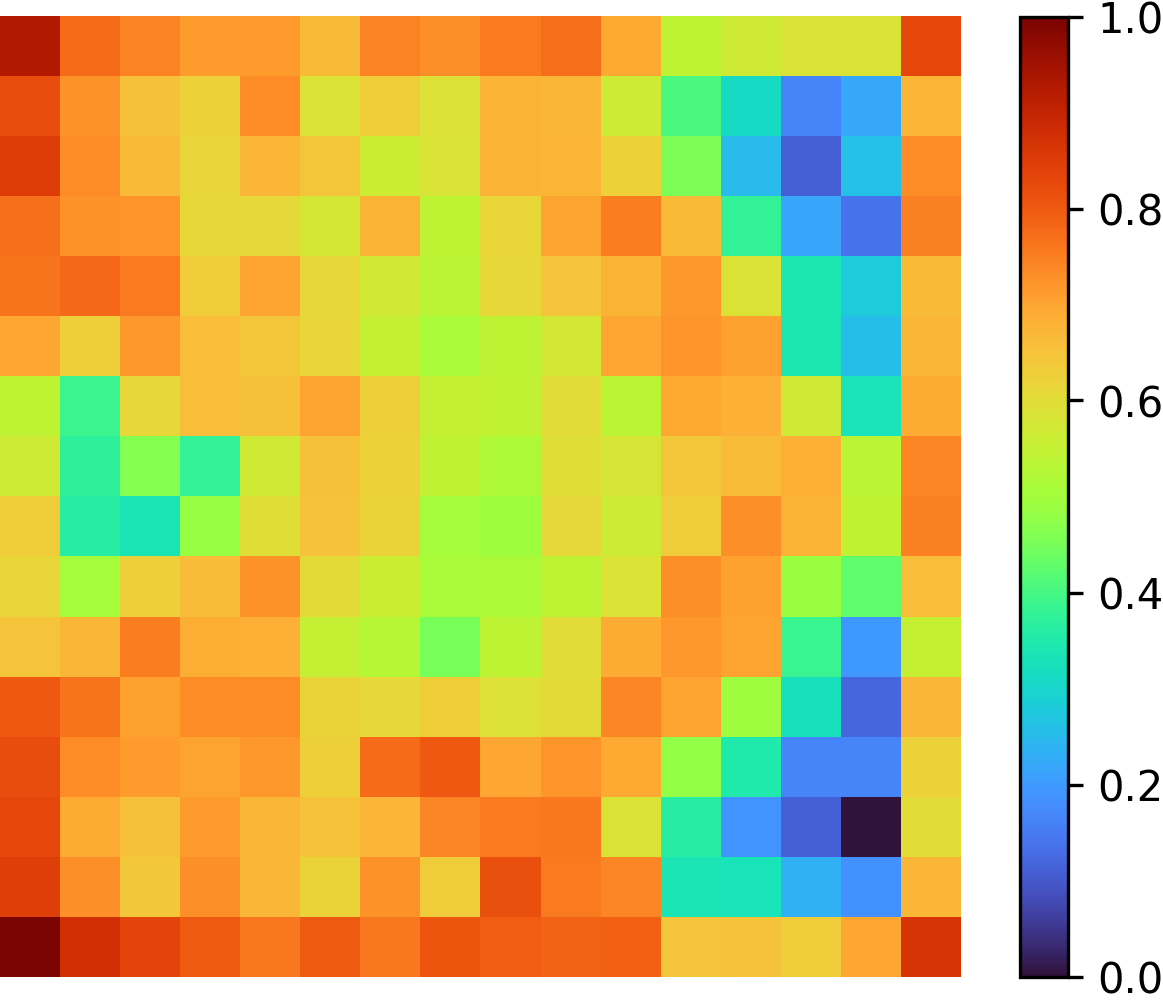}
    \end{minipage}  
    \begin{minipage}[b]{.19\linewidth}
        \centering
        \includegraphics[height=2.8cm,width=3.1cm]{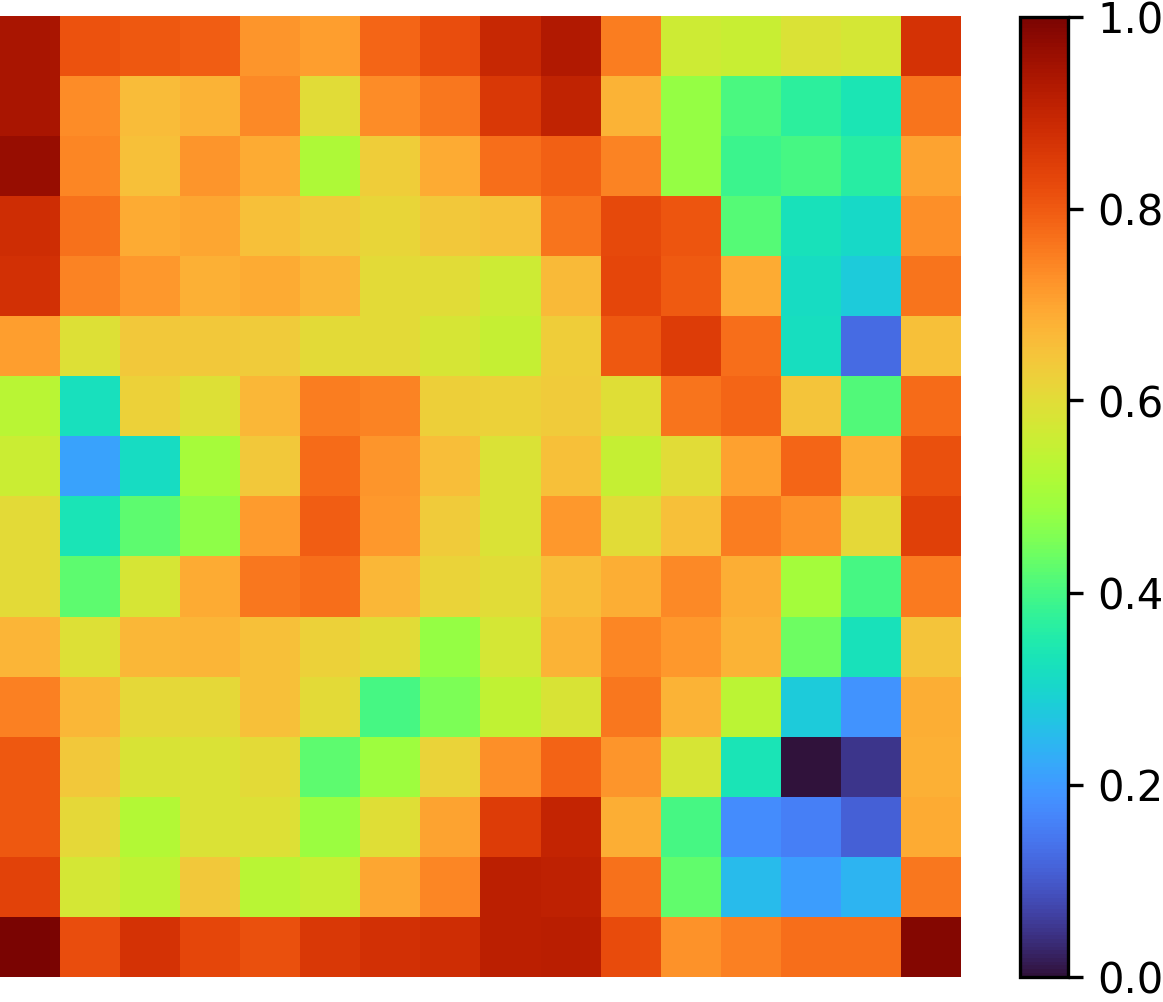}
    \end{minipage}
    \begin{minipage}[b]{.19\linewidth}
        \centering
        \includegraphics[height=2.8cm,width=3.1cm]{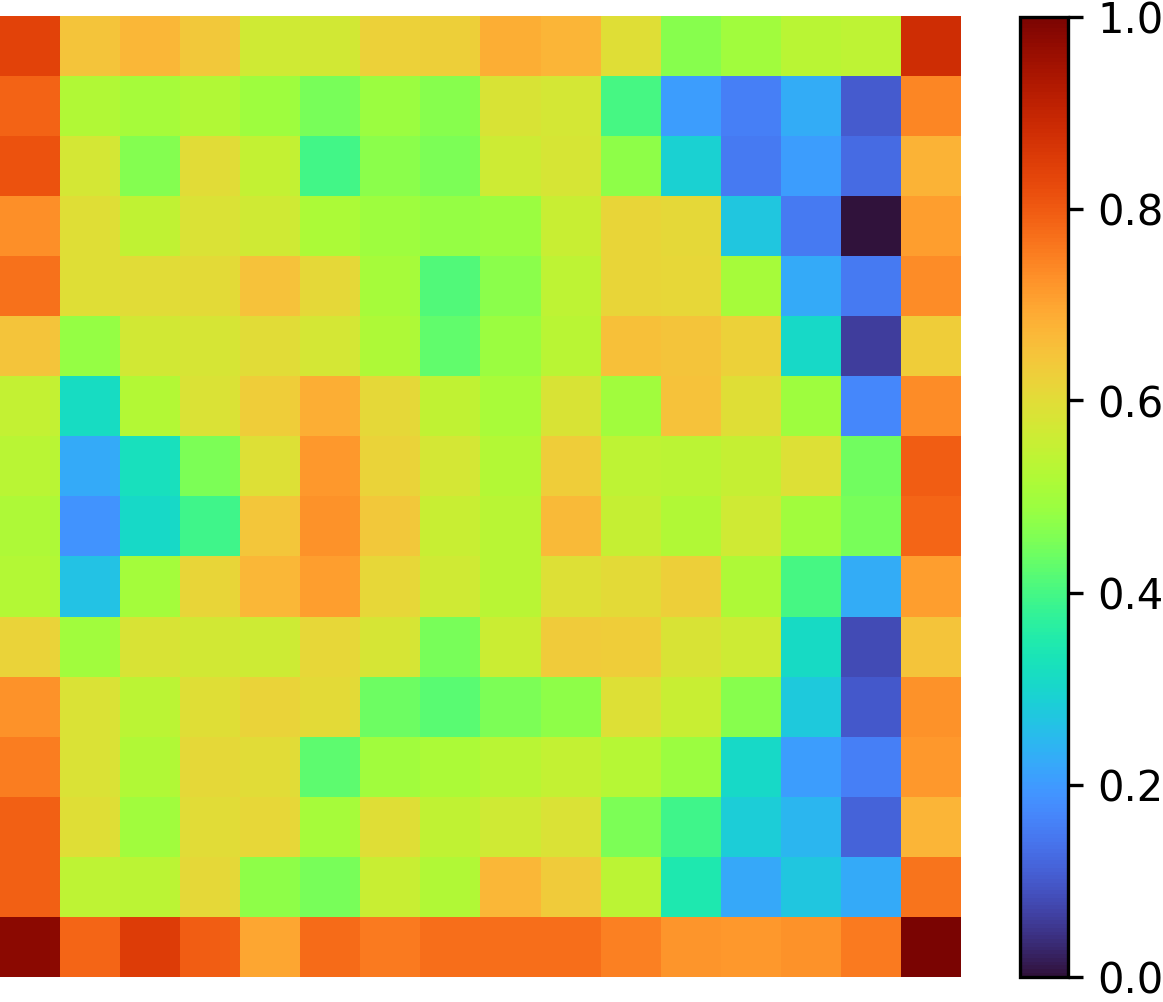}
    \end{minipage}  
}\\
\subfigure{
    \begin{minipage}[b]{.19\linewidth}
        \centering
        \includegraphics[height=3.1cm,width=3.1cm]{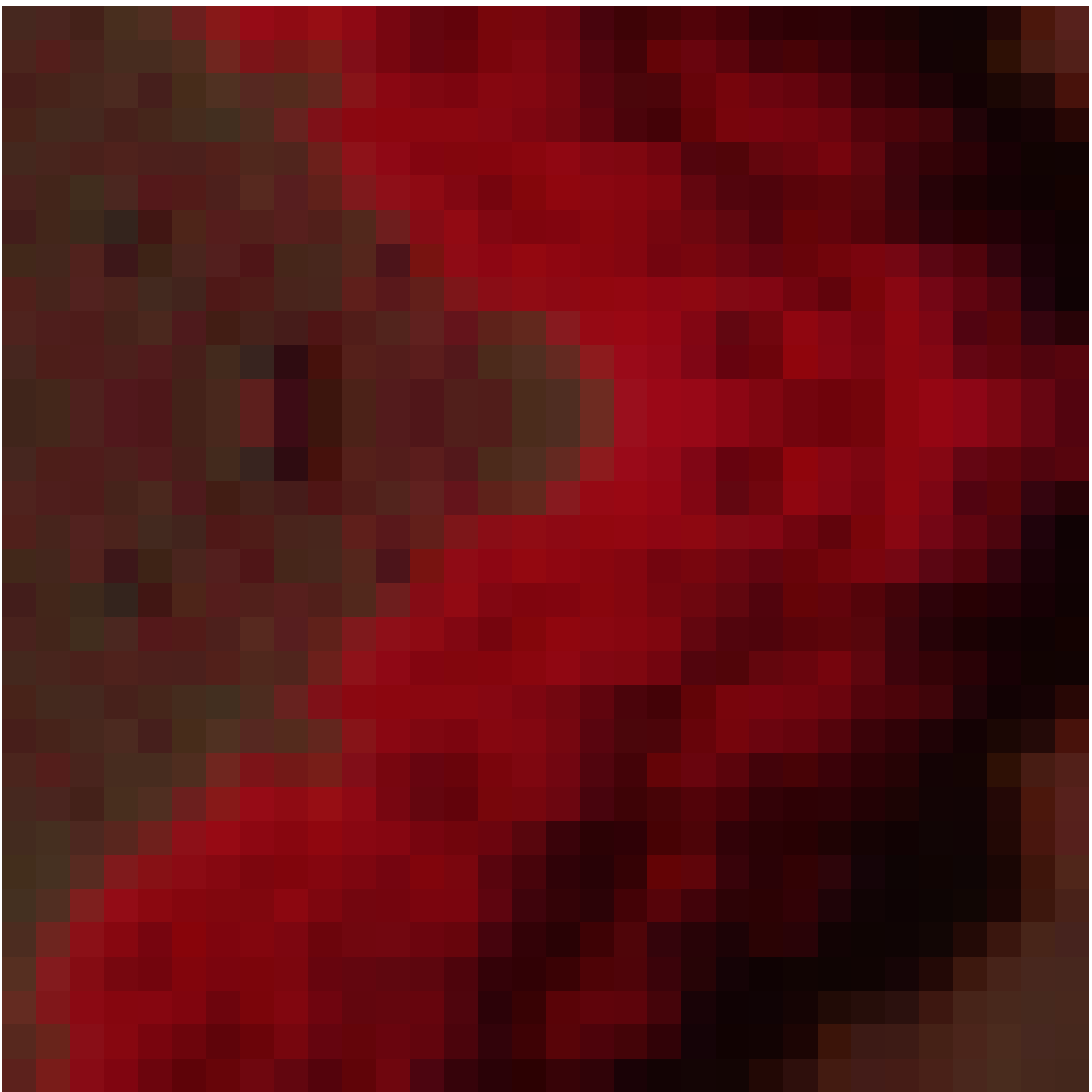}
    \end{minipage}
    \begin{minipage}[b]{.19\linewidth}
        \centering
        \includegraphics[height=3.1cm,width=3.1cm]{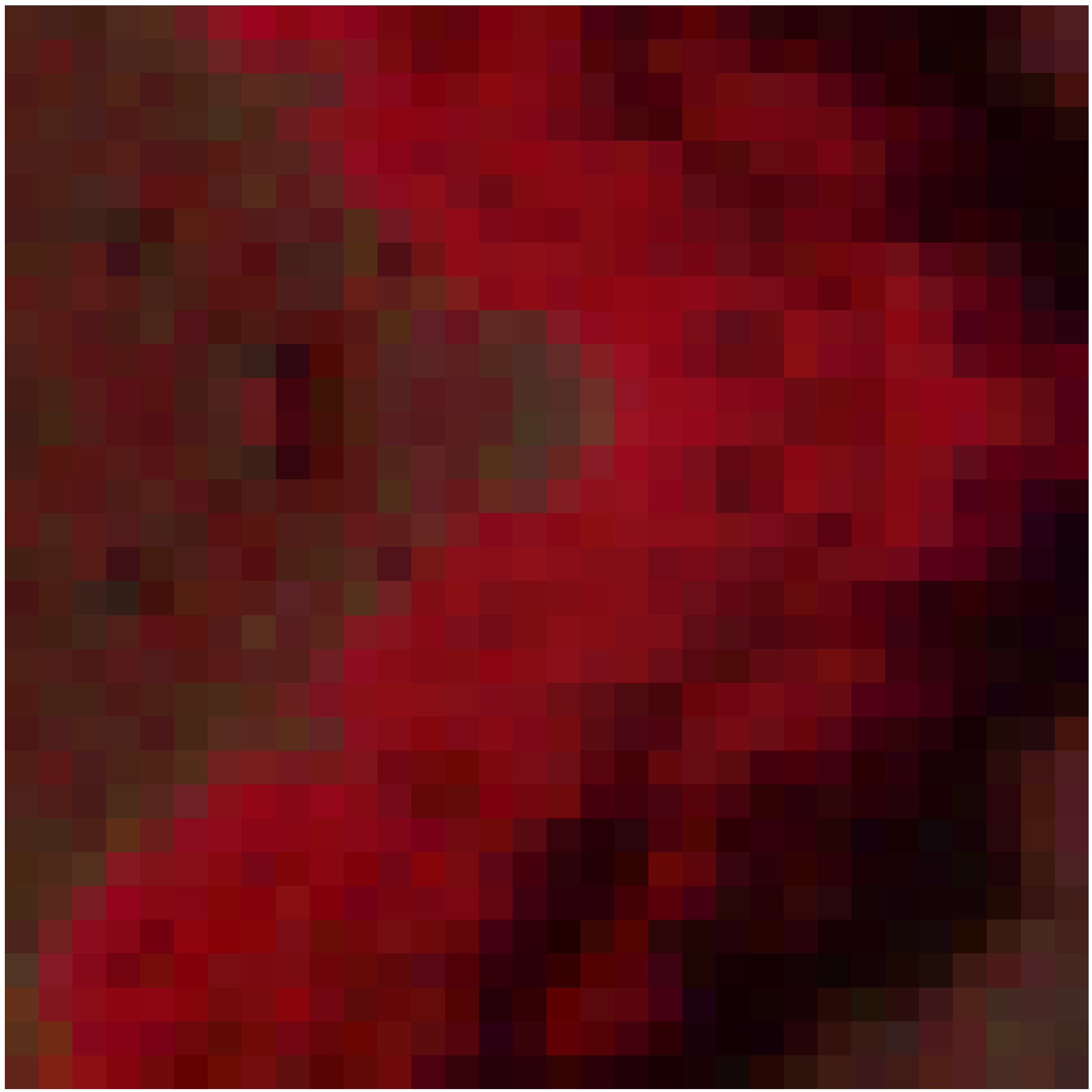}
    \end{minipage}  
    \begin{minipage}[b]{.19\linewidth}
        \centering
        \includegraphics[height=3.1cm,width=3.1cm]{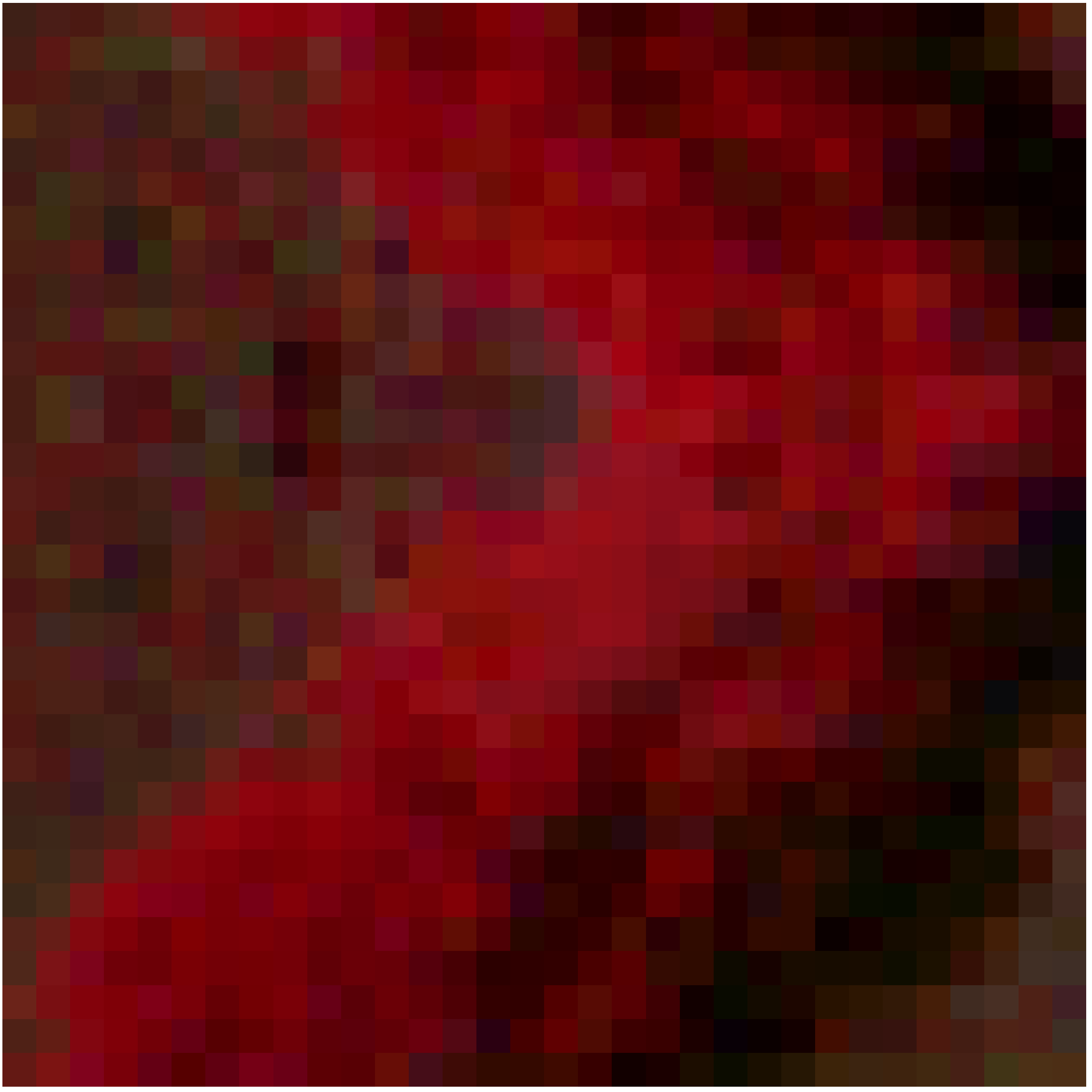}
    \end{minipage}
    \begin{minipage}[b]{.19\linewidth}
        \centering
        \includegraphics[height=3.1cm,width=3.1cm]{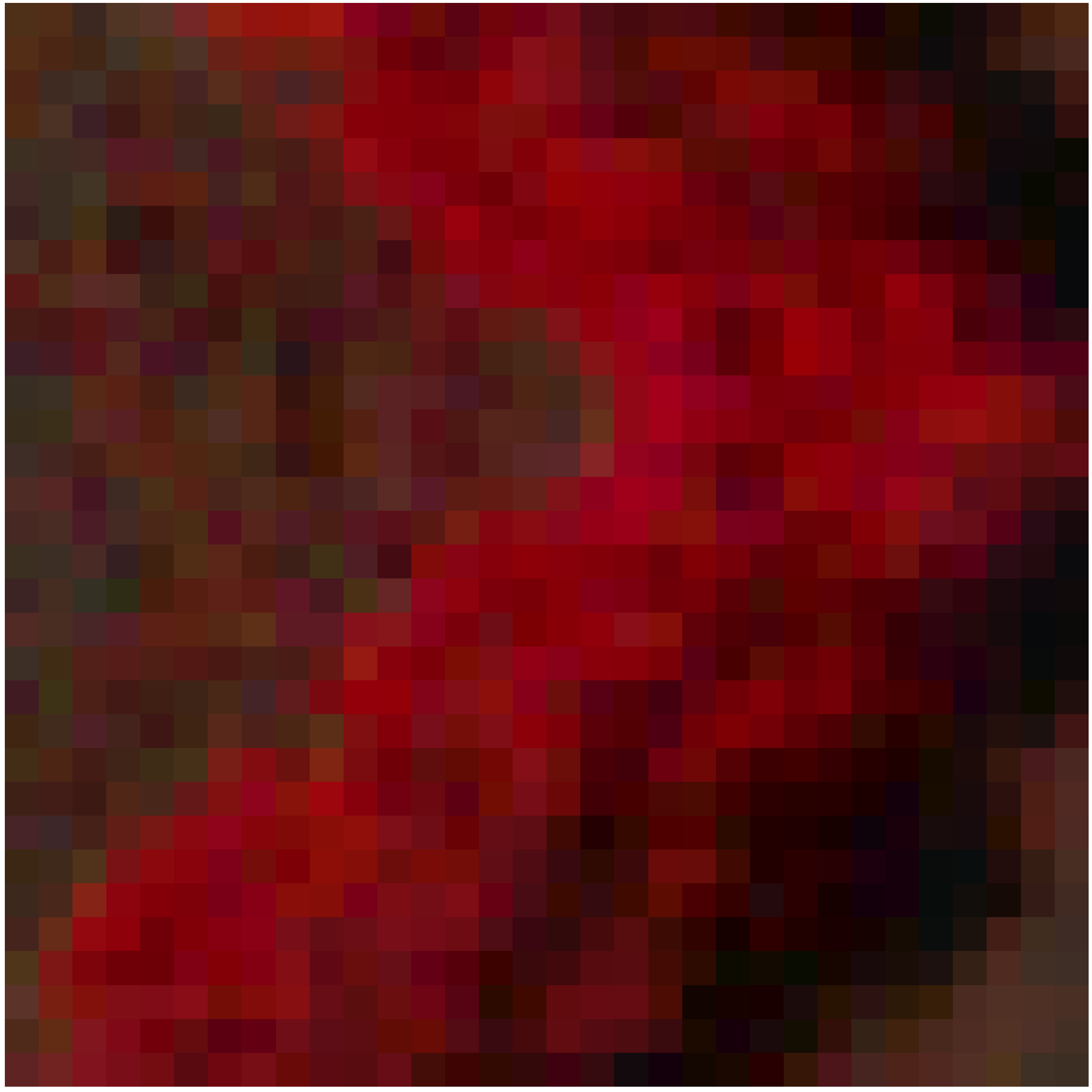}
    \end{minipage}
    \begin{minipage}[b]{.19\linewidth}
        \centering
        \includegraphics[height=3.1cm,width=3.1cm]{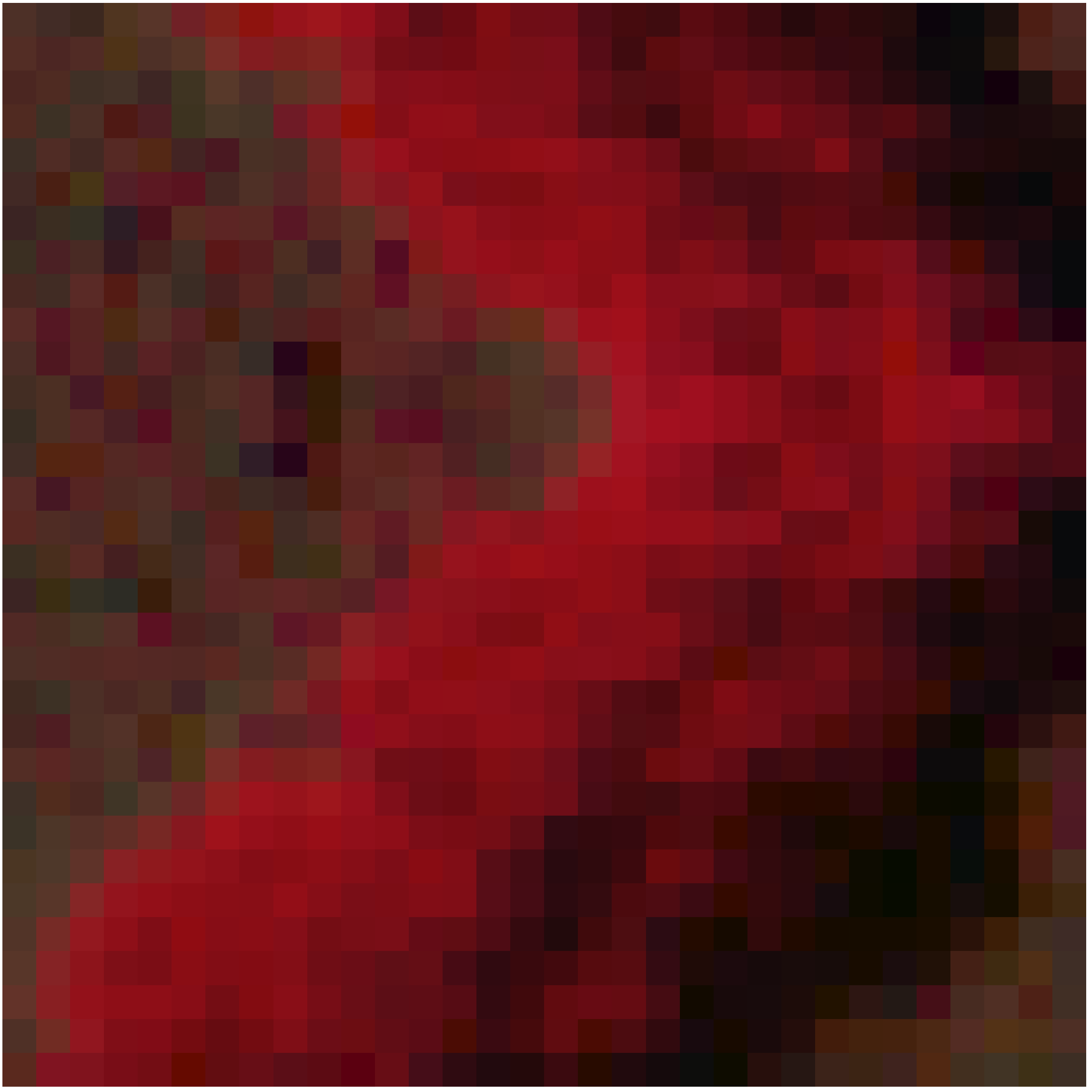}
    \end{minipage}
}
\subfigure{
    \begin{minipage}[b]{.19\linewidth}
        \centering
        \includegraphics[height=2.8cm,width=3.1cm]{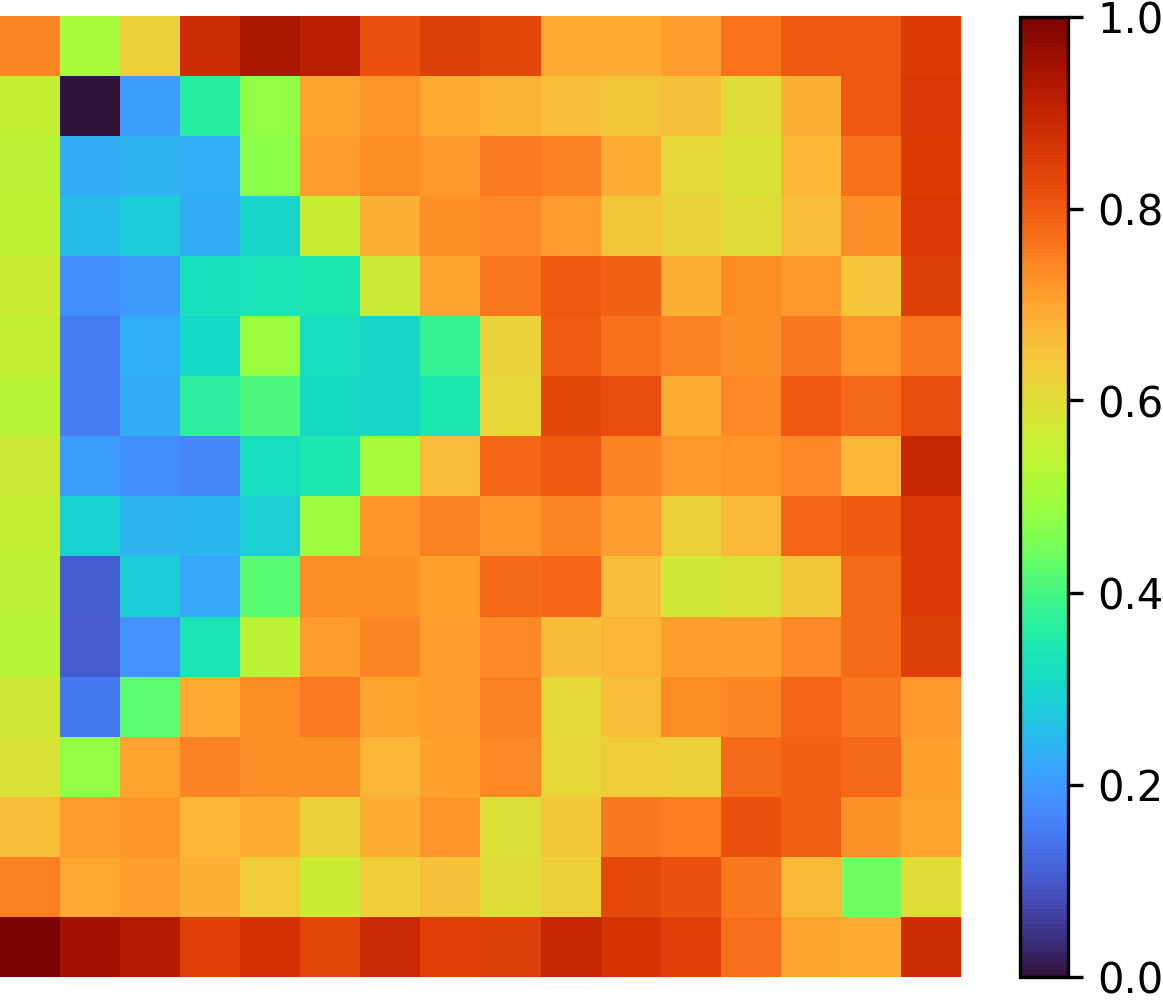}
        \\ Original
    \end{minipage}
     \begin{minipage}[b]{.19\linewidth}
        \centering
        \includegraphics[height=2.8cm,width=3.1cm]{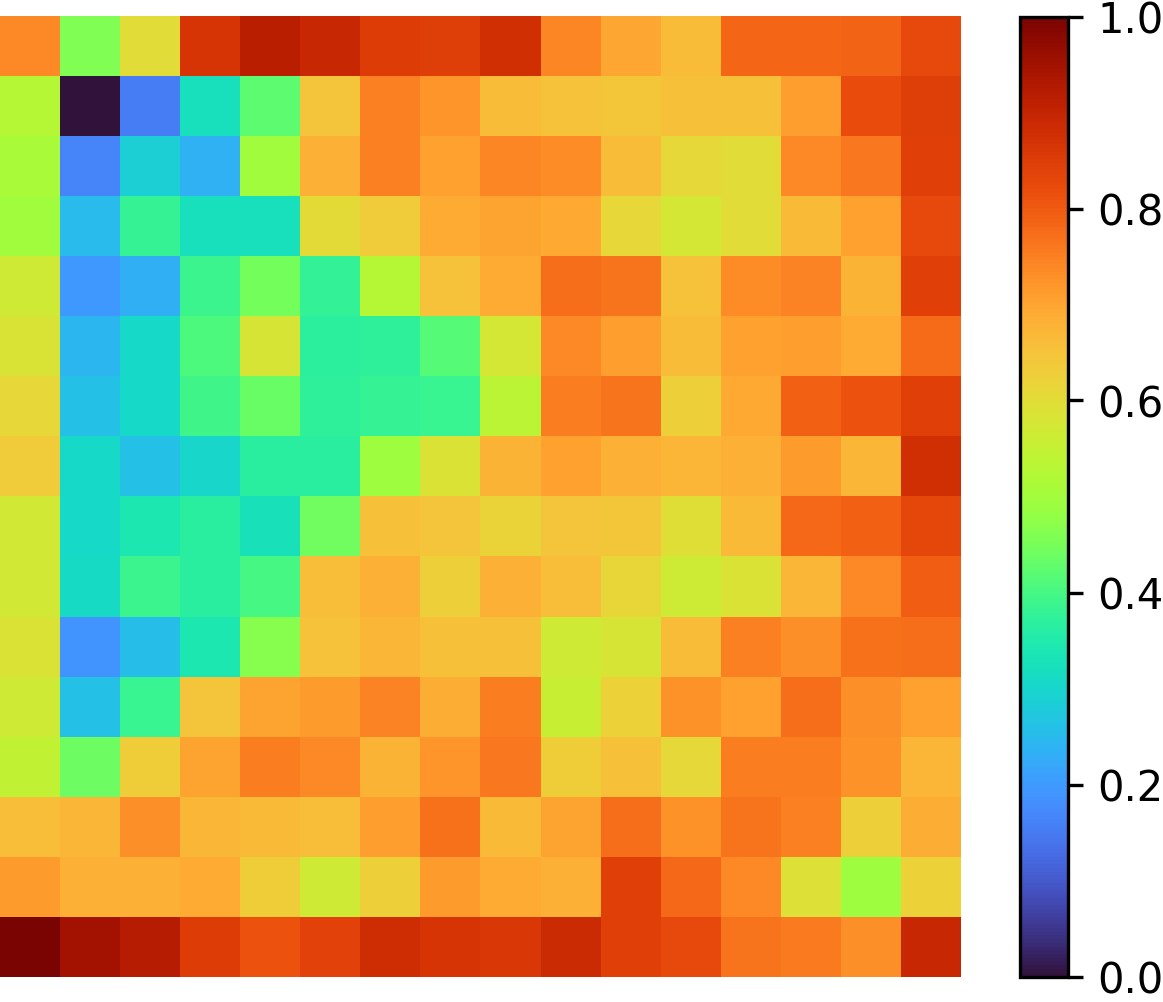}
        \\ SS
    \end{minipage}
    \begin{minipage}[b]{.19\linewidth}
        \centering
        \includegraphics[height=2.8cm,width=3.1cm]{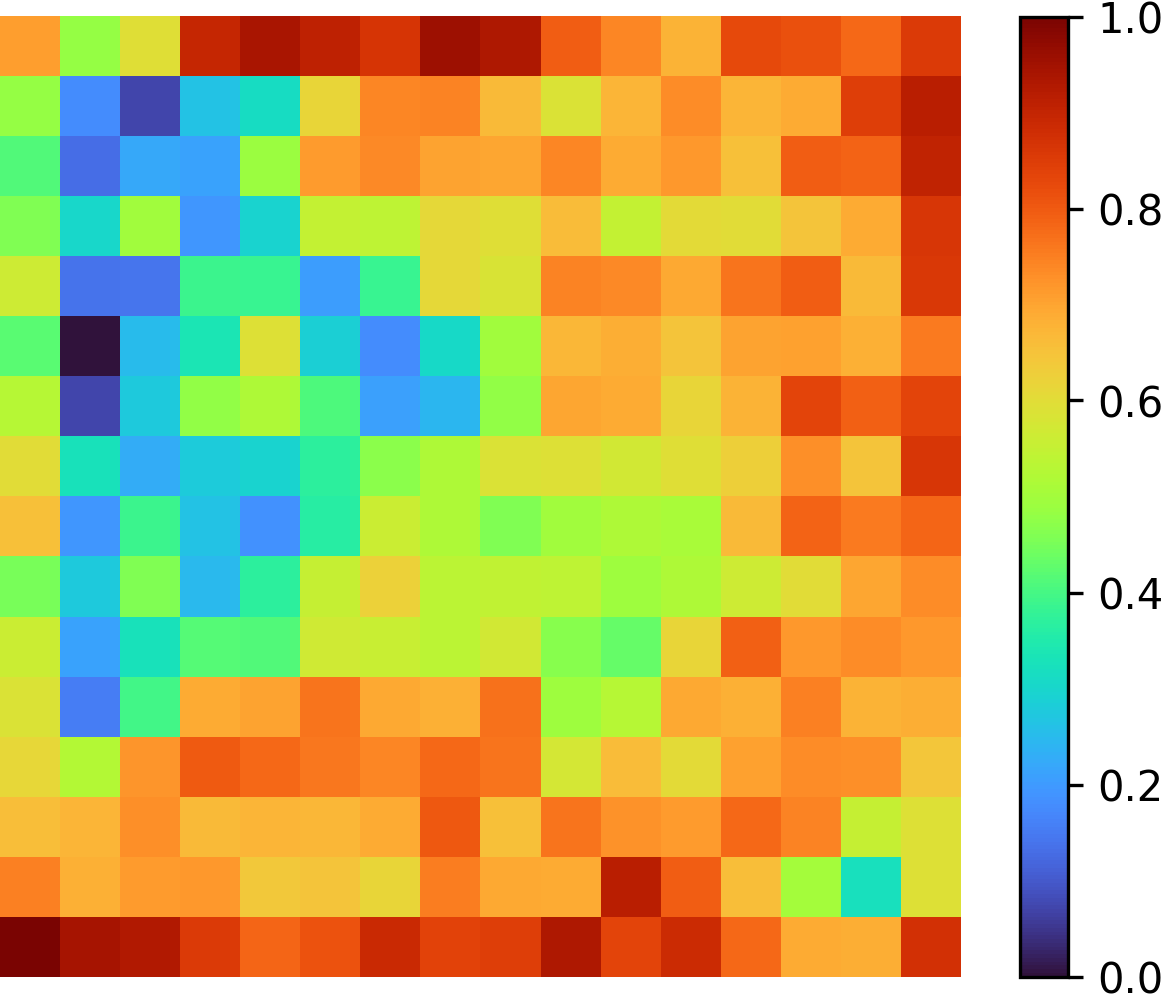}
        \\SIA
    \end{minipage}  
    \begin{minipage}[b]{.19\linewidth}
        \centering
        \includegraphics[height=2.8cm,width=3.1cm]{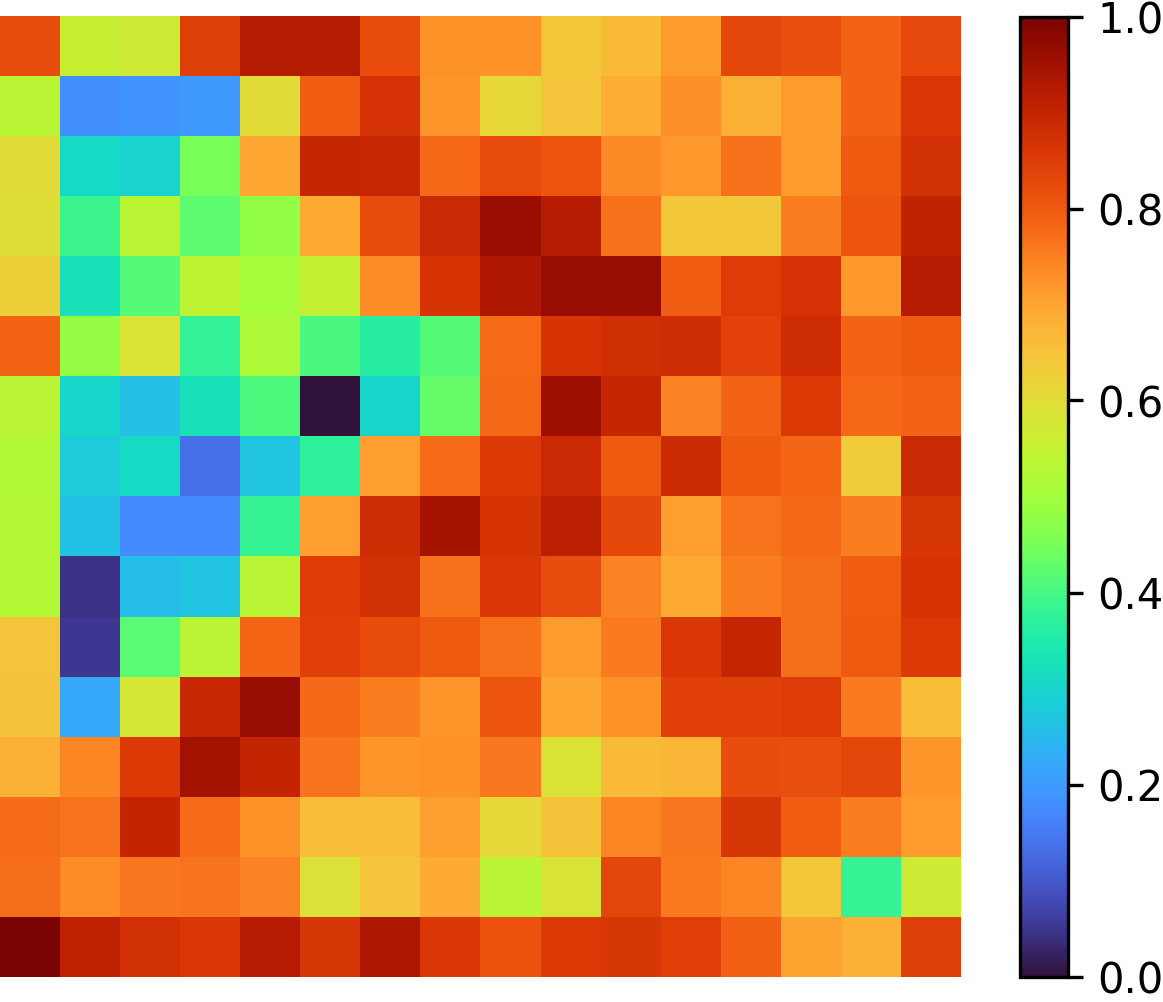}
        \\BSR
    \end{minipage}
    \begin{minipage}[b]{.19\linewidth}
        \centering
        \includegraphics[height=2.8cm,width=3.1cm]{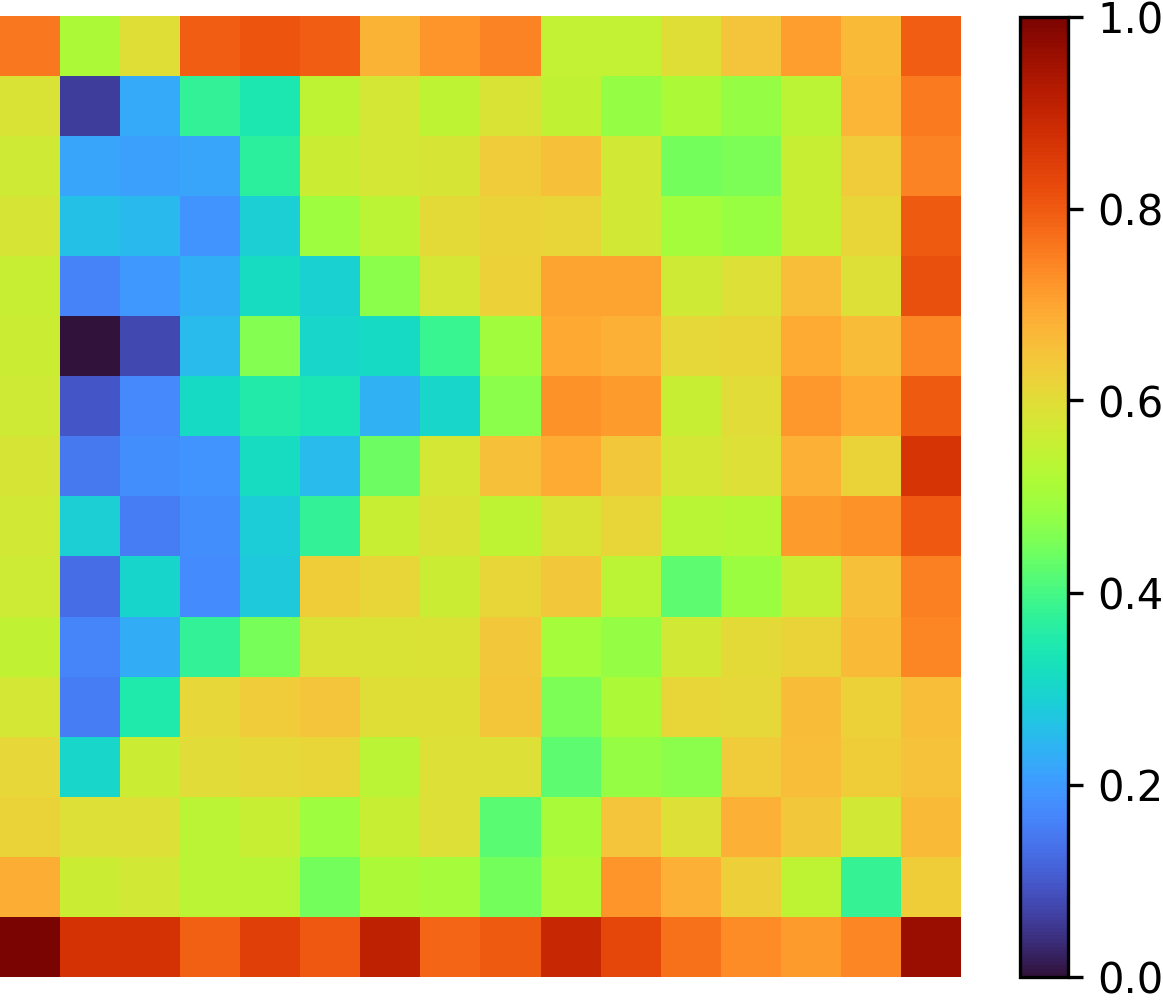}
        \\Ours
    \end{minipage}
}
\caption{The first row shows the false color images of the original example and its adversarial examples, while the second row shows the feature maps extracted from the VGG-11 model. The first column represents the original example, followed by the adversarial examples generated by SS, SIA, BSR, and our method. Two sets of examples are shown here.}
\label{Visualization}
\end{figure*}

To gain clearer insight into the internal differences between adversarial examples generated by our method and those produced by other approaches, we extracted and compared the feature maps of the example from the VGG-11 model, as shown in Fig. \ref{Visualization}. A darker orange indicates higher activation values, suggesting that these regions exert greater influence on the model's classification performance. Although the other methods sometimes shift the locations of originally high-activation regions, they do not markedly suppress them. In contrast, although there is no obvious difference between the pseudo-color images of the two examples, the adversarial example feature map generated by our method shows the biggest difference with the original feature map: the regions most attended to in the clean example are largely replaced by lighter colors and thus become far less salient. This prevents the model from focusing on the most critical regions and thereby achieves the desired attack effect.

\section{Conclusion}
For the HSI classification problem, existing methods for generating adversarial examples often suffer from overfitting to the substitute model and fail to fully utilize the feature maps extracted during training. Building on this, we propose a comprehensive attack method to improve the transferability of adversarial examples in HSI classification tasks. Considering that HSI contains a large number of spectral bands, we perform 3D block partitioning on the input samples in both spatial and spectral dimensions, and apply random transformations to each block. Additionally, we design a weighted feature divergence loss that enhances the difference between the feature maps of the original and adversarial examples, suppressing features related to the true class in the adversarial example, thereby improving transferability. Experiments show that the adversarial examples produced by this method achieve significant transferability on multiple black-box models, which not only deepens the understanding of adversarial vulnerabilities in HSI classification but also provides an important reference for developing more robust defense strategies in this field.

\section*{Acknowledgment}
Acknowledge for IEEE GRSS IADF as well as the Hyperspectral Image Analysis Laboratory at the University of Houston for providing the HoustonU 2018 dataset. This study was funded by the National Key Laboratory of Integrated Aircraft Control Technology Fund Project (WDZC2024601B02) and the National Natural Science Foundation of China (No.42371433).

\bibliographystyle{IEEEtran}
\bibliography{reference}

\end{document}